\documentclass[lettersize,journal]{IEEEtran}
\usepackage{amsmath,amsfonts, amssymb}
\usepackage{algorithmic}
\usepackage{algorithm}
\usepackage{array}
\usepackage{textcomp}
\usepackage{stfloats}
\usepackage{url}
\usepackage{verbatim}
\usepackage{graphicx, caption, subcaption}
\usepackage{cite}
\usepackage{diagbox}
\usepackage{amsthm}
\usepackage{relsize}
\usepackage[dvipsnames]{xcolor}
\usepackage{xcolor, soul}
\sethlcolor{yellow}

\newtheorem{theorem}{Theorem}

\newtheorem{lemma}{Lemma}

\newtheorem{assumption}{Assumption}

\newcommand{\R}{\mathbb{R}}

\newcommand{\modelOrderP}{\ensuremath{P}}

\newcommand{\conj}[1]{\ensuremath{\text{conj}({#1})}}

\newcommand{\eye}{\ensuremath{\mathbf{I}}}

\newcommand{\modelOrderK}{\ensuremath{K}}
\newcommand{\modelOrderQ}{\ensuremath{Q}}
\newcommand{\modelOrderM}{\ensuremath{M}}

\newcommand{\muA}{\ensuremath{\mu_A}}
\newcommand{\muB}{\ensuremath{\mu_B}}

\newcommand{\filterKernel}{\ensuremath{g}}

\newcommand{\graph}{\ensuremath{\mathcal{G}}}
\newcommand{\nodeSet}{\ensuremath{\mathcal{V}}}
\newcommand{\edgeSet}{\ensuremath{\mathcal{E}}}
\newcommand{\graphDim}{\ensuremath{N}}
\newcommand{\timeDim}{\ensuremath{T}}

\newcommand{\iNode}{\ensuremath{i}}
\newcommand{\jNode}{\ensuremath{j}}
\newcommand{\ijNode}{\ensuremath{ij}}
\newcommand{\node}{\ensuremath{v}}
\newcommand{\weightMat}{\ensuremath{\mathbf{W}}}

\newcommand{\graphSignal}{\ensuremath{\mathbf{x}}}
\newcommand{\graphSignalFilterOut}{\ensuremath{\mathbf {y}}}

\newcommand{\degreeMat}{\ensuremath{\mathbf{D}}}

\newcommand{\timeVertexSignalMat}{\ensuremath{\mathbf{X}}}

\newcommand{\timeVertexFilterOutSignalMat}{\ensuremath{\mathbf{Y}}}

\newcommand{\laplacian}{\ensuremath{\mathbf{L}}}
\newcommand{\graphLaplacian}{\ensuremath{\laplacian_\graph}}
\newcommand{\timeLaplacian}{\ensuremath{\laplacian_T}}
\newcommand{\jointLaplacian}{\ensuremath{\laplacian_J}}

\newcommand{\graphEigenvectorMat}{\ensuremath{\mathbf{U}_\graph }}
\newcommand{\graphEigenvalue}{\ensuremath{\lambda}}
\newcommand{\graphEigenvalueMat}{\ensuremath{\mathbf{\Lambda}}}

\newcommand{\kNN}[1]{{$#1$-{NN}}}

\newcommand{\timeEigenvalueMat}{\ensuremath{\mathbf{\Omega}}}

\newcommand{\timeSymbol}{\ensuremath{T}}

\newcommand{\iTime}{\ensuremath{t}}

\newcommand{\timeEigenvectorMat}{\ensuremath{\mathbf{U}_\timeSymbol }}
\newcommand{\jointEigenvectorMat}{\ensuremath{\mathbf{U}_J }}
\newcommand{\NT}{\ensuremath{{\graphDim \timeDim}}}
\newcommand{\covMat}{\ensuremath{{\mathbf{\Sigma}}}}

\newcommand{\xv}{\ensuremath{\bar {\mathbf{x}}}}
\newcommand{\yv}{\ensuremath{\bar {\mathbf{y}}}}
\newcommand{\tr}{\ensuremath{\text{tr}}}
\newcommand{\test}{\ensuremath{\theta}} 
\newcommand{\wt}{\ensuremath{{\mathbf{w}_t}}}
\newcommand{\wSignal}{\ensuremath{{\mathbf{w}}}}
\newcommand{\xSignal}{\ensuremath{{\mathbf{x}}}}
\newcommand{\ySignal}{\ensuremath{{\mathbf{y}}}}
\newcommand{\zSignal}{\ensuremath{{\mathbf{z}}}}
\newcommand{\indSet}{\ensuremath{\mathcal{I}}}
\newcommand{\indTriSet}{\ensuremath{\mathcal{T}}}
\newcommand{\psdCone}{\ensuremath{{\mathbb{S}}}}
\newcommand{\unitvect}{\ensuremath{\mathbf{u}}}

\newcommand{\avect}{\ensuremath{\mathbf{a}}}
\newcommand{\avectaug}{\ensuremath{\tilde{\mathbf{a}}}}
\newcommand{\bvect}{\ensuremath{\mathbf{b}}}
\newcommand{\ag}{\ensuremath{\avect^0}} 
\newcommand{\bg}{\ensuremath{\bvect^0}}
\newcommand{\cg}{\ensuremath{{\boldsymbol{\zeta}^0}}}
\newcommand{\as}{\ensuremath{\avect^\est}} 
\newcommand{\bs}{\ensuremath{\bvect^\est}}
\newcommand{\cs}{\ensuremath{{\boldsymbol{\zeta}^\est}}}
\newcommand{\cvect}{\ensuremath{{\boldsymbol{\zeta}}}}

\newcommand{\csp}{\ensuremath{\mathcal{S}}}
\newcommand{\uvect}{\ensuremath{\mathbf{u}}}
\newcommand{\vvect}{\ensuremath{\mathbf{v}}}
\newcommand{\vvectaug}{\ensuremath{\tilde{\mathbf{v}}}}
\newcommand{\Amat}{\ensuremath{\mathbf{A}}}
\newcommand{\Bmat}{\ensuremath{\mathbf{B}}}
\newcommand{\Rmat}{\ensuremath{\mathbf{R}}}
\newcommand{\est}{\ensuremath{\star}}
\newcommand{\locvect}{\ensuremath{\mathbf{v}}}

\newcommand{\Tbnd}{\ensuremath{\Upsilon}}
\newcommand{\Kbnd}{\ensuremath{\kappa}}
\newcommand{\Hc}{\ensuremath{\mathcal{H}}}
\newcommand{\Cg}{\ensuremath{C_{\cg}}}

\newcommand{\hest}{\ensuremath{{h^\est}}}

\newcommand{\hvect}{\ensuremath{\mathbf{h}}}
\newcommand{\eh}{\ensuremath{\tilde{ \mathbf{e}}}} 
\newcommand{\hinit}{\ensuremath{\tilde \hvect}} 
\newcommand{\hg}{\ensuremath{\hvect_\cg}} 
\newcommand{\hs}{\ensuremath{\hvect_\cs}} 
\newcommand{\hgapp}{\ensuremath{\breve{\hvect}_\cg}} 
\newcommand{\gh}{\ensuremath{\tilde{g}}}
\newcommand{\rvect}{\ensuremath{\mathbf{r}}}

\newcommand{\ztests}{\ensuremath{({\bar \zSignal}^\test)^\est}}
\newcommand{\ztestg}{\ensuremath{({\bar \zSignal}^\test)^0}} 
\newcommand{\ytest}{\ensuremath{\bar \ySignal^\test}}
\newcommand{\Sigxh}{\ensuremath{{\covMat_{\bar \xSignal}^\est}}}
\newcommand{\Sigx}{\ensuremath{{\covMat_{\bar \xSignal}}}}
\newcommand{\Sigzytesth}{\ensuremath{{(\covMat^{\test}_{\bar \zSignal \bar \ySignal})^\est}}}
\newcommand{\Sigzytest}{\ensuremath{{\covMat^{\test}_{\bar \zSignal \bar \ySignal}}}}
\newcommand{\Sigytesth}{\ensuremath{{(\covMat^{\test}_{\bar \ySignal})^\est}}}
\newcommand{\Sigytest}{\ensuremath{{\covMat^{\test}_{\bar \ySignal}}}}
\newcommand{\delSigzytest}{\ensuremath{{\mathbf{\Delta}^{\test}_{_{\bar \zSignal \bar \ySignal} }}}}
\newcommand{\delSigytest}{\ensuremath{{\mathbf{\Delta}^{\test}_{_{ \bar \ySignal} }}}}

\newcommand{\dc}{\ensuremath{d}}

\newcommand{\Wis}{\ensuremath{\mathbf{\Phi}}}

\begin{document}


\title{Learning Graph ARMA Processes from Time-Vertex Spectra}

\author{Eylem Tu\u g\c ce G\"uneyi, Berkay Yald\i z, Abdullah Canbolat and Elif Vural

\thanks{The authors are with the Dept. of Electrical and Electronics Engineering, METU, Ankara. This work was supported by the Scientific and Technological Research Council of Turkey (T\"UB\.ITAK) under grant 120E246. The codes of our method are available at https://github.com/eylemtugce/JS-ARMA.}}



\maketitle

\begin{abstract}
The modeling of time-varying graph signals as stationary time-vertex stochastic processes permits the inference of missing signal values by efficiently employing the correlation patterns of the process across different graph nodes and time instants. In this study, we propose an algorithm for computing graph autoregressive moving average (graph ARMA) processes based on learning the joint time-vertex power spectral density of the process from its incomplete realizations for the task of signal interpolation. Our solution relies on first roughly estimating the joint spectrum of the process from partially observed realizations and then refining this estimate by projecting it onto the spectrum manifold of the graph ARMA process through convex relaxations. The initially missing signal values are then estimated based on the learnt model. Experimental results show that the proposed approach achieves high accuracy in time-vertex signal estimation problems.

\end{abstract}


\section{Introduction}
\IEEEPARstart{M}{any} modern digital platforms involve the acquisition of data over networks, while network data has a typically time-varying structure. For instance, measurements acquired on a sensor network or user data in a social network often vary over time. Such data can be modeled as time-varying graph signals, or \textit{time-vertex} signals. In many practical applications, time-vertex signals may have missing observations due to issues such as sensor failure, connection loss, and partial availability of user statistics. Hence, the spatio-temporal interpolation of time-vertex signals arises as an important problem of interest. Similarly, in forecasting applications, one would like to predict future values of a time-vertex signal based on its past values. All these problems necessitate the computation of signal models that can accurately fit to the characteristics of data. Stationary graph process models are of potential interest for a wide range of data types where the correlation patterns between different nodes evolve in line with the topology of the graph, such as data resulting from message passing, diffusion, or filtering operations over irregular networks. In this work, we consider a setting where possibly partial observations of a collection of time-vertex signals are available, and study the problem of learning parametric stochastic graph processes from data for signal inference tasks such as interpolation and forecasting.

\begin{figure}[t]
\centering
  \includegraphics[width=8cm]{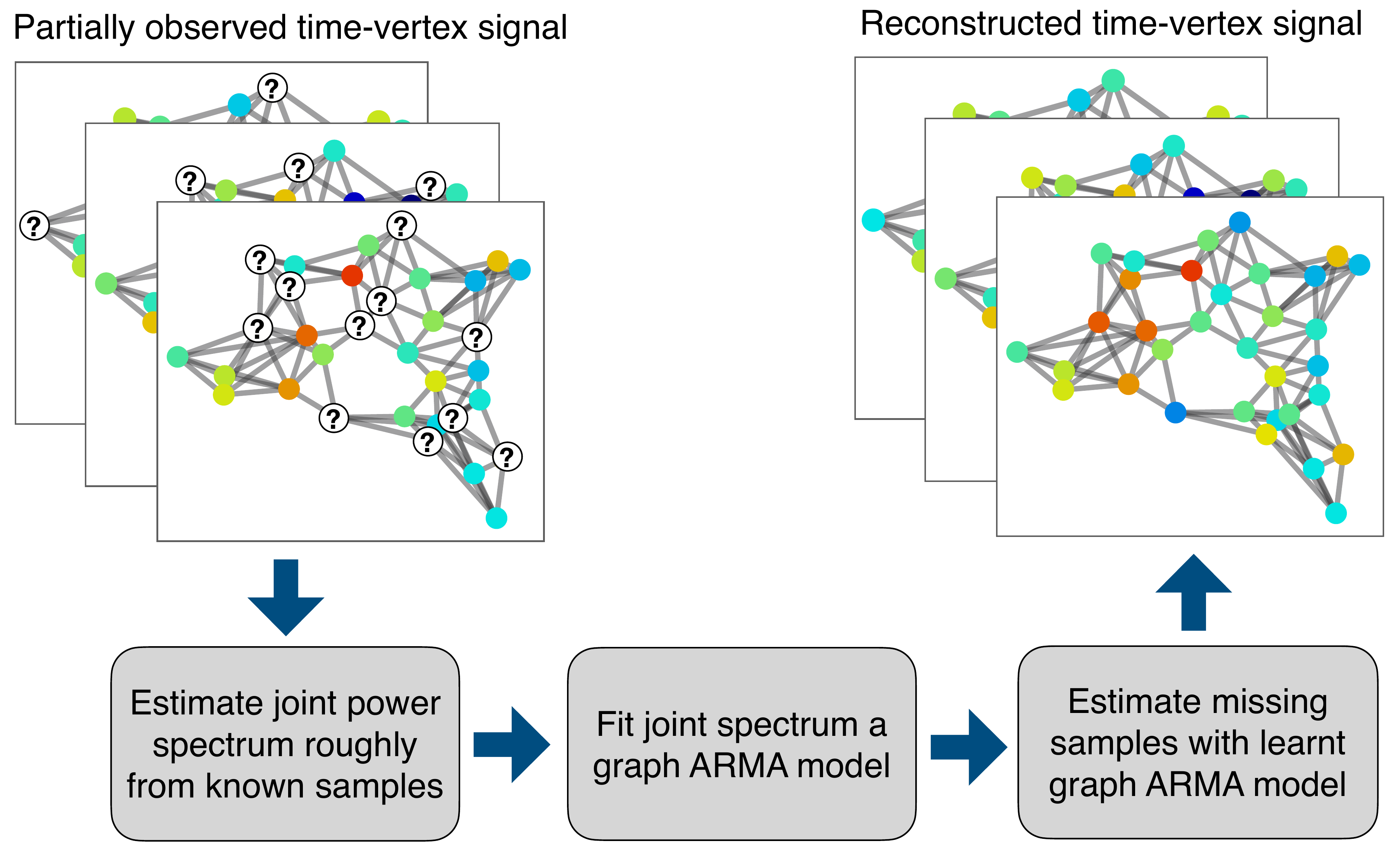}
    \caption{Illustration of the proposed method}
  \label{fig_illus_method}
\end{figure}

The modeling of time-vertex signals via stochastic processes has been addressed in several previous works, where the traditional definition of random processes in regular domains has been extended to graph domains so as to permit the modeling of both graph signals (vertex signals) \cite{Girault-Isometric-shift, Natheneal-Graph-Stationarity, Marques-Stationarity} and time-vertex signals \cite{Natheneal-Joint-Stationarity} as stationary stochastic processes.  Among stationary process models, ARMA models widely used in classical signal processing have also been adapted to graph domains in several recent works \cite{MeiM17, IsufiLPL19}. Meanwhile, the computation of an ARMA process model is a challenging problem in graph domains as it typically involves the solution of highly nonlinear and nonconvex optimization problems. The problem of learning graph ARMA process models has been addressed in the previous studies \cite{MeiM17, IsufiLPL19, Marques-Stationarity}; however, none of these studies explicitly aim to capture the specific time-vertex joint spectral characteristics of graph data. The method in \cite{Marques-Stationarity} aims to fit AR (autoregressive) and ARMA vertex process models to the observed power spectral density and tackles the nonconvexity of the problem through several strategies; however, treats graph signals only in the vertex domain and does not consider the time dimension of graph data. The estimation algorithm in \cite{MeiM17} and the GP-VAR method in \cite{IsufiLPL19} consider time-vertex signals, however adopt the relatively simpler AR model instead of the more elaborate ARMA model. The G-VARMA algorithm in  \cite{IsufiLPL19} breaks down the nonconvex graph ARMA model learning problem into the computation of individual temporal ARMA models, which are still nonconvex but relatively well-studied in the time domain. However, this algorithm has the disadvantage of resulting in a large number of unknown model variables due to the underlying nonparametric signal model. Moreover, focusing particularly on the forecasting problem, the methods in \cite{IsufiLPL19} do not seek to capture the joint time-vertex spectral characteristics of signals, as the model parameters are fit directly to the process realizations in the mean squared error sense.


In this paper, we propose a novel algorithm for learning parametric graph ARMA process models from multiple realizations\footnote{The assumption regarding the availability of multiple realizations of the process is often met in practice, e.g., due to the time-periodicity inherent in many data types. For instance, each 24-hour meteorological measurement sequence over a network can be regarded as a different realization of a time-vertex process. In lack of such periodicity, a solution would be to parse the data along the time dimension with an appropriately chosen time length.} of the process as illustrated in Fig.~\ref{fig_illus_method}. Our main contribution over the previous studies \cite{Marques-Stationarity, Natheneal-Joint-Stationarity, MeiM17, IsufiLPL19} is that it is the first method to learn graph ARMA process models by explicitly employing the information of the joint time-vertex spectrum of signals. The main difficulty regarding the computation of graph ARMA models lying in the nonconvexity of the associated problems, we demonstrate that the original nonconvex problem can be relaxed into a convex problem that can be solved efficiently.  We consider the relatively challenging setting where the available realizations of the process are assumed to be partially observed, i.e., the time-vertex signals at hand may have missing values at arbitrary time instants and graph nodes. Our method relies on the idea of first obtaining a rough estimate of the time-vertex joint power spectral density (JPSD) of the process from its partially known observations through the estimation of its sample covariance matrix. This initial estimate of the JPSD is typically corrupted due to noise and the fact that it is obtained from a finite number of samples. Under the assumption that the data conforms to a graph ARMA model, we fit the initial JPSD estimate the parameters of a graph ARMA process with the aim of improving its accuracy.  The resulting optimization problem being nonconvex and thus difficult to solve, we relax it into a convex optimization problem through a series of approximations, which can then be solved accurately via semidefinite quadratic linear programming. Once the graph process model is learnt in this way, the initially missing observations of the realizations are estimated from the second-order statistics of the process. Experimental results on several real data sets show that the proposed method achieves state-of-the-art performance in time-vertex signal estimation tasks. We also provide a theoretical sample complexity analysis of the problem of learning graph ARMA models from a finite set of realizations and show that the estimation error of the JPSD and the missing process observations decrease  at rate $O(1/\sqrt{L})$ as the number $L$ of realizations increases. 

The rest of the paper is organized as follows: In Section \ref{sec_rel_work}, we discuss the related literature. In Section \ref{sec_basics_gsp}, we give a brief overview of the theory of stochastic graph processes. Then in Section \ref{sec_prop_method}, we present the proposed method for learning graph ARMA processes. In Section \ref{sec_samp_complx}, we present our theoretical sample complexity analysis for learning graph ARMA models. In Section \ref{sec_exp_res} we experimentally evaluate the performance of our method and in Section \ref{sec_concl} we conclude.

\section{Related Work}
\label{sec_rel_work}

The inference of graph signals is a well-studied problem that allows for a wide scope of solutions. Methods based on traditional semi-supervised learning techniques typically rely on regularization on graph domains  \cite{ZhouBLWS03, JungHMJHE19, BergerHM17}, which have been extended to time-vertex signals in various recent works through smoothness priors along the vertex and the time dimensions \cite{GiraldoMGTB22, JiangTSO20, QiuMSWLG17}. The iterative reconstruction techniques in these papers have also motivated deep algorithm unrolling methods \cite{ChenE21a}, \cite{KojimaNYT23}, \cite{NagahamaYTCE22}, while several other works have proposed graph neural network models for the reconstruction of time-vertex signals \cite{HadouKR23}, \cite{CastroCorreaGMBBM23}. In the recent years, another common approach for reconstructing graph signals has been the band-limitedness assumption \cite{LorenzoBBS16, LorenzoBIBL18, YangYYH21, ChamonR18}, which has been employed in time-vertex signal inference problems as well \cite{IsufiBLL20}. The representation of time-vertex signals has also been addressed via the concept of product graphs, where the temporal dimension of graph signals is modeled by edges that connect vertices at different time instants. Several studies have focused on the design of time-vertex analysis and synthesis filter banks \cite{JiangFTX21}, the sampling \cite{OrtizJimenezCC18}, and the reconstruction of time-vertex signals  \cite{OrtizJimenezCC18}, \cite{RomeroIG17} through product graphs.


Among the graph signal inference methods relying on stochastic models,  stationary process models are of particular interest. The concept of stationarity in classical signal processing theory has first been extended to irregular graph domains in the leading studies \cite{Girault-Stationarity, Natheneal-Graph-Stationarity, Marques-Stationarity, Natheneal-Joint-Stationarity}. The common idea in these works is to explicitly account for the graph topology in the definition of stationarity, differently from traditional multivariate process models \cite{Lutkepohl05},  \cite{Jung15}. Girault et al.~have defined wide sense stationary graph processes  \cite{Girault-Stationarity} via isometric graph translations \cite{Girault-Isometric-shift}, while the definition of stationarity is based on graph localization operators in \cite{Natheneal-Graph-Stationarity} and graph shift operators in \cite{Marques-Stationarity}. These works have been extended to time-varying graph signals through the definition of joint time-vertex stationarity in the succeeding studies \cite{Natheneal-Towards-Joint-Stationarity,Natheneal-Joint-Stationarity}, which show that joint stationarity can be characterized through the time-vertex filtering of white processes \cite{Time-vertex-signal-processing, IsufiLSL16}.

The methods in \cite{Natheneal-Graph-Stationarity}, \cite{Natheneal-Joint-Stationarity} estimate the joint power spectral density (JPSD) of graph processes through the joint time-vertex Fourier transform of process realizations. While these algorithms use nonparametric representations, in several other works stationary graph processes have been defined via parametric models.  The concept of  AR and ARMA processes traditionally used in the modeling of time-series data has been extended to graph domains in the recent studies  \cite{IsufiLPL19, Marques-Stationarity, MeiM17}. In classical signal processing theory, the computation of ARMA process models for time-series data requires the solution of a nonlinear equation system. Although various techniques exist in the classical literature such as approximations using Durbin's method or PSD factorization solutions via modified Yule-Walker equations \cite{hayes96}, it is not straightforward to adapt these methods to graph domains due to the presence of the vertex dimension in addition to the time dimension. The computation of an ARMA model is a nonlinear and nonconvex problem in graph domains, to which approximate solutions have been proposed in the  previous studies \cite{IsufiLPL19, Marques-Stationarity}.

Our work essentially differs from these previous approaches in that it aims to learn parametric ARMA graph process models by explicitly matching the process parameters to the joint time-vertex spectrum of the process.  As for the theoretical contributions, the study in  \cite{MeiM17} presents an analysis of the estimation error of the algorithm therein and reports similar convergence rates to ours; however, their analysis addresses AR process models instead of ARMA. Lastly, a preliminary version of our study has been presented in \cite{GuneyiCV21}. The current paper builds on \cite{GuneyiCV21} by significantly extending the experimental results and including a detailed theoretical analysis.


\section{Brief Overview of Graph Processes}
\label{sec_basics_gsp}

In this section, we give a brief overview of basic concepts related to the stochastic modeling of time-varying signals on graphs. Throughout the paper, matrices (e.g.~$\Amat$) and vectors (e.g.~$\avect$) are shown with boldface capital letters and boldface lowercase letters, respectively. The notation $\conj{\cdot}$ represents the complex conjugate of a complex number or matrix;  $(\cdot)^H$ denotes the Hermitian (transpose conjugate);  $(\cdot)^\intercal$ stands for the transpose; and $(\cdot)_{ij}$ denotes the $(i,j)$-th entry of a matrix. $\eye_N \in \R^{\graphDim \times \graphDim} $ represents the identity matrix. The operation $\otimes$ denotes the Kronecker product and $\text{tr}(\cdot)$ represents the trace of a matrix. The notation $\| \cdot \|$ stands for the $\ell_2$-norm for vectors and the operator norm for matrices,  $\| \cdot \|_F$ denotes the Frobenius norm, and $\| \cdot \|_1$ represents the $\ell_1$-norm. The expectation and the variance of a random variable are respectively denoted as $E[\cdot]$ and $\text{Var}(\cdot)$. The notation $\text{diag}(\cdot)$ represents the vector formed by extracting the entries on the diagonal of a matrix.

\subsection{Time-Vertex Signal Processing}
We consider an undirected weighted graph model $\graph = (\nodeSet,\edgeSet, \weightMat)$ consisting of a set of nodes $\nodeSet = \{\node_1,\node_2,\cdots,\node_N \}$, a set of undirected edges $\edgeSet$ that represents the connections between the nodes (vertices), and a weight matrix $\weightMat \in \R^{\graphDim \times \graphDim}$ representing the edge weights. The degree matrix $\degreeMat \in \R^{\graphDim \times \graphDim}$ is defined as a diagonal matrix with entries given by ${\degreeMat}_{\iNode\iNode}  = \sum_{\jNode=1}^{\graphDim} \weightMat_{i \jNode}$. The combinatorial graph Laplacian matrix is defined as $ \graphLaplacian = \degreeMat - \weightMat \in \R^{\graphDim \times \graphDim} $, while the symmetrically normalized graph Laplacian $ \degreeMat^{-1/2} (\degreeMat - \weightMat) \degreeMat^{-1/2} $ is also used commonly in graph signal processing.  The graph Laplacian has the eigenvalue decomposition $\graphLaplacian = \graphEigenvectorMat \graphEigenvalueMat_\graph \graphEigenvectorMat^\intercal
$, where the matrix $\graphEigenvectorMat$ forms a graph Fourier basis.


A graph signal  $\graphSignal : \nodeSet  \rightarrow \R$ is a mapping from the set of nodes to real numbers, which can alternatively be represented as a vector $\graphSignal \in \R^\graphDim $. The graph Fourier transform (GFT) $\hat{\graphSignal}$ of a graph signal $\graphSignal$ is defined as $\hat{\graphSignal} = \graphEigenvectorMat^\intercal  \, \graphSignal$ \cite{DavidEmerging}. The filtering operation on graphs is done as
\begin{equation}
	\graphSignalFilterOut = g(\graphLaplacian) \graphSignal = \graphEigenvectorMat \, g(\graphEigenvalueMat_\graph) \, \graphEigenvectorMat^\intercal \graphSignal
\end{equation}
where $\filterKernel : \{0\} \cup \R^+ \rightarrow\R$ denotes a filter kernel,  $\filterKernel (\graphLaplacian) \in \R^{\graphDim \times \graphDim} $ is a graph filter, $\graphSignal$ is the input signal, and $\graphSignalFilterOut$ is the output signal. Here  $\filterKernel (\graphEigenvalueMat_\graph)$ is a diagonal matrix with diagonals $\filterKernel (\graphEigenvalue_n)$, where $\graphEigenvalue_n $ are the eigenvalues of $\graphLaplacian $.

A time-varying graph signal, or a \textit{time-vertex} signal observed on a graph $\graph $ during the time instants $t=1, \dots, T$ can be represented as a matrix $\timeVertexSignalMat = \big[\graphSignal_1 \ \graphSignal_2 \ \cdots \  \graphSignal_T\big] \in \R^{\graphDim \times T}$. Here each column $\graphSignal_t \in \R^N$  of $\timeVertexSignalMat $  is a graph signal observed at time $t$, and each row of $\timeVertexSignalMat $ is a time signal observed on a graph node. The joint time-vertex frequency behavior of time-varying graph signals can be analyzed using the joint Fourier transform (JFT) \cite{Natheneal-Joint-Stationarity}. The JFT $\widehat{\timeVertexSignalMat}$ of a time-vertex signal $\timeVertexSignalMat $ is defined in such a way that it takes the GFT along the node dimension and the DFT along the time dimension as \cite{Natheneal-Joint-Stationarity}
\begin{equation}
	\widehat{\timeVertexSignalMat} = JFT\{\timeVertexSignalMat\} = \graphEigenvectorMat^\intercal \, \timeVertexSignalMat \, \conj{\timeEigenvectorMat}
\end{equation}
where $\timeEigenvectorMat $ is the normalized DFT matrix given by
\begin{equation}
\label{eq_defn_UT}
	\timeEigenvectorMat(t,\tau) = {e^{j\omega_\tau \iTime} \over \sqrt{T}}, \quad \omega_\tau = {{2\pi(\tau-1)} \over {T}} \:
	\text{ for } t, \tau = 1, \cdots,T.
\end{equation}
%
%
%
Denoting the vectorized form of a time-vertex signal $\timeVertexSignalMat \in \R^{\graphDim \times T}$ as $\xv \in \R^{\NT}$, the JFT can also be expressed as \cite{Natheneal-Joint-Stationarity}
\begin{equation}
	\hat{\xv} = \jointEigenvectorMat ^H \xv
\end{equation}
where $\jointEigenvectorMat = \timeEigenvectorMat \otimes \graphEigenvectorMat$. The filtering of time-vertex signals can similarly be defined in the joint spectral domain through the use of the joint Laplacian operator $\jointLaplacian = \timeLaplacian \otimes \eye_N + \eye_T \otimes \graphLaplacian$, where $\timeLaplacian $ is the Laplacian of a cyclic graph with eigenvector matrix $\timeEigenvectorMat $ \cite{Natheneal-Joint-Stationarity}. Note that this definition of the joint Laplacian $\jointLaplacian$ corresponds to the Laplacian of a Cartesian product graph, while other graph products such as the strong product and generalized products also exist in the literature \cite{JiangFTX21}. For a joint time-vertex filter, the relation between the input  and the output time-vertex signals $\timeVertexSignalMat $ and $\timeVertexFilterOutSignalMat $ can be represented in terms of their vectorized forms as
\begin{equation}
	\yv =  h(\graphLaplacian, \timeLaplacian) \, \xv = \jointEigenvectorMat \, h(\graphEigenvalueMat_\graph, \timeEigenvalueMat) \, \jointEigenvectorMat ^H \xv .
\end{equation} 
Here $h(\graphEigenvalueMat_\graph, \timeEigenvalueMat)$ is a diagonal matrix with $[h(\graphEigenvalueMat_\graph, \timeEigenvalueMat)]_{ii}=h(\graphEigenvalue_n, \omega_\tau)$ for $i=(\tau-1)N+n$, hence containing on its diagonals the joint filter kernel values $h(\graphEigenvalue_n, \omega_\tau)$  representing the desired filter response at graph frequency $\graphEigenvalue_n$ and time frequency $\omega_\tau$.

\subsection{Joint Time-Vertex Wide Sense Stationary Processes}
Let $\timeVertexSignalMat \in \R^{N\times T}$ be a random time-vertex process, with vectorized form $\xv \in \R^{\NT}$. If $\timeVertexSignalMat $ satisfies the following conditions, it is called a joint time-vertex wide sense stationary (JWSS) process \cite{Natheneal-Joint-Stationarity}:

\begin{itemize}
	\item $\xv$ has a constant mean  $E[\xv] = c \, \eye_{\NT}$.
	\item The covariance matrix $\covMat_{\xv}$  of the process $\xv$ is a joint time-vertex filter  
	\begin{equation}
	\label{eq_cov_jwss}
	 \covMat_{\xv} = h(\graphLaplacian, \timeLaplacian)= \jointEigenvectorMat \, h(\graphEigenvalueMat_\graph , \timeEigenvalueMat) \, \jointEigenvectorMat ^H.
	 \end{equation}
\end{itemize}

Hence, if $\xv$ is a  JWSS process, the covariance matrix  $\covMat_{\xv}$ of the process has the same eigenvector matrix $\jointEigenvectorMat $ as the joint Laplacian $\jointLaplacian$. Moreover, the eigenvalues $h(\graphEigenvalue_n, \omega_\tau)$ of  $\covMat_{\xv}$ give the joint power spectral density (JPSD) of the time-vertex process.

\subsection{Autoregressive Moving Average Graph Processes}

The concept of autoregressive moving average (ARMA) filters in classical signal processing has been extended to graph domains in several previous works \cite{IsufiLSL16,IsufiLPL19}. A JWSS time-vertex process $\timeVertexSignalMat $ can be modeled as an ARMA graph process if it is generated by filtering a zero-mean white process with an ARMA graph filter. We consider an input white process of normal distribution $\wt \sim \mathcal{N}(0, I_N)$ whose instances  $\wt $ at distinct time instants $\iTime$ are independent. The graph process $\graphSignal_t$ at each time $\iTime$ is then related to the past values $\graphSignal_{\iTime-p}$ of the process and the input process $\wt $ as \cite{IsufiLPL19}
\begin{equation}
	\graphSignal_t = -\sum_{p= 1}^Pa_p(\graphLaplacian) \graphSignal_{\iTime-p} + \sum_{q= 0}^Qb_q( \graphLaplacian) \wSignal_{\iTime-q} 
\end{equation}
where $a_p(\graphLaplacian)$, $b_q(\graphLaplacian)$ are graph filters. If $a_p(\graphLaplacian)$ and $b_q(\graphLaplacian)$  are polynomial filters of the form $a_p(\graphLaplacian)=  \sum_{k } a_{pk}\graphLaplacian^k$ and   $b_q(\graphLaplacian)= \sum _{m } b_{qm}\graphLaplacian^m$, where $\graphLaplacian^k$ represents the $k$-th power of the graph Laplacian,  then the graph ARMA process model becomes \cite{IsufiLSL16,IsufiLPL19}
\begin{equation}
	\label{ARMA polynomial model}
	\graphSignal_t = -\sum_{p= 1}^P \sum_{k = 0 }^K a_{pk}\graphLaplacian^k \, \graphSignal_{\iTime-p} + \sum_{q= 0}^Q \sum _{m = 0}^M b_{qm}\graphLaplacian^m \, \wSignal_{\iTime-q}.	
\end{equation}
Here $a_{pk}$ and $b_{qm}$ are the ARMA graph filter coefficients. Since the input process $\wt$ is assumed to be Gaussian, the filter output process $\graphSignal_t $ is also a Gaussian process. In our work, we consider \eqref{ARMA polynomial model} as our time-vertex process model. Taking the JFT of the process $\timeVertexSignalMat $, the time-vertex spectral domain representation of the graph filter in \eqref{ARMA polynomial model} can be obtained as \cite{IsufiLPL19} 

\begin{equation}
\label{eq_joint_spectrum_arma}
	H(\graphEigenvalue_n,\omega_\tau) = \frac{\sum_{q = 0} ^Q \sum_{m = 0} ^M b_{qm} \, \graphEigenvalue_n^m \, e^{-j\omega_\tau q}}{1 + \sum_{p = 1} ^P \sum_{k = 0} ^K a_{pk} \, \graphEigenvalue_n^k \, e^{-j\omega_\tau p}}.
\end{equation}

\section{Proposed Method for Learning Parametric Time-Vertex Processes}
\label{sec_prop_method}

In this work, we consider a setting where $L$ realizations $\{\timeVertexSignalMat^l \}_{l=1}^L$ of the time-vertex process $\timeVertexSignalMat $ are available. Each realization $\timeVertexSignalMat ^l \in \R^{\graphDim \times T}$ is assumed to be only partially observed, such that the value  $\timeVertexSignalMat ^l_{i\iTime}$ of the realization is known only at some of the graph nodes $i \in \{1, \dots, N\}$ for some of the time instances $t \in \{1, \dots, T \}$. Let $\indSet^l$ denote the index set of node-time pairs for which the realization $\timeVertexSignalMat ^l$ is observed.
\[
\indSet ^l = \{ (i, t) \ | \ \timeVertexSignalMat_{i\iTime}^l \text{ is observed} \} 
\]
Also, let $\bar \indSet ^l$ denote the complement of $\indSet ^l$, i.e., the index set for which the observation of the realization $\timeVertexSignalMat ^l$ is missing. We consider the problem of learning a process model that allows the estimation of the missing observations $\{ \timeVertexSignalMat^l_{i\iTime} | \ (i,t)\in \bar \indSet ^l, \ l=1, \dots, L \}$, given the available process observations $\{ \timeVertexSignalMat^l_{i\iTime} | \ (i,t)\in \indSet ^l, \ l=1, \dots, L \} $.

Our approach is based on first obtaining an initial rough estimate $	\tilde \covMat_{\xv} $ of the covariance matrix  $\covMat_{\xv} $ \eqref{eq_cov_jwss} from the available process observations, which yields a rough estimate  $\tilde h(\graphEigenvalue_n, \omega_\tau)$ of the joint power spectral density. We then learn the ARMA model parameters by fitting the joint time-vertex spectrum \eqref{eq_joint_spectrum_arma}  of the ARMA filter to the initial estimate $\tilde h(\graphEigenvalue_n, \omega_\tau)$ of the JPSD. An improved estimate of the covariance matrix $\covMat_{\xv} $ is finally obtained from the learnt ARMA model, from which we infer the initially unknown process values. We discuss these steps in detail in the following sections.

\subsection{Initial Estimation of the JPSD}
\label{ssec_jpsd_estimate}

We first describe the initial estimation of the JPSD, which will be used in the computation of ARMA process models in Section \ref{ssec_learn_arma}. We compute the initial JPSD by employing a variant of the algorithm proposed in \cite{Natheneal-Joint-Stationarity}. The covariance matrix $\covMat_{\xv}$ of a zero-mean time-vertex process $\timeVertexSignalMat $ is given by 
\begin{equation}
\label{eq_sigmax_defn}
\covMat_{\xv} = E[ \xv \xv^\intercal] =
\begin{bmatrix}
\covMat_{1,1} & \covMat_{1,2} & \cdots & \covMat_{1,T} \\
\covMat_{2,1} & \covMat_{2,2} & \cdots & \covMat_{2,T} \\
\vdots  & \vdots  & \ddots & \vdots  \\
\covMat_{T,1} & \covMat_{T,2} & \cdots & \covMat_{T,T} 
\end{bmatrix}
\end{equation}
where $\covMat_{\iTime,u}=E[\graphSignal_t \, \graphSignal_u^\intercal]$ stands for the covariance matrix of the values of the process at time instants $\iTime$ and $u$. When $\timeVertexSignalMat $ is a JWSS process, the covariance matrix $\covMat_{\xv}$  is known to have the following special property, which simplifies its estimation: Each covariance matrix $\covMat_{\iTime,u}$ is a graph filter $\covMat_{\iTime,u}=g_{\iTime,u}(\graphLaplacian) $, which depends only on the time difference $t-u$. This leads to a block-Toeplitz structure in $\covMat_{\xv}$ \cite{Natheneal-Joint-Stationarity}. The observation that any graph filter $\filterKernel (\graphLaplacian)$ needs to be symmetric, as well as the overall covariance matrix $\covMat_{\xv}$, leads to the equality $\covMat_{\iTime,u}= \covMat_{u,\iTime}$. Hence, the estimation of $\covMat_{\xv}$ boils down to the estimation of the smaller matrices $\covMat_{\Delta}\in \R^{\graphDim \times \graphDim} $, for $\Delta=0, 1, \dots, T-1$, where $\covMat_{\iTime,u} = \covMat_{\Delta}$ with $\Delta = | t-u |$. We obtain an estimate $\tilde \covMat_{\Delta}$  of each $\covMat_{\Delta}$ by estimating its entries $[ \tilde \covMat_{\Delta}]_{ij}$ from the sample covariance of the available process observations as
\begin{equation}
[ \tilde \covMat_{\Delta}]_{ij}  = \frac{1}{| \indTriSet_{i,j}^\Delta |} 
\sum_{(t,u,l) \in \indTriSet_{i,j}^\Delta } 
\timeVertexSignalMat ^l_{i t} \timeVertexSignalMat ^l_{ju}
\end{equation}
where 
\[
\indTriSet_{i,j}^\Delta = \{ (t,u,l) \ | \ (i,t) \in \indSet^l , (j,u) \in \indSet^l, \ |t-u| = \Delta , \  1 \leq l \leq  L \}
\]
denotes the set of time and realization indices of available observations\footnote{In applications where the process is permanently unobserved at some graph nodes, e.g. as in sensor networks with permanently malfunctioning sensors, this leads to missing rows and columns in the covariance estimate. One can interpolate the missing covariance values, e.g.,~based on the neighbors of the permanently unobserved nodes.} and $| \indTriSet_{i,j}^\Delta |$ stands for the cardinality of $\indTriSet_{i,j}^\Delta$.
 
Once we compute the estimate $\tilde \covMat_{\xv}$ of  the covariance matrix  $\covMat_{\xv}$, using the relation in \eqref{eq_cov_jwss}, we obtain the initial estimate $\tilde h (\graphEigenvalue_n , \omega_\tau)$ of the JPSD simply by extracting the diagonal entries of the matrix
\begin{equation}
\label{eq_jpsd_est}
\tilde h(\graphEigenvalueMat_{\graph}, \timeEigenvalueMat) = \jointEigenvectorMat ^H \tilde \covMat_{\xv} \jointEigenvectorMat .
\end{equation}

\subsection{Computation of the ARMA Graph Process Model}
\label{ssec_learn_arma}

We now propose our problem formulation for learning an ARMA process model coherent with the initially estimated JPSD. We first rewrite the filter spectrum in \eqref{eq_joint_spectrum_arma} as
\begin{equation}
	H(\graphEigenvalue_n,\omega_\tau) = \frac{\bvect^H \uvect_{n,\tau}}{1 + \avect^H \vvect_{n,\tau} }
\end{equation} 
where the vectors $\avect \in \R ^{P(K+1) \times 1} $ and $\bvect \in \R ^{(Q+1)(M+1) \times 1}$ respectively consist of the filter coefficients $a_{pk}$  and  $b_{qm}$ as
\begin{equation}
\label{eq_defn_ab1}
\begin{split}
\avect &= [a_{10} \ a_{11} \ \cdots a_{pk}  \ \cdots a_{PK} ]^H  \\
\bvect &= [b_{00} \ b_{01} \ \cdots b_{qm} \ \cdots b_{QM} ]^H.
\end{split}
\end{equation}
The vectors $\vvect_{n,\tau} \in \mathbb{C} ^{P(K+1) \times 1}$ and $\uvect_{n,\tau} \in \mathbb{C} ^{(Q+1)(M+1) \times 1} $ consist of the constant coefficients 
\begin{equation}
\label{eq_defn_uv1}
\begin{split}
\vvect_{n,\tau} &= [\graphEigenvalue_n^0 e^{j\omega_\tau 1} \ \graphEigenvalue_n^1 e^{j\omega_\tau 1} \ \cdots \ \graphEigenvalue_n^ke^{j\omega_\tau p} \ \cdots  \ \graphEigenvalue_n^Ke^{j\omega_\tau P}]^H  \\
\uvect_{n,\tau} &= [\graphEigenvalue_n^0 e^{j\omega_\tau 0} \ \graphEigenvalue_n^1 e^{j\omega_\tau 0} \ \cdots \ \graphEigenvalue_n^me^{j\omega_\tau q} \  \cdots \ \graphEigenvalue_n^Me^{j\omega_\tau Q}]^H 
\end{split}
\end{equation}
where $\graphEigenvalue_n^k$ denotes the $k$-th power of the $n$-th graph eigenvalue  $\graphEigenvalue_n$, and the frequency variables $\omega_\tau$ are as defined in \eqref{eq_defn_UT}.

Similarly to the filtering of white noise processes in classical signal processing, the JPSD $h(\graphEigenvalue_n, \omega_\tau)$ of the process is related to the filter spectrum in \eqref{eq_joint_spectrum_arma} as \cite{Natheneal-Joint-Stationarity} 
\begin{equation}
\label{eq_rel_h_ab}
h(\graphEigenvalue_n, \omega_\tau)  =  | H(\graphEigenvalue_n, \omega_\tau) |^2
= \left | \frac{\bvect^H \uvect_{n,\tau}}{1 + \avect^H \vvect_{n,\tau} } \right |^2.
\end{equation}
We then formulate the estimation of the ARMA model from the initially estimated JPSD $\tilde h(\graphEigenvalue_n,\omega_\tau)$ as 
\begin{equation}
\label{eq_opt_prob1}
	\min_{\avect,\bvect} \sum_{n=1}^N \sum_{\tau=1}^T \left | \ \ \left|  \frac{\bvect^H \uvect_{n,\tau}}{1 + \avect^H \vvect_{n,\tau} } \right|^2 - \tilde h(\graphEigenvalue_n,\omega_\tau) \right|^2.	
\end{equation}
Due to the fourth-order dependence of the objective function on the model parameters $\avect$ and $\bvect$, the optimization problem in \eqref{eq_opt_prob1} is nonconvex with non-unique minima, and hence difficult to solve. In order to develop a convex relaxation of this problem, we first reformulate the relation in \eqref{ARMA polynomial model} as
\begin{equation}
	\label{modified_arma}
	\sum_{p= 0}^P \sum_{k = 0 }^K a_{pk}\graphLaplacian^k \graphSignal_{\iTime-p}  = \sum_{q= 0}^Q \sum _{m = 0}^M b_{qm}\graphLaplacian^m \wSignal_{\iTime-q} 	
\end{equation}
where we set $a_{00} = 1 $ and $a_{0k} = 0$ for $k = 1,2, \cdots, K$. This new formulation has the advantage that the JPSD of the process has the relatively simple form
\begin{equation}
	h(\graphEigenvalue_n,\omega_\tau) = \left| \frac{\bvect^H \uvect_{n,\tau}}{\avectaug^H \vvectaug_{n,\tau} } \right|^2
\end{equation} 
where we define the vectors $\avectaug \in \R ^{(P+1)(K+1) \times 1} $ and $\vvectaug_{n,\tau} \in \mathbb{C} ^{(P+1)(K+1) \times 1}$ to be augmented versions of $\avect$ and $\vvect_{n,\tau}$ as
\begin{equation}
\label{eq_newdefn_ab}
\begin{split}
	\avectaug &= [a_{00} \ a_{01} \ \cdots a_{pk} \ \cdots a_{PK} ]^H  \\
	\vvectaug_{n,\tau} &= [\graphEigenvalue_n^0 e^{j\omega_\tau 0} \ \graphEigenvalue_n^1 e^{j\omega_\tau 0} \ \cdots  \
	\graphEigenvalue_n^k e^{j\omega_\tau p}  \ \cdots \ \graphEigenvalue_n^K e^{j\omega_\tau P}]^H. 
\end{split}	
\end{equation}
%
%
In addition to this change of variables, we also remove the term in the denominator and hence propose to substitute the objective in \eqref{eq_opt_prob1} with
\begin{equation}
\label{eq_opt_ab_v3}
\begin{split}
	 \sum_{n=1}^N \sum_{\tau=1}^T \left |  {\uvect_{n,\tau}^H \bvect \bvect^H \uvect_{n,\tau}} - { \vvectaug_{n,\tau}^H \avectaug \avectaug^H \vvectaug_{n,\tau} } \  \tilde h(\graphEigenvalue_n,\omega_\tau) \right|^2.	
\end{split}
\end{equation}
The objective function \eqref{eq_opt_ab_v3} is a proxy for the one in \eqref{eq_opt_prob1} and results in a different solution in general. Nevertheless, provided that the spectrum has bounded magnitude, i.e., the denominator term in \eqref{eq_opt_prob1} does not take arbitrarily small values, the minimization of \eqref{eq_opt_ab_v3} is likely to provide a satisfactory estimate for the model that we aim to fit in \eqref{eq_opt_prob1}. The dependence of the objective \eqref{eq_opt_ab_v3} on the vectors $\avectaug$ and $\bvect$ is still nonconvex. We thus propose to relax it into a convex function of the matrices $\Amat$ and $\Bmat$, defined as $\Amat \triangleq \avectaug \avectaug^H$ and $\Bmat \triangleq \bvect \bvect^H$. However, for these definitions to be valid, the matrices $\Amat$ and $\Bmat$ must be rank-1 and positive semidefinite. Hence, we get the optimization problem
\begin{equation}
\label{eq_opt_AB_v4}
\begin{split}
	 \min_{\Amat, \Bmat} \sum_{n=1}^N \sum_{\tau=1}^T  & \mu(\lambda_n,\omega_\tau)  \left |   {\uvect_{n,\tau}^H \Bmat \, \uvect_{n,\tau}} - { \vvectaug_{n,\tau}^H \Amat  \, \vvectaug_{n,\tau} }  \ \tilde h(\graphEigenvalue_n,\omega_\tau) \right|^2 \\
	 \text{ subject to } \quad	&\text{rank}(\Amat)=1, \ \text{rank}(\Bmat)=1, \\
	& \Amat \in \psdCone_+^{(P+1)(K+1)},  \ \ \Bmat \in \psdCone_+^{(Q+1)(M+1)}, \\
	& a_{00} = 1, \ \ a_{0k} = 0 \text{ for } k = 1, 2, \ \cdots ,  \ K
\end{split}
\end{equation}
where $\psdCone_+^{R}$ denotes the cone of $R\times R$ positive semidefinite matrices. Here $\mu(\cdot, \cdot)$ stands for an optional weight function for adaptively penalizing the error at particular zones of the joint spectrum, which can be chosen as $\mu(\lambda_n,\omega_\tau)=1$ under no priors.   Lastly, we apply a convex relaxation of the rank constraints as follows. The positive semidefinite matrices $\Amat$ and $\Bmat$ can be pushed to be low-rank by minimizing the sums of their singular values, or equivalently, their traces $\text{tr}(\Amat)$ and $\text{tr}(\Bmat)$. We hence obtain our final optimization problem as 
\begin{equation}
\label{eq_opt_AB_v5}
\begin{split}
	& \min_{\Amat, \Bmat} \sum_{n=1}^N \sum_{\tau=1}^T  \mu(\lambda_n,\omega_\tau)  \left |   {\uvect_{n,\tau}^H \Bmat \, \uvect_{n,\tau}} - { \vvectaug_{n,\tau}^H \Amat  \, \vvectaug_{n,\tau} }  \ \tilde h(\graphEigenvalue_n,\omega_\tau) \right|^2  \\
	& + \mu_A \text{tr}(\Amat) 	 
	 + \mu_B \text{tr}(\Bmat), \quad
	 \text{ subject to }  \ \ \ \Amat \in \psdCone_+^{(P+1)(K+1)},  \\ 
	 & \Bmat \in \psdCone_ +^{(Q+1)(M+1)},  \ \ 
	 a_{00} = 1, \ \ a_{0k} = 0 \text{ for } k = 1, 2, \cdots,   K
\end{split}
\end{equation}
where $\mu_A$ and $\mu_B$ are positive weight parameters. The objective function in  \eqref{eq_opt_AB_v5} is quadratic and jointly convex in $\Amat$ and $\Bmat$. We also observe that the constraint set consists of linear equality constraints and the constraint that $\Amat $ and $\Bmat $ be positive semidefinite matrices. Hence, \eqref{eq_opt_AB_v5} is a convex problem that can be solved using convex optimization techniques \cite{cvx}, \cite{gb08} relying on semidefinite quadratic linear programming \cite{optimization-SDPT3,optimization-quadratic-linear}. Once the matrices $\Amat $ and $\Bmat $ are computed by solving  \eqref{eq_opt_AB_v5}, the ARMA model parameter vectors $\avect$ and $\bvect$ can be recovered through rank-1 decompositions of $\Amat $ and $\Bmat $.  

The final convex problem formulation in \eqref{eq_opt_AB_v5} has the clear advantage that its global minimum can be computed; however, this comes at the expense of a likely deviation between the solution of  \eqref{eq_opt_AB_v5} and that of the original problem \eqref{eq_opt_prob1} due the various relaxations and approximations done along the way. In fact, we study this trade-off through several experiments in Section \ref{sec_exp_res}. In realistic settings where signals deviate from the underlying process model due to noise, the convex relaxation of the problem in  \eqref{eq_opt_AB_v5}  improves the overall model estimation accuracy significantly compared to attempting to solve the original nonconvex problem \eqref{eq_opt_prob1} via, e.g.,~local optimization techniques.

\subsection{Estimation of Missing Observations of the Process}
\label{ssec_lmmse_estimate}

Having estimated the ARMA model parameters $\avect$ and $\bvect$ as described in Section \ref{ssec_learn_arma}, we now discuss the estimation of the missing observations $\{ \timeVertexSignalMat ^l_{i\iTime} | \ (i,t)\in \bar \indSet^l \}$ of the process. Following the relation in \eqref{eq_rel_h_ab}, the learnt model parameters $\avect$ and $\bvect$ provide an improved estimate of the JPSD, which we may denote as $\hest (\graphEigenvalue_n, \omega_\tau)$. Rearranging $\hest(\graphEigenvalue_n, \omega_\tau)$ in matrix form as $\hest(  \graphEigenvalueMat_{\graph}, \timeEigenvalueMat )$, we can obtain an improved estimate $\covMat^\est_{\xv}$ of the covariance matrix $\covMat_{\xv}$ as 
\begin{equation}
\label{eq_sigma_from_jpsd}
\covMat^\est_{\xv} =  \jointEigenvectorMat  \, \hest(  \graphEigenvalueMat_{\graph}, \timeEigenvalueMat ) \, \jointEigenvectorMat ^H
\end{equation} 
which follows from the relation in \eqref{eq_cov_jwss}.

Finally, denoting the vectorized form of each realization $\timeVertexSignalMat ^l$ of the time-vertex process as $\bar \graphSignal ^l$, let us form two new vectors $\bar \ySignal^l$ and $\bar \zSignal^l$, consisting respectively of the known and the missing entries of $\bar \xSignal^l$, i.e., the process values in the sets  $\{ \timeVertexSignalMat ^l_{i\iTime} | \ (i,t)\in \indSet^l \}$ and $\{ \timeVertexSignalMat ^l_{i\iTime} | \ (i,t)\in \bar \indSet^l \}$. The vector of missing process values $\bar \zSignal^l$ for $l=1, \dots, L$ can then be estimated as follows with the classical minimum mean square error (MMSE) estimation approach, which is the same as the linear MMSE estimate since $\bar \ySignal^l$ and $\bar \zSignal^l$ are jointly Gaussian \cite{Natheneal-Joint-Stationarity}
\begin{equation} 
\label{eq_lmmse_est_z}
	({\bar \zSignal}^l)^\est = (\covMat^l_{\bar \zSignal \bar \ySignal})^\est \, ((\covMat ^l_{\bar \ySignal})^\est)^{-1} \, \bar \ySignal^l .
\end{equation}
Here $(\covMat ^l_{\bar \zSignal \bar \ySignal})^\est$ and $(\covMat ^l_{\bar \ySignal})^\est$  respectively denote  the estimates of the cross-covariance matrix of $\bar \zSignal^l$ and  $\bar \ySignal^l$, and the covariance matrix of $\bar \ySignal^l$. These matrices can be formed by extracting the corresponding entries of $\covMat^\est_{\xv} $ for each realization $\timeVertexSignalMat ^l$.

We call the proposed method for learning graph ARMA processes from joint spectra as JS-ARMA, and give its summary in Algorithm \ref{alg_tv_process}.

\begin{algorithm}[h]
\footnotesize
\caption{Proposed JS-ARMA Method }
\begin{algorithmic}[1]

\STATE
\label{algInput}
\textbf{Input: } Graph $\graph$, available process observations $\{ \bar \ySignal^l \}_{l=1}^L$\\

\STATE
Compute $\tilde h$ from $\{ \bar \ySignal^l \}$ as explained in Section \ref{ssec_jpsd_estimate}

\STATE
Compute $\Amat$ and $\Bmat$ by solving the optimization problem \eqref{eq_opt_AB_v5}
	
\STATE
Find $\avect$ and $\bvect$ through rank-1 decompositions of $\Amat$ and $\Bmat$

\STATE
Compute the JPSD $h^\est(\graphEigenvalue_n, \omega_\tau)$ from $\avect$ and $\bvect$ using \eqref{eq_rel_h_ab}
	
\STATE
Find the covariance matrix $\covMat^\est_{\xv} $ from the JPSD using the relation \eqref{eq_sigma_from_jpsd}

\STATE
Find MMSE estimates $\{ (\bar \zSignal^l)^\est \}_{l=1}^L$ of observations using \eqref{eq_lmmse_est_z}

\STATE
\label{algOutput}
\textbf{Output: } 
Estimated process observations $\{ (\bar \zSignal^l)^\est \}_{l=1}^L$\\

\end{algorithmic}
\label{alg_tv_process}
\end{algorithm}
\normalsize
\vspace{-0.3cm}

\subsection{Complexity Analysis of the Algorithm}
\label{ssec_complexity_anlys}

Here we analyze the computational complexity of the proposed JS-ARMA method. First,  we study the computations required for Step-2 of Algorithm \ref{alg_tv_process}. Assuming that the graph Laplacian $\graphLaplacian$ is known, the time complexities of computing  $\graphEigenvectorMat$ and $\graphEigenvalueMat_\graph$ is of $O(N^3)$. The joint Fourier transform matrix $\jointEigenvectorMat $ is found via the Kronecker product of $\graphEigenvectorMat$ and $\timeEigenvectorMat $ with a complexity of $O(N^2T^2)$. Since $\tilde\covMat_{\xv}$ is a block-Toeplitz covariance matrix, it is computed with a complexity of $O(N^2TL)$. Finally, the initial JPSD estimate $\tilde h$ is obtained from \eqref{eq_jpsd_est} with a complexity of $O(N^3T^3)$. Next, in Step-3, the optimization problem \eqref{eq_opt_AB_v5} can be solved with semidefinite quadratic linear programming, via e.g., the HKM algorithm \cite{optimization-SDPT3,optimization-quadratic-linear}. The objective function in \eqref{eq_opt_AB_v5} can be implemented by grouping together the constant terms through $O(NT)$ operations only once before calling the HKM algorithm, in which case the number of variables and the number of equality constraints in HKM become independent of $N$ and $T$. Assuming that $P, K, Q, M \ll NT$, the complexity of Step-3 is then obtained as $O(NT)$. In Step-4, the complexities of the rank-1 decompositions of the matrices $\Amat$ and $\Bmat$ are of $O((P+1)^3(K+1)^3)$ and $O((Q+1)^3(M+1)^3)$ respectively. Then, in Step-5,  $ h^\est(\graphEigenvalue_n, \omega_\tau)$ can be found from \eqref{eq_rel_h_ab} with a complexity of $O(NT(PK+QM))$. The covariance matrix in Step-6 can be computed using \eqref{eq_sigma_from_jpsd} with $O(N^3T^3)$ operations, and lastly, the 
complexity of finding the MMSE estimates via \eqref{eq_lmmse_est_z} in Step-7 is of $O(N^3T^3)$. Hence, assuming $P, K, Q, M \ll NT$, the overall complexity of our method can be summarized as $O(N^3 T^3)$.


\section{Sample Complexity Analysis of Learning Graph ARMA Models}
\label{sec_samp_complx}

In this section we theoretically analyze the sample complexity of learning graph ARMA models. We consider a time-vertex process $\xv$ conforming to the ARMA model \eqref{ARMA polynomial model}. We denote as $\ag $ and $\bg$ the true but unknown parameter vectors generating the process as defined in \eqref{eq_defn_ab1}. Let us represent the overall true parameter vector as $\cg=[(\ag)^H \ (\bg)^H]^H $. We consider that the initial estimate of the process covariance matrix is obtained from the sample covariance of fully observed $L$ independent realizations $\{\xv^l \}_{l=1}^L$ of the process as
\begin{equation*}
\tilde \covMat_{\xv} = \frac{1}{L} \sum_{l=1}^L \xv^l (\xv^l)^\intercal.
\end{equation*}
The initial covariance estimate $\tilde \covMat_{\xv}$ gives the initial JPSD estimate $\tilde h(\graphEigenvalueMat_{\graph}, \timeEigenvalueMat) $ via  \eqref{eq_jpsd_est}. In order to make the derivations tractable, we base our analysis on the original form \eqref{eq_opt_prob1} of the objective function. We recall that the original problem \eqref{eq_opt_prob1} we would normally like to solve in this paper is nonconvex and impractical to tackle, hence the algorithm proposed in Section \ref{ssec_learn_arma} relies on developing a convex approximation for it. While it is also important to understand the effect of the convex relaxations made in \eqref{eq_opt_AB_v5}, we leave this issue to the experimental analyses in Section \ref{sec_exp_res} and focus here on how the quality of the solution of the original problem evolves with the number of realizations $L$ and the model complexity. Hence, denoting as 
\begin{equation}
\label{eq_defn_h_n_tau}
h_{\cvect}(\graphEigenvalue_n, \omega_\tau) 
= \left | \frac{\bvect^H \uvect_{n,\tau}}{1 + \avect^H \vvect_{n,\tau} } \right |^2
\end{equation}
the JPSD associated with an arbitrary process parameter vector $\cvect \triangleq [\avect^H \ \bvect^H]^H \in \R^\dc$ where $\dc \triangleq P(K+1)+(Q+1)(M+1)$ denotes the model order, we consider the ARMA process parameter vector
\begin{equation}
\label{eq_defn_ab_star}
	\cs = \arg \min_{\cvect \in  \csp} \sum_{n=1}^N \sum_{\tau=1}^T
	 \left | \ h_{\cvect}(\graphEigenvalue_n, \omega_\tau)   - \tilde h(\graphEigenvalue_n,\omega_\tau) \right|^2	
\end{equation}
that best matches the initial JPSD estimate $\tilde h(\graphEigenvalue_n,\omega_\tau)$ as formulated in \eqref{eq_opt_prob1}. Here $\csp \subset \R^\dc$ is assumed to be a compact set of feasible parameter vectors, with boundary, and excluding degenerate $\avect$ and $\bvect$ vectors of zero norm. The JPSD estimate given by the learnt graph ARMA model is thus $h_\cs(\graphEigenvalue_n,\omega_\tau)$. Our purpose in this section is then to characterize the deviation between the learnt JPSD $h_\cs(\graphEigenvalue_n,\omega_\tau)$ and the true JPSD $h_\cg(\graphEigenvalue_n,\omega_\tau) $.


Let us simply denote the vectorized form of the JPSD for the parameter vector $\cvect$ as 
\[  
\hvect_\cvect \triangleq 
[ h_ \cvect(\graphEigenvalue_1, \omega_1) 
\ h_ \cvect(\graphEigenvalue_2, \omega_1)
\ \dots
\ h_ \cvect(\graphEigenvalue_N, \omega_T)
]^\intercal
\in \R^{\NT}.
\]
Let us also similarly denote the vectorized form of the initial JPSD estimate  $\tilde h(\graphEigenvalue_n,\omega_\tau)$ as $\hinit \in \R^\NT$. Before presenting our results, we make the following mild assumptions: 
\begin{assumption}
\label{assum_tan_bound}
Let $\Hc \triangleq \{ \hvect_\cvect : \cvect \in \csp \} \subset \R^\NT$ denote the JPSD manifold parameterized by $\cvect$. Then, there exists a positive constant $\Tbnd >0$ such that
\begin{equation}
\label{eq_defn_Tbnd}
\left \|   \frac{d \, \hvect_{\cvect + t \unitvect}}{dt}  \bigg |_{t=0}  \right \| \geq \Tbnd
\end{equation}
for all $\cvect \in \csp$ and all $\unitvect \in \R^\dc$ of unit norm $\| \uvect \|=1$. 
\end{assumption}
In Assumption \ref{assum_tan_bound}, $t \in \R$, and the vector $\frac{d \, \hvect_{\cvect + t \unitvect}}{dt} \in \R^\NT$ represents a tangent to the manifold $\Hc $, which consists of the derivatives of the entries of $\hvect_{\cvect + t \unitvect}$. The constant $\Tbnd $ thus stands for a lower bound on the tangent norms of the manifold  $\Hc $, thus imposing $\Hc $ to have a non-degenerate geometry free of zero tangents. In Appendix A, the existence of the constant $\Tbnd $ is further analyzed, suggesting that if the length of the process is sufficiently large with respect to the number of process parameters so as to satisfy $\NT > \dc = P(K+1)+(Q+1)(M+1)$, then $\Tbnd$ is very likely to exist. 
%


\begin{assumption}
\label{assum_curv_bnd}
The JPSD of the process is finite over $\csp$.
\end{assumption}
In Appendix A, we also show that if Assumption \ref{assum_curv_bnd} holds, then 
for each $(n, \tau)$ pair, there exists a positive constant $\Kbnd_{n,\tau}$ such that
\begin{equation}
\label{eq_defn_Kbnd_ntau}
 \left |   \frac{d^2 }{dt^2} \  \hvect_{\cvect + t \unitvect}(\graphEigenvalue_n,\omega_\tau)    \right | \leq  \Kbnd_{n,\tau}
\end{equation}
for any $\cvect \in \csp$, any $\unitvect \in \R^\dc$  with $\| \unitvect \|=1$, and any $t$ with $\cvect + t \unitvect \in \csp$. One can then define a geometric constant
\begin{equation}
\label{eq_defn_Kbnd}
\Kbnd= \left(  \sum_{n=1}^\graphDim \sum_{\tau=1}^\timeDim  \Kbnd_{n,\tau}^2 \right)^{1/2}
\end{equation}
which can be regarded as a global upper bound on the curvature of the manifold $\Hc $. 

We first study in the following lemma the deviation between the true JPSD $\hg$ of the process and its estimate $\hs$ obtained by solving \eqref{eq_defn_ab_star}.

\begin{lemma}
\label{lem_hard_bnd_hs_hg}
Let $\eh \triangleq \hinit - \hg$ denote the error vector representing the deviation of the initial (sample covariance) JPSD estimate $\hinit$ from the true JPSD $\hg$. Then the JPSD estimation error of an algorithm solving \eqref{eq_defn_ab_star} can be bounded as
\begin{equation}
\label{eq_jspd_error_lemma}
\begin{split}
\|  \hs - \hg \| & \leq  
 \left (
 \left( \| \eh \| + \frac{\Kbnd}{2} \, \| \cs - \cg \|^2 \right)^2 
 - \| \hinit - \hs \|^2
\right )^{1/2}
 \\
 +
& \frac{\Kbnd}{2 \Tbnd^2}
\left (
 \left( \| \eh \| + \frac{\Kbnd}{2} \, \| \cs - \cg \|^2 \right)^2 
 - \| \hinit - \hs \|^2
 \right ).
\end{split}
\end{equation}
\end{lemma}

The proof of the lemma is given in Appendix B. Lemma \ref{lem_hard_bnd_hs_hg} intuitively states the following: The optimization problem \eqref{eq_defn_ab_star} seeks for the projection $\hs$ of the initial JPSD estimate $\hinit$ on the JPSD manifold $\Hc$. Assuming that the curvature $\Kbnd$ of the manifold is sufficiently small, this projection operation improves the JPSD estimation performance, since the error $\| \hs - \hg \|$ remaining after the projection operation reduces the initial error $\| \eh \|= \| \hinit - \hg \|$ by an amount proportional to the distance $ \| \hinit - \hs \|$ of $\hinit$ to the manifold $\Hc$. Consequently, the graph ARMA model learnt by solving \eqref{eq_defn_ab_star} will perform better than the initial JPSD estimate $\hinit $ relying on the sample covariance matrix of the process.

We are now ready to state our first main result on the sample complexity of learning  graph ARMA models.

\begin{theorem}
\label{thm_jpsd_error_conv_rate}
Consider a graph ARMA model learnt by solving \eqref{eq_defn_ab_star}. Then as the number of realizations $L$ increases, with probability at least $1-\delta$, the estimation errors of the model parameters and the JPSD decrease at the following rates
\begin{equation}
\begin{split}
 \| \cs - \cg \| &= O \left(\sqrt{ \frac{ N^2 T^2 \dc^4}{L\delta}  } \right) \\
\| \hs - \hg \| &= O \left( \sqrt{ \frac{N^2 T^2 \dc^4 }{L \delta}} \right).
\end{split}
\end{equation}
\end{theorem}

The proof of Theorem \ref{thm_jpsd_error_conv_rate} is given in Appendix C. In the proof, we first examine the convergence rate of the initial error $\eh$. We then build on Lemma \ref{lem_hard_bnd_hs_hg} to derive the rates of convergence of first the model parameter estimation error $\| \cs - \cg\|$, and then the JPSD estimation error $\| \hs - \hg \|$.  Theorem \ref{thm_jpsd_error_conv_rate} states that the estimation error of the JPSD of the process converges at rate $O(1/\sqrt{L})$ with $L$, confirming that the reliability of the estimate improves as the number of realizations $L$ increases. The estimation error scales with $N$ and $T$ as expected, due to the growth in the dimension of $\hvect_{\cvect}$ with process dimensions $N$ and $T$.  Meanwhile, the JPSD estimation error also depends on the model order $d$, as bounded by the quadratic rate $O(d^2)$. Hence, as $d$ increases, the number of realizations $L$ must also increase so as to measure up to the model order. 

One may wonder about the implications of Theorem \ref{thm_jpsd_error_conv_rate} for the estimation of the observations of the process. In line with the setting in Section \ref{ssec_lmmse_estimate}, let us consider a partially observed test realization $\bar \xSignal^\test \in \R^{\NT}$  of the time-vertex process $\xv$; and denote as  $\bar \ySignal^ \test $ and $\bar \zSignal^ \test $ its components respectively with known and missing entries. Let  $\covMat_{\xv}$ and $\covMat^\est_{\xv} $ stand for the true covariance matrix of the process and its estimate given by the JPSD $\hs $ learnt by solving \eqref{eq_defn_ab_star}.  Similarly, let $\covMat^ \test_{ \bar \ySignal}$, $ \covMat^ \test_{\bar \zSignal \bar \ySignal}$ and $(\covMat^ \test_{ \bar \ySignal})^\est$, $ (\covMat^\test_{\bar \zSignal \bar \ySignal})^\est$  denote the covariance and cross-covariance matrices obtained by extracting the submatrices of  $\covMat_{\xv}$ and $\covMat^\est_{\xv} $ corresponding to the known and the missing parts of a given test realization $\bar \xSignal^\test$. In the following main result, we study the rate of convergence of the estimate of $\bar \zSignal^ \test $ based on the MMSE estimation scheme in \eqref{eq_lmmse_est_z}.

\begin{theorem}
\label{thm_lmmse_error_conv_rate}
For a given test realization  $\bar \xSignal^\test$ with observed component $\ytest$, assume that $\covMat^ \test_{ \bar y}$, $(\covMat^ \test_{ \bar y})^\est$ and their difference are invertible. Let
\begin{equation*}
\begin{split}
\ztests &= \Sigzytesth \, (\Sigytesth)^{-1}  \ytest \\
\end{split}
\end{equation*}
denote the MMSE estimate of the missing component  $\bar \zSignal^ \test $ obtained through the covariance matrix $\covMat^\est_{\xv} $ estimated by solving  
\eqref{eq_defn_ab_star}; and let
\begin{equation*}
\begin{split}
\ztestg &=\Sigzytest \, (\Sigytest)^{-1} \ytest
\end{split}
\end{equation*}
denote the oracle estimate given by the true covariance matrix  $\covMat_{\xv}$ of the process. Then, as the number of realizations $L$ increases, $\ztests $ converges to $\ztestg$; such that with probability at least $1-\delta$, the deviation between them decreases at rate
\[
\| \ztests - \ztestg \| = O \left( \sqrt{\frac{NT \dc^2}{L \delta}} \right).
\]
\end{theorem}

The proof of Theorem \ref{thm_lmmse_error_conv_rate} is given in Appendix D, where the result in Theorem \ref{thm_jpsd_error_conv_rate} is used to bound the deviation between the MMSE estimates $\ztests $ and $\ztestg$. Theorem \ref{thm_lmmse_error_conv_rate} states that as the number of realizations increases, the estimate of the missing process observations obtained with the proposed formulation improve progressively, thus converging to the reference ideal estimate $\ztestg $ one would have if the process covariance matrix was perfectly known. In particular, the   estimate $\ztests $ with the learnt model converges to the ideal estimate  $\ztestg$ at rate $O(\sqrt{NT \dc^2 /L\delta})$, whose dependence on $L$ is the same as that of the JPSD convergence rate presented in Theorem \ref{thm_jpsd_error_conv_rate}.

\section{EXPERIMENTAL RESULTS}
\label{sec_exp_res}

In this section, we evaluate the performance of our method on real and synthetic time-vertex data sets.

\subsection{Performance and Sensitivity Analysis of JS-ARMA}
\label{sec_exp_sensit_analy}

Here we analyze the sensitivity of the proposed JS-ARMA method to factors such as number of realizations, noise, algorithm parameters and evaluate its complexity through the following experiments.

\subsubsection{Model estimation accuracy} 
\begin{figure}[t]
\begin{center}
     \subfloat[JPSD $h(\lambda_n, \omega_\tau)$]
     {\label{fig_jpsd_syn}\includegraphics[height=2.5cm]{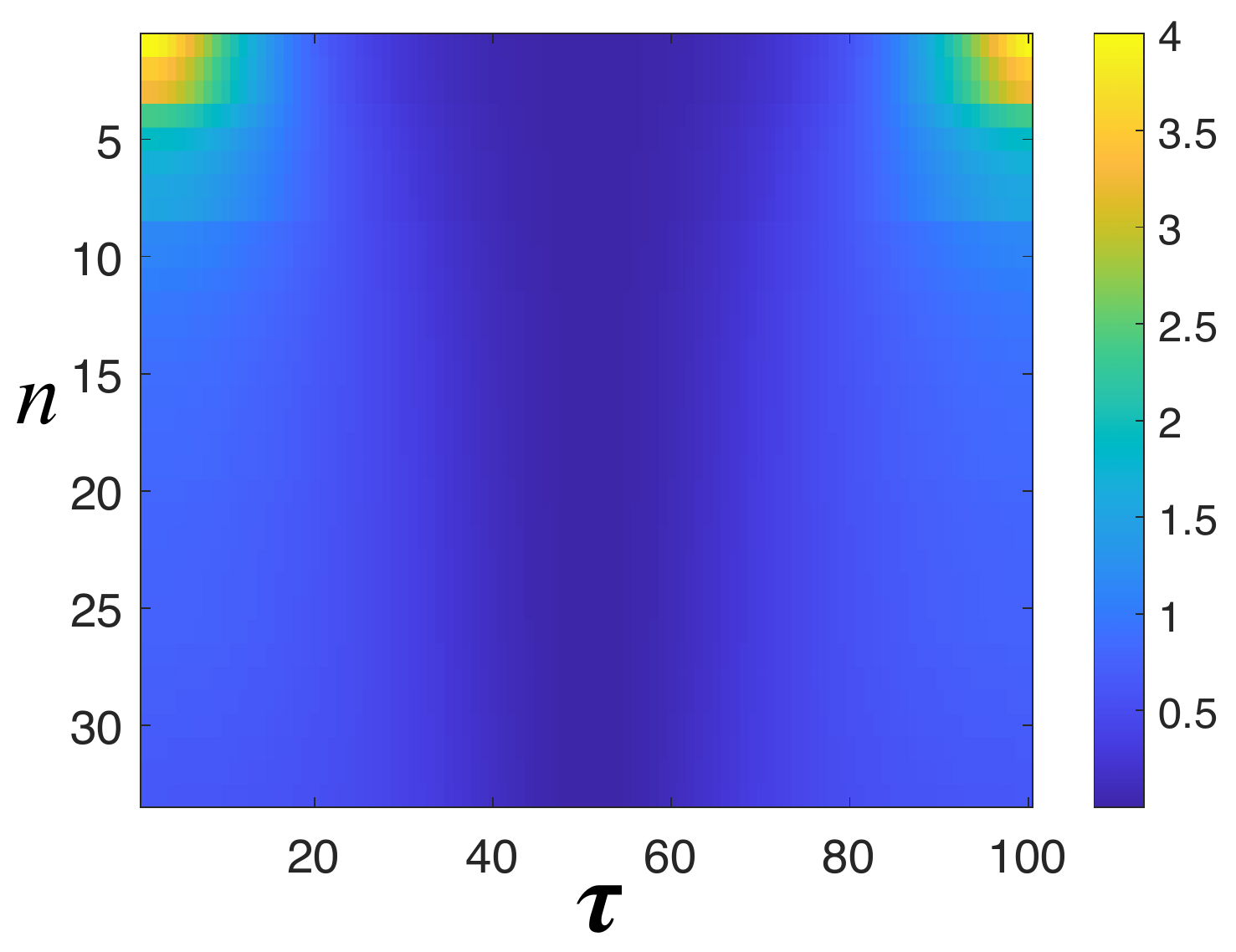}}
     \hspace{0.5cm}
    \subfloat[Realization at $t=25$]
    {\label{fig_proc_real_t25}\includegraphics[height=2.5cm]{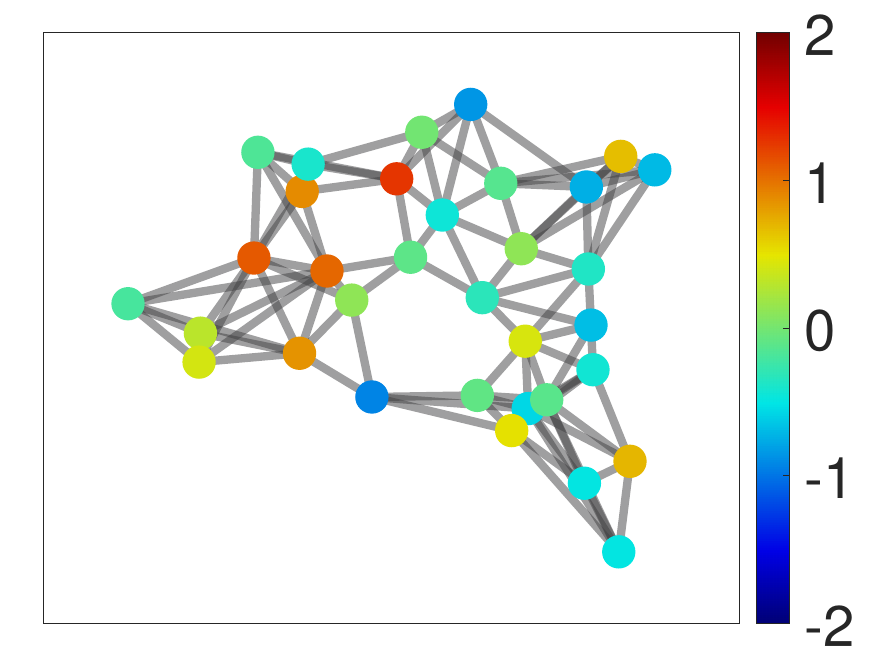}}\\
     \subfloat[Realization for nodes $i=7,15,27$]
    {\label{fig_proc_real_nodes7_15_27}\includegraphics[height=3.0cm]{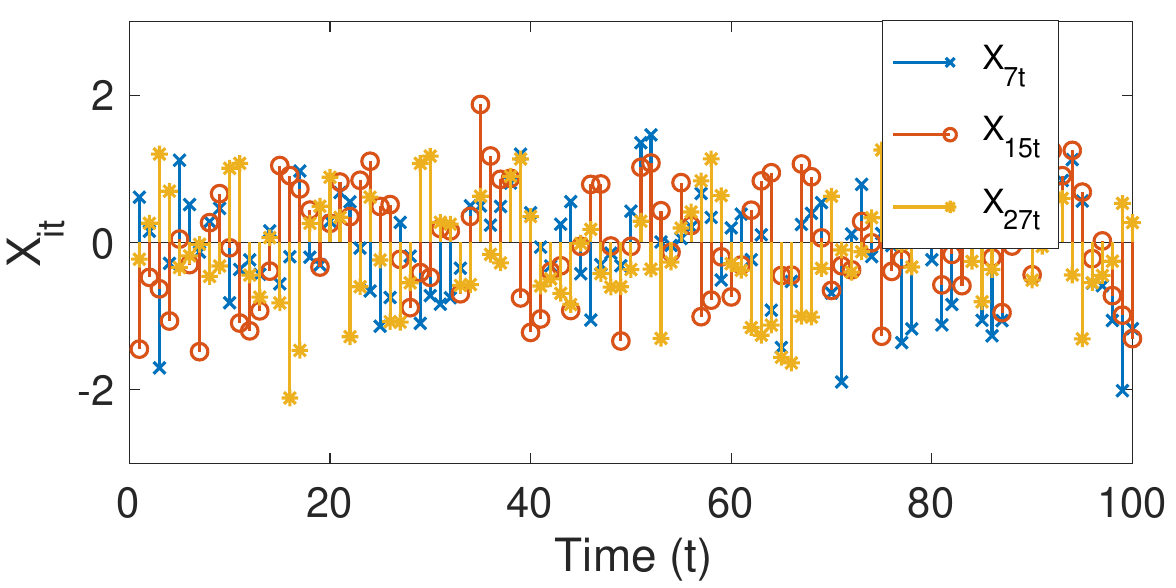}}\\   
 \end{center}
	\caption{JPSD and an example realization of the synthetic process}
 \label{fig_proc_jpsd_real}
\end{figure}

We begin with analyzing the model estimation performance on a synthetically generated graph ARMA process. We experiment on a real graph topology constructed from  the Mol\`ene weather data set  \cite{Girault-Stationarity}, consisting of $\graphDim=33$ meteorological observation stations each of which is represented as a graph node. We form a \kNN{5} graph with Gaussian edge weights computed as  $\weightMat_{\ijNode}=\exp(-\| \locvect_\iNode - \locvect_\jNode \|^2/\sigma^2)$, where $ \{ \locvect_\iNode \}$ denote the locations of the stations and $\sigma$ is a scale parameter. The normalized graph Laplacian has been used in the performance analysis experiments with synthetic data. We synthetically generate realizations of an ARMA graph process with time length $T=100$ according to the process model \eqref{ARMA polynomial model} on this topology, where the model order parameters are set as $\modelOrderP=1$, $\modelOrderK=1$, $\modelOrderQ=1$, $\modelOrderM=0$. The parameter vectors defined in \eqref{eq_defn_ab1} are set to have the ground truth values $\avect = \big[ -0.5  \ \ 0.5 \big]^H$ and $ \bvect= \big[ 0.5 \ \ 0.5 \big]^H$. The JPSD of the process generated with these parameters is shown in Fig.~\ref{fig_proc_jpsd_real}(a) and an example realization of the process is shown in Fig.~\ref{fig_proc_jpsd_real}(b)-\ref{fig_proc_jpsd_real}(c) at one time instant and three graph nodes. 

In order to clearly observe the effect of the number of realizations $L$ on the model estimation accuracy, we learn a model from $L$ complete realizations of the process without any missing observations by solving the relaxed optimization problem \eqref{eq_opt_AB_v5}. The realizations of the process are corrupted with additive white Gaussian noise at several noise levels, and the variation of the model estimation accuracy with the number of realizations $L$ is studied at each noise level. We evaluate the normalized estimation errors $\| \as - \ag\| / \| \ag \|$ and $\| \bs - \bg\| / \| \bg \|$ for the model parameter vectors $\avect$, $\bvect$; and the normalized estimation error $\| \hs - \hg\| / \| \hg \|$ for the JPSD vector, where  $\ag$, $\bg$, $\hg$ denote the true vectors and $\as$, $\bs$, $\hs$ denote their estimates.

\begin{figure}[t]
\begin{center}
     \subfloat[JS-ARMA]
     {\label{fig_a_err_snr}\includegraphics[height=3.4cm]{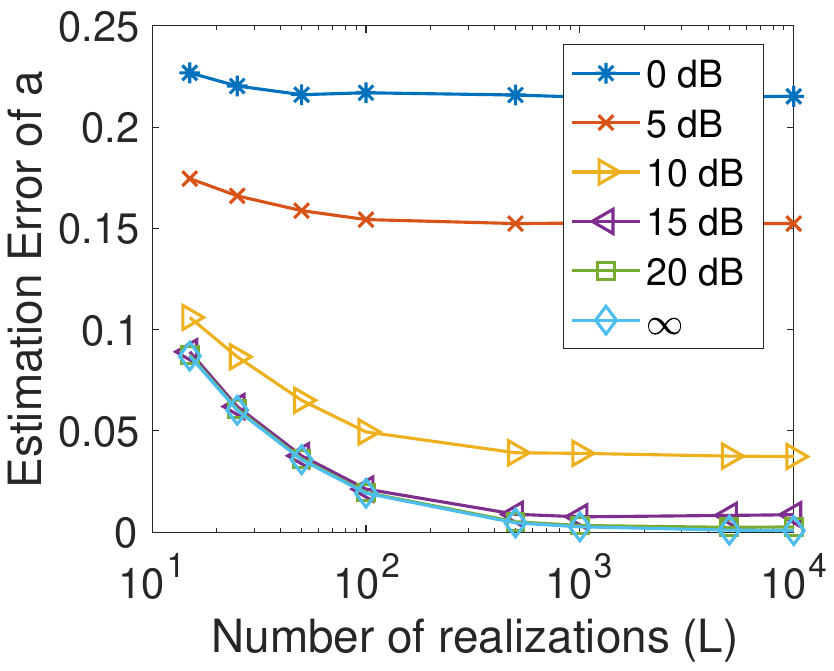}}
     \subfloat[JS-ARMA]
     {\label{fig_b_err_snr}\includegraphics[height=3.4cm]{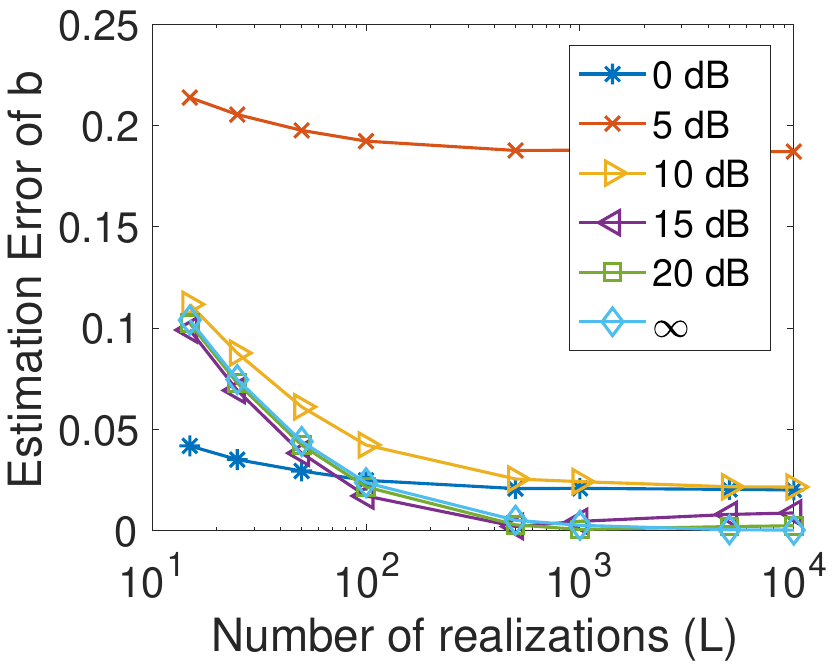}}\\
     \subfloat[JS-ARMA]
     {\label{fig_jpsd_err_snr}\includegraphics[height=2.4cm]{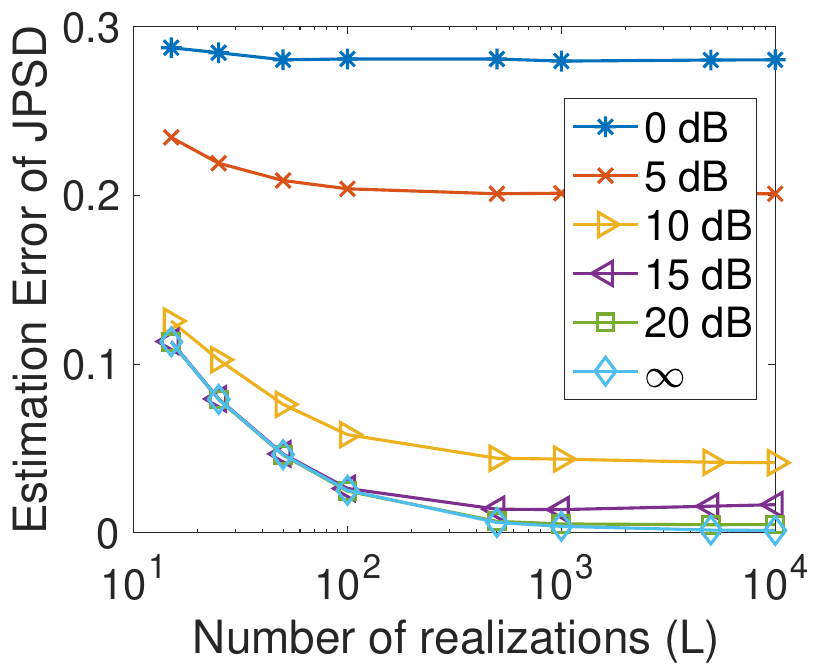}}
       \subfloat[JSNC-ARMA]
     {\label{fig_jpsd_err_snr_ncarma}\includegraphics[height=2.4cm]{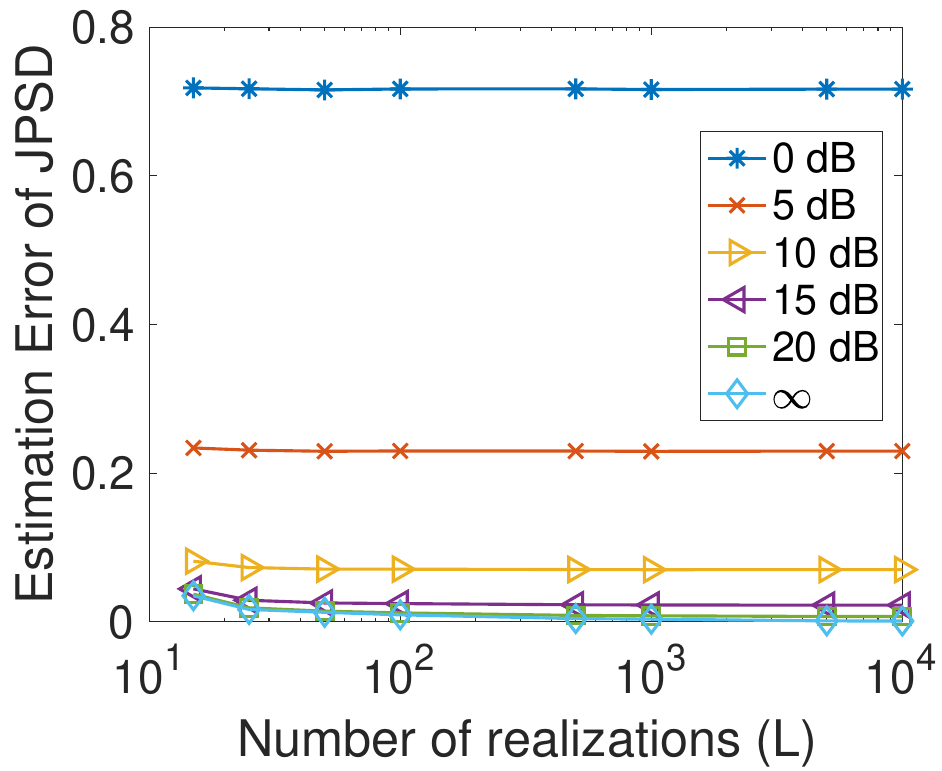}}
       \subfloat[JWSS]
     {\label{fig_jpsd_err_snr_jwss}\includegraphics[height=2.4cm]{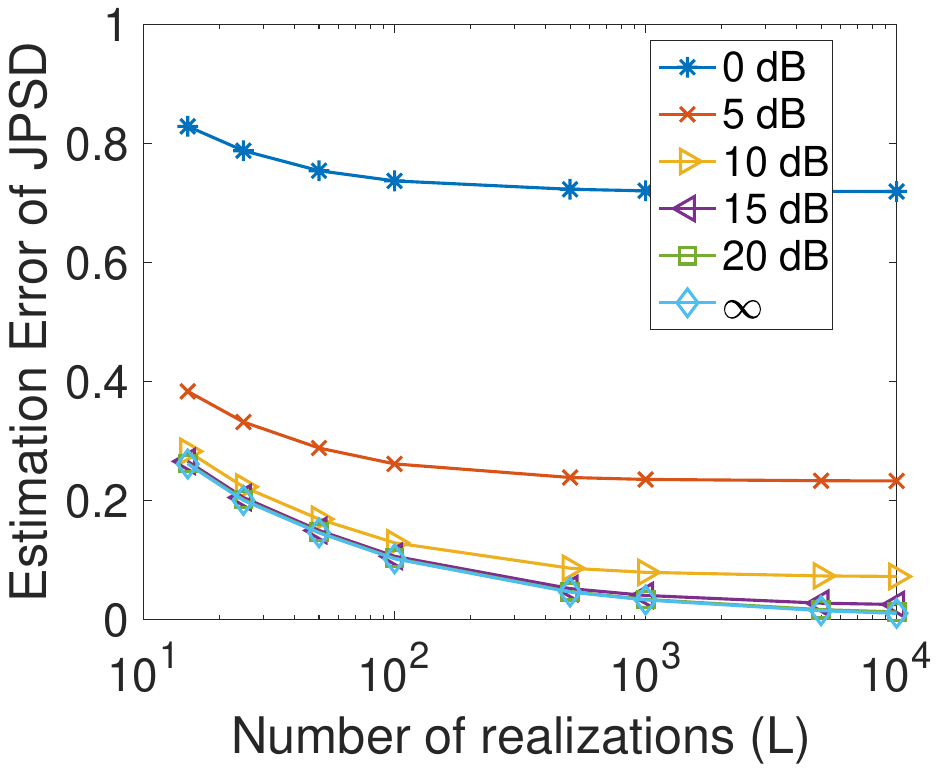}}
 \end{center}
	\caption{Variation of the estimation error under noise}
 \label{fig_err_vs_L_noise}
\end{figure}

The estimation errors of the parameter vectors $\avect$, $\bvect$ and the JPSD are given in Fig.~\ref{fig_err_vs_L_noise}  at different SNR (signal to noise ratio) levels. In order to better understand the performance of our algorithm, in Fig.~\ref{fig_jpsd_err_snr_ncarma}-\ref{fig_jpsd_err_snr_jwss} we also present the errors obtained with the two following methods: The first method (JSNC-ARMA) is based on minimizing the nonconvex objective function \eqref{eq_opt_prob1} with a local descent-type optimizer, by initializing it with 10 different randomly selected vectors in the solution space and choosing the one that yields the smallest objective function value after optimization. The second method is the non-parametric JWSS process model (JWSS) \cite{Natheneal-Joint-Stationarity}, which provides our JS-ARMA algorithm with the initial estimate of the joint spectrum as discussed in Section \ref{ssec_jpsd_estimate}. 

In Fig.~\ref{fig_err_vs_L_noise}, an SNR value of 15 dB is sufficient to provide a model estimation accuracy close to the ideal case of infinite SNR. At high SNR, the estimation errors efficiently converge to 0 as the number of realizations increases. At small SNR values, the solution given by the nonconvex objective  \eqref{eq_opt_prob1} in JSNC-ARMA results in higher estimation error than JS-ARMA, demonstrating the difficulty of the original problem when signals show some divergence from the assumed model. This difficulty is efficiently addressed by the convex relaxations employed in JS-ARMA. At higher SNR values, while JS-ARMA has slightly larger error than JSNC-ARMA for small $L$, the estimation error of  JS-ARMA drops to $0$ as the number of realizations $L$ increases. This confirms that the convex problem \eqref{eq_opt_AB_v5} derived from the original nonconvex problem \eqref{eq_opt_prob1}  through several approximations and relaxations is capable of accurately recovering the true process model. We can also observe that the initial spectrum given by the JWSS method is less accurate than that of the proposed JS-ARMA method at all noise levels and number of realizations. The  initial JWSS estimate is affected by two principal error sources; namely, the deviation of the data from the stationary process model due to noise, and the finite sample effects in the estimation of the covariance matrix. The proposed JS-ARMA method alleviates the effects of both of these error sources by projecting the initial JWSS estimate onto the spectrum manifold of graph ARMA processes.




%
%


\begin{figure}[t]
\begin{center}
    \subfloat[JS-ARMA]
    {\label{fig_a_err_QMPK_L}\includegraphics[height=3.4cm]{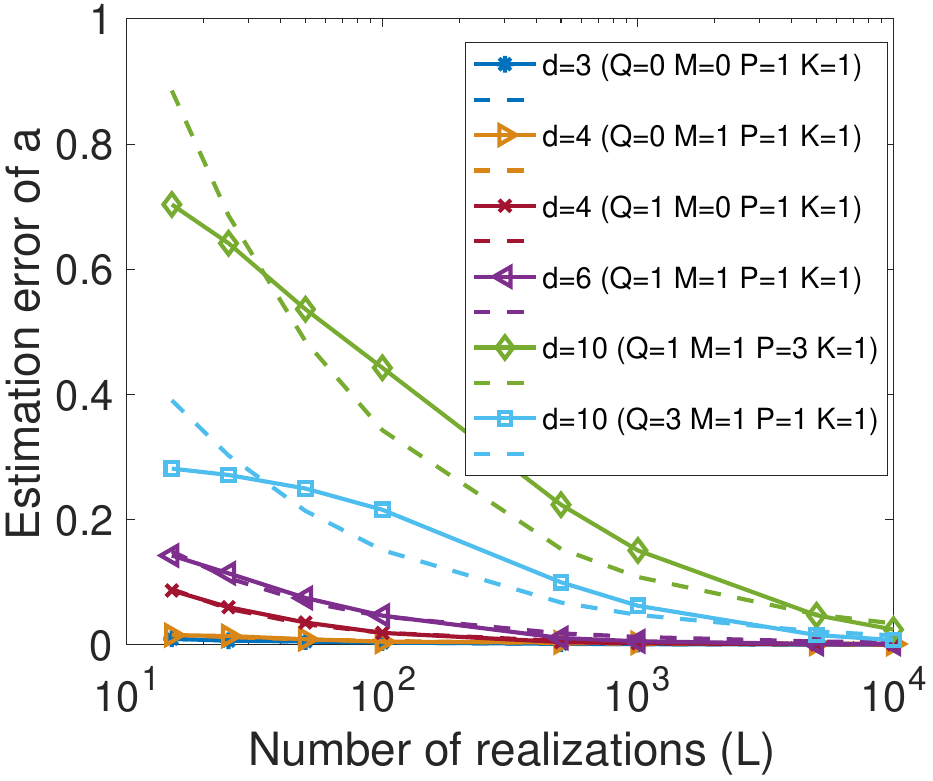}}
     \subfloat[JS-ARMA]
    {\label{fig_b_err_QMPK_L}\includegraphics[height=3.4cm]{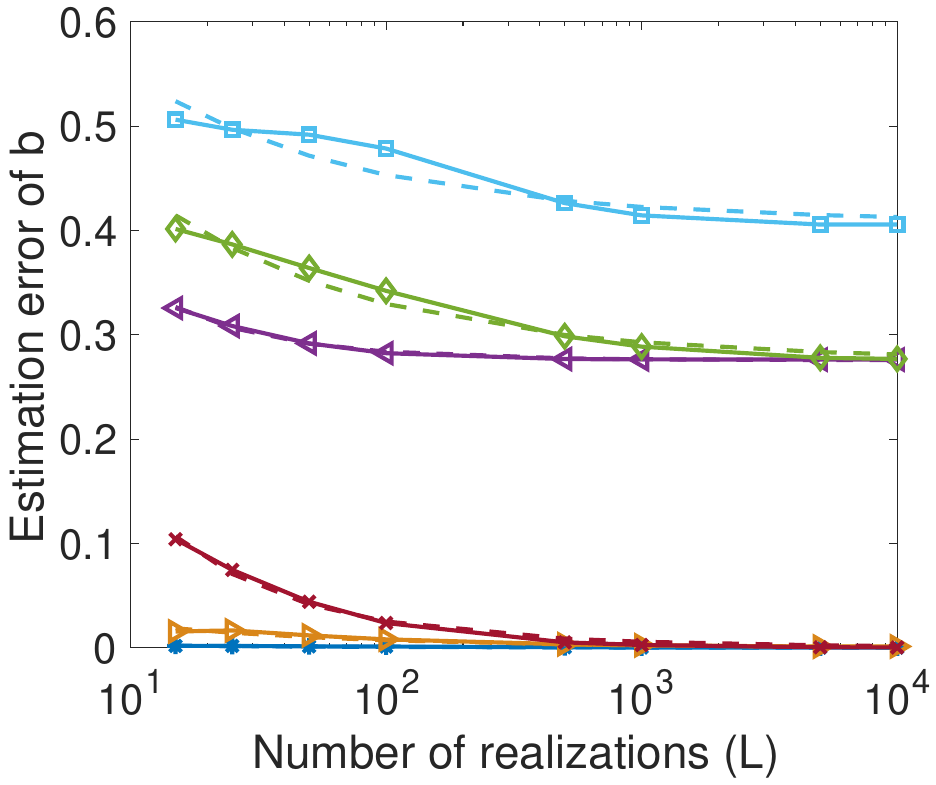}}\\
     \subfloat[JS-ARMA]
    {\label{fig_jpsd_err_QMPK_L}\includegraphics[height=3.4cm]{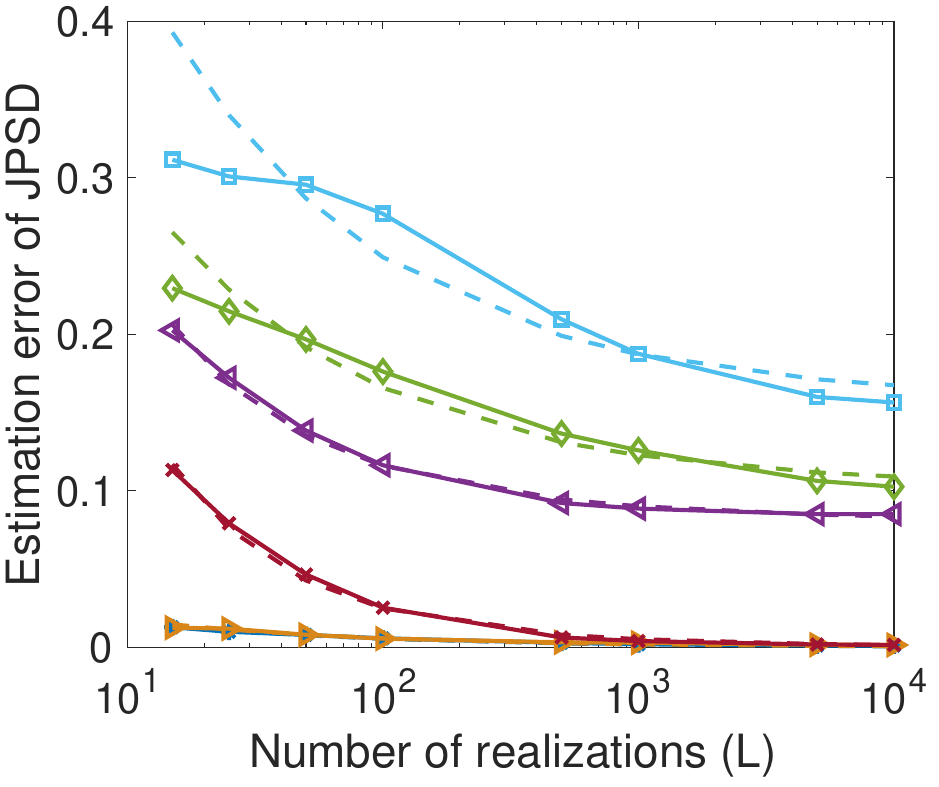}}
      \subfloat[JWSS]
    {\label{fig_jpsd_err_QMPK_jwss_L}\includegraphics[height=3.4cm]{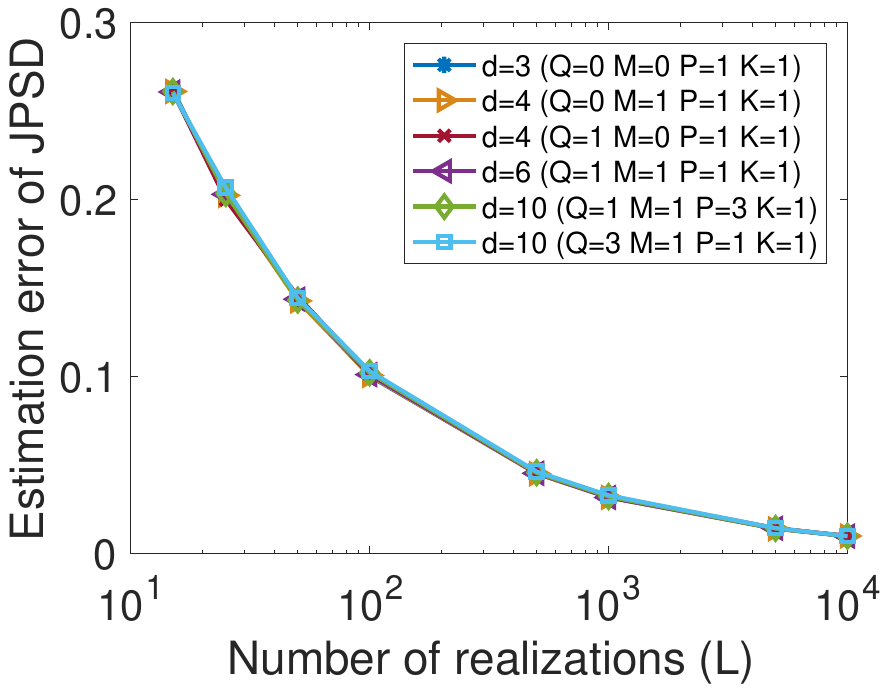}}
    \\
     \subfloat[JS-ARMA]
    {\label{fig_a_err_QMPK_d}\includegraphics[height=2.7cm]{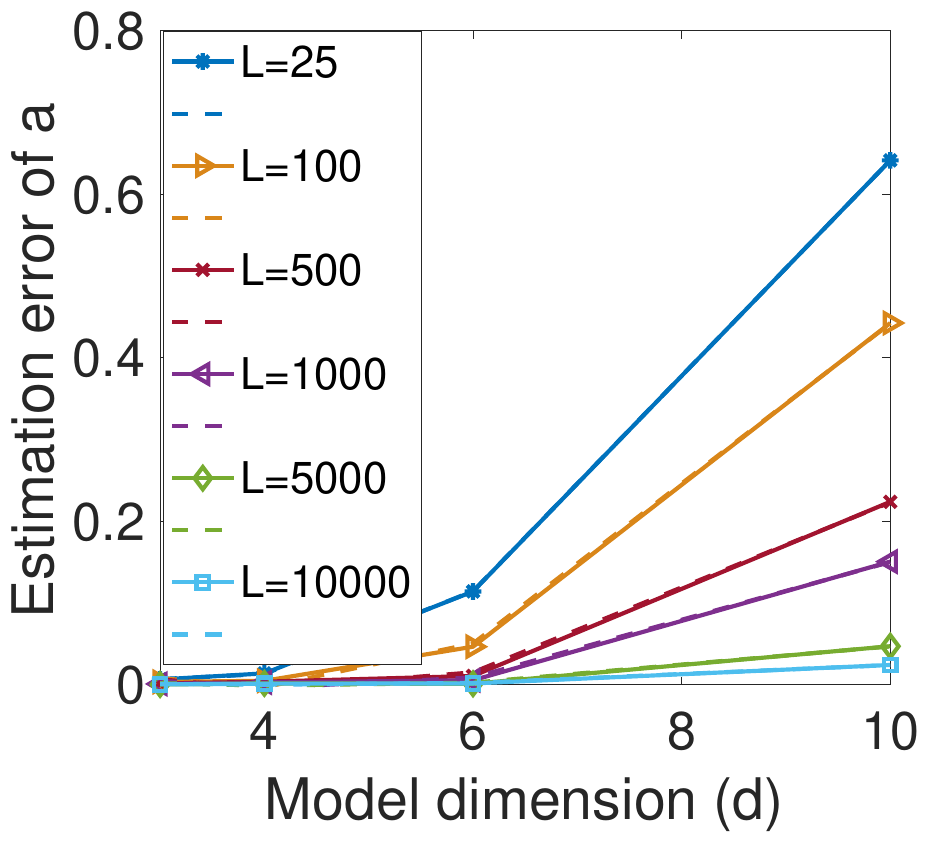}}
         \subfloat[JS-ARMA]
    {\label{fig_b_err_QMPK_d}\includegraphics[height=2.7cm]{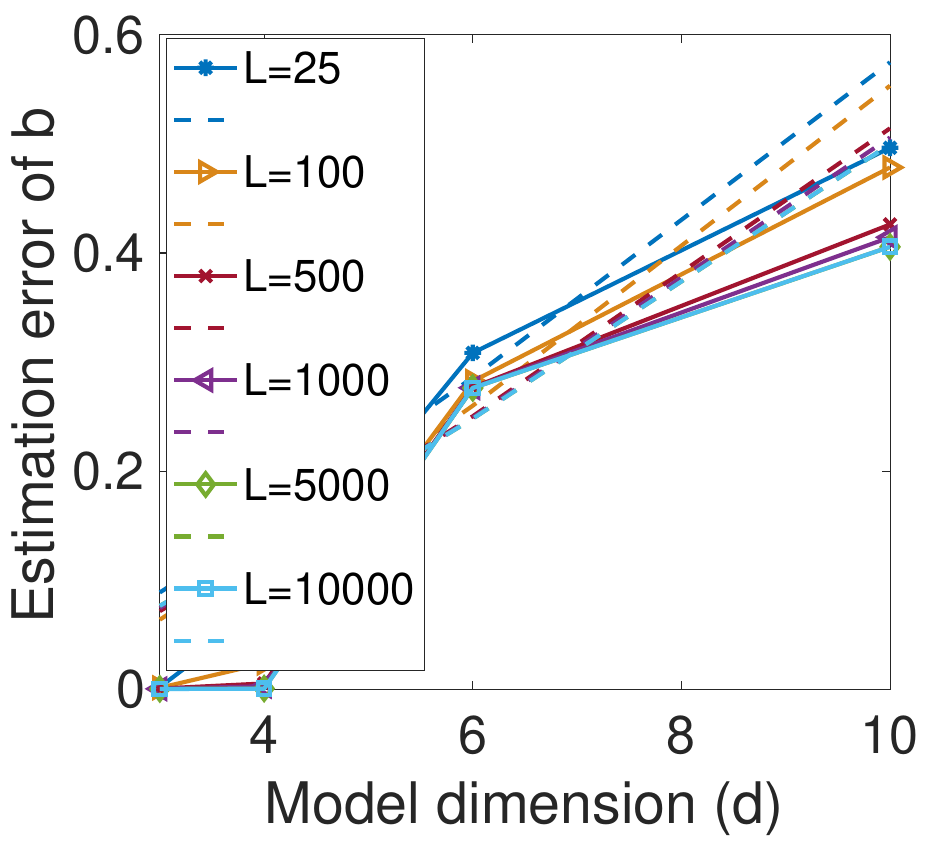}}
         \subfloat[JS-ARMA]
    {\label{fig_jpsd_err_QMPK_d}\includegraphics[height=2.6cm]{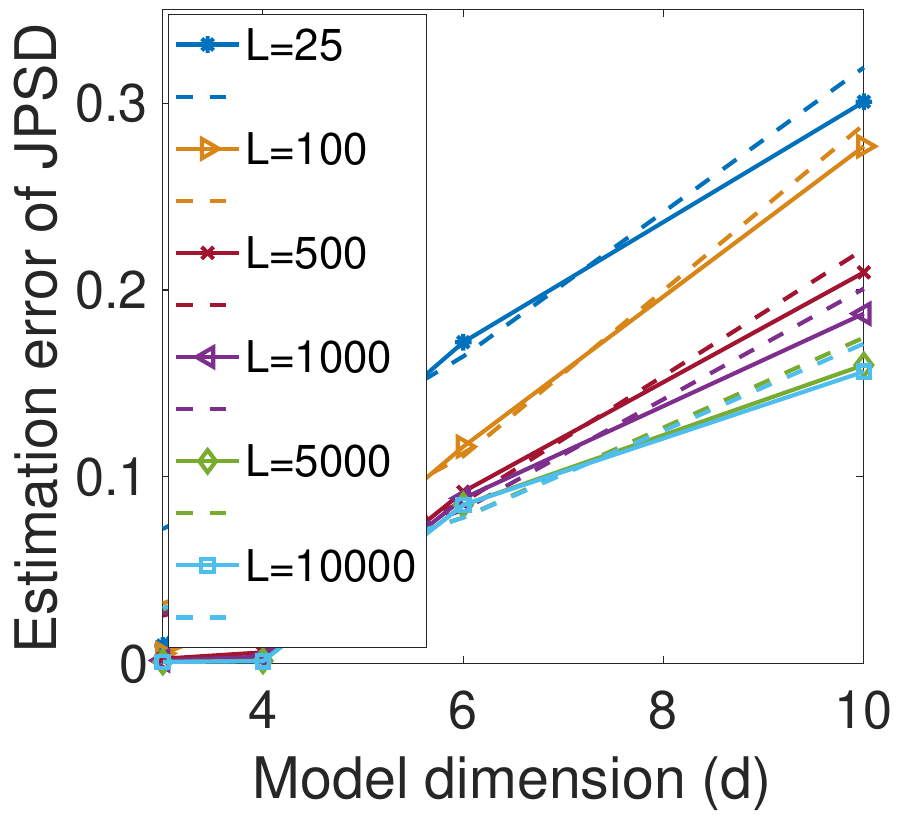}}
 \end{center}
	\caption{Variation of the estimation error with the number of realizations and model dimension. Panels (b) and (c) share the same legends as (a).}
 \label{fig_err_vs_L_d}
\end{figure}

\subsubsection{Effect of model complexity}  We next study how the number of realizations required for accurate model estimation evolves in relation to the model complexity. 6 different ARMA processes are generated with variable model orders $\modelOrderP$, $\modelOrderK$, $\modelOrderQ$, $\modelOrderM$ on the graph topology used in the first experiment. In order to restrict the scope of the experiment to the estimation of the model parameters $\avect$, $\bvect$, and the JPSD, the ground truth model orders are provided to the algorithm. For each combination of  $\modelOrderP$, $\modelOrderK$, $\modelOrderQ$, $\modelOrderM$, the normalized estimation errors of $\avect$, $\bvect$, and the JPSD are plotted in Fig.~\ref{fig_a_err_QMPK_L}-\ref{fig_jpsd_err_QMPK_L} for the proposed JS-ARMA method. The JPSD errors are also presented for the JWSS algorithm \cite{Natheneal-Joint-Stationarity} in Fig.~\ref{fig_jpsd_err_QMPK_jwss_L}, which provides the initial estimate of the JPSD to our algorithm as input.

We recall from Theorem \ref{thm_jpsd_error_conv_rate} that the estimation errors are expected to converge at rate $O(\sqrt{N^2 T^2 \dc ^4 / L})$ as the number of realizations $L$ increases. In order to experimentally verify the theoretical convergence rate of $O(1/\sqrt{L})$, in Fig.~\ref{fig_a_err_QMPK_L}-\ref{fig_jpsd_err_QMPK_L}, for each estimation error curve, we fit a polynomial containing the term $1/\sqrt{L}$ and its higher-order powers that decay faster, which are shown with dashed lines. No constant terms have been allowed in the polynomials in Fig.~\ref{fig_a_err_QMPK_L}, where the consistency between the experimental and the theoretical curves indicates that the estimation error of $\avect$ indeed approaches 0 at a rate no slower than $O(1/\sqrt{L})$ as theoretically predicted. Although the theoretical and experimental plots in Fig.~\ref{fig_b_err_QMPK_L} and \ref{fig_jpsd_err_QMPK_L} also exhibit strong agreement, in these plots, constant terms have been excluded from the theoretical curves for small model orders $d=3, 4$ and included in them for larger model orders $d=6,10$. While the results in Fig.~\ref{fig_b_err_QMPK_L} and \ref{fig_jpsd_err_QMPK_L} confirm that the $\bvect$ and the JPSD estimation errors change with $L$ at rate $O(1/\sqrt{L})$ for all $d$; the convergence of the error to 0 at small model orders is replaced by convergence to a nonzero error component at higher model orders. A probable explanation for this nonzero error component may be that the solution of the modified convex problem \eqref{eq_opt_AB_v5} obtained through several approximations and relaxations might deviate from that of the original problem \eqref{eq_opt_prob1} at large model orders. This hypothesis is also supported by the fact that the initial JWSS estimate outperforms the JS-ARMA estimate at large model orders in Fig.~\ref{fig_jpsd_err_QMPK_jwss_L}. On the other hand, at small model orders $d=3,4$,  the convex problem \eqref{eq_opt_AB_v5} successfully approximates the nonconvex problem \eqref{eq_opt_prob1} and JS-ARMA recovers the true solution, performing better than JWSS. Similarly to the infinite SNR scenario studied in Fig.~\ref{fig_err_vs_L_noise}, the JSNC-ARMA algorithm yields quite small estimation error in this noiseless setting, whose plots are skipped here for brevity.

 
 As for the dependence of the error on the model order, Theorem \ref{thm_jpsd_error_conv_rate} states that the JPSD error $\| \hs - \hg \|$ and the non-normalized parameter estimation error $\| \cs - \cg\|$ increase with the model dimension $\dc=P(K+1) + (Q+1)(M+1)$ at a rate bounded by $O(\dc ^2) $. One may then expect the relation $\| \cs - \cg\| / \| \cg \| = O(\dc^{3/2}) $ for the normalized parameter error. In order to verify these bounds, in Fig.~\ref{fig_a_err_QMPK_d}-\ref{fig_jpsd_err_QMPK_d} we present the experimental variations of $\avect$, $\bvect$, and the JPSD errors with the model dimension $d$ at different realization numbers. We also plot with dashed lines the corresponding theoretical curves fit to the experimental data, with highest-order terms set as $\dc^{3/2}$ for the $\avect$ and $\bvect$ errors and $\dc ^2$ for the JPSD errors. The agreement between the experimental and the theoretical curves confirms the validity of the theoretical bounds. In fact, in contrast to the $\avect$ error, we visually observe that the actual rate of increase of the $\bvect$ and the JPSD errors may even be slower, e.g., closer to the linear rate $O(\dc)$ than the theoretical rates $O(\dc^{3/2})$ and $O(\dc ^2) $, which are valid but not necessarily always tight upper bounds.


\subsubsection{Sensitivity to weight parameters} We next examine the sensitivity of JS-ARMA to the weight parameters $\muA$ and $\muB$ of the objective function. We conduct the experiment on the COVID-19 pandemic data set\footnote{A real data set is preferred in this experiment, since synthetic data largely adheres to the assumed graph ARMA model and the algorithm naturally tends to learn rank-1 $\Amat$ and $\Bmat$ matrices even at very small $\muA$ and $\muB$ values.} described in Section \ref{ssec_exp_realdata}, which consists of the number of daily new COVID-19 cases reported on the graph of European countries. In each repetition of the experiment, an ARMA process model of order $P=2$, $K=0$, $Q=1$, $M=1$ is learnt from partially observed process realizations with varying $(\muA, \muB)$ combinations, and the missing entries of the realizations are computed via MMSE estimation as explained in Section \ref{ssec_lmmse_estimate}.  The normalized mean errors of the  estimates of the missing process observations are computed as 
\begin{equation}
\label{eq_nme_defn}
NME = \left( \sum_{l=1}^L  \|   {\bar \zSignal}^l - ({\bar \zSignal}^l)^\est  \|^2 /  \sum_{l=1}^L  \|  {\bar \zSignal}^l \|^2 \right)^{1/2}
\end{equation}
where ${\bar \zSignal}^l$ and $({\bar \zSignal}^l)^\est$ denote respectively the missing observations and their estimates in \eqref{eq_lmmse_est_z}. The NME values are reported in Table \ref{parameter_sensitivity} for different $(\muA, \muB)$ combinations, which are averaged over  $18$ repetitions of the experiment with different random selections of the missing observations and over a range of missing observation ratios varying between 10\% and 80\%.  We first observe that setting $\muA$ and $\muB$ to 0 or too small values results in very high estimation errors, which serves as an ablation study for the $\tr(\Amat)$ and $\tr(\Bmat)$ terms in the objective function in \eqref{eq_opt_AB_v5}. Although the minimum NME of $0.17$ is attained at relatively high values of $\muA$,  the intervals $\muB \in [10,100]$, $\muA \in [0.001, 0.1]$ define a safe region that provides stable estimation performance, offering a suitable trade-off between fitting the model to the initially estimated spectrum and ensuring the low-rank structures of the $\Amat$ and $\Bmat$ matrices. The results on other data sets have led to similar conclusions as well, which are skipped here for brevity.




\begin{table}[t]
	\centering
	\scriptsize
	\begin{tabular}{|l|l|l|l|l|l|l|l|l|} \hline
\diagbox[width=1cm]{$\muA$}{$\muB$}& $0$        & $0.001$        & $0.01$        & $0.1$          & $1$ & $10$ & $100$ & $1000$    \\   \hline
	$0$               & 15.10 & 11.61 & 10.01 & 8.50         & 5.65   & 0.22   & 0.30   & 23.68   \\	\hline
	$0.001$               & 10.00 & 9.52 &10.68 & 8.45          & 7.16   & \textbf{0.22}   & \textbf{0.21}   & 20.62   \\	\hline
	$0.01$               & 9.42 & 9.47 & 9.61 & 7.83          & 5.70   & \textbf{0.22}   & \textbf{0.22}   & 21.71   \\	\hline
	$0.1$               & 10.17 & 10.68 & 10.75 & 9.81 & 4.79   & \textbf{0.22}   & \textbf{0.21}   & 11.75   \\	\hline
	$1$                & 8.98          & 18.84          & 6.22          & 3.29          & 3.12   & 0.20   & 0.35   & 21.29   \\	\hline
	$10$                & 5.39          & 5.22          & 4.56          & 2.67          & 2.50   & 0.19   & 0.26   & 5.08   \\	\hline
	$100$                & 2.14          & 2.04          & 2.00          & 1.84          & 1.84   & 0.33   & \textbf{\underline{0.17}}   & 0.18   \\	\hline
	$1000$                & 0.43          & 0.43          & 0.43          & 0.42          & 0.35   & 0.52   & \textbf{\underline{0.17}}   & \textbf{\underline{0.17}}   \\	\hline
	\end{tabular}
	\caption{Variation of the NME with $\muA$ and $\muB$}
		\label{parameter_sensitivity}
\end{table}

\begin{figure}[h]
\begin{center}
     \subfloat[]
     {\label{fig_time_comp_N}\includegraphics[height=2.5cm]{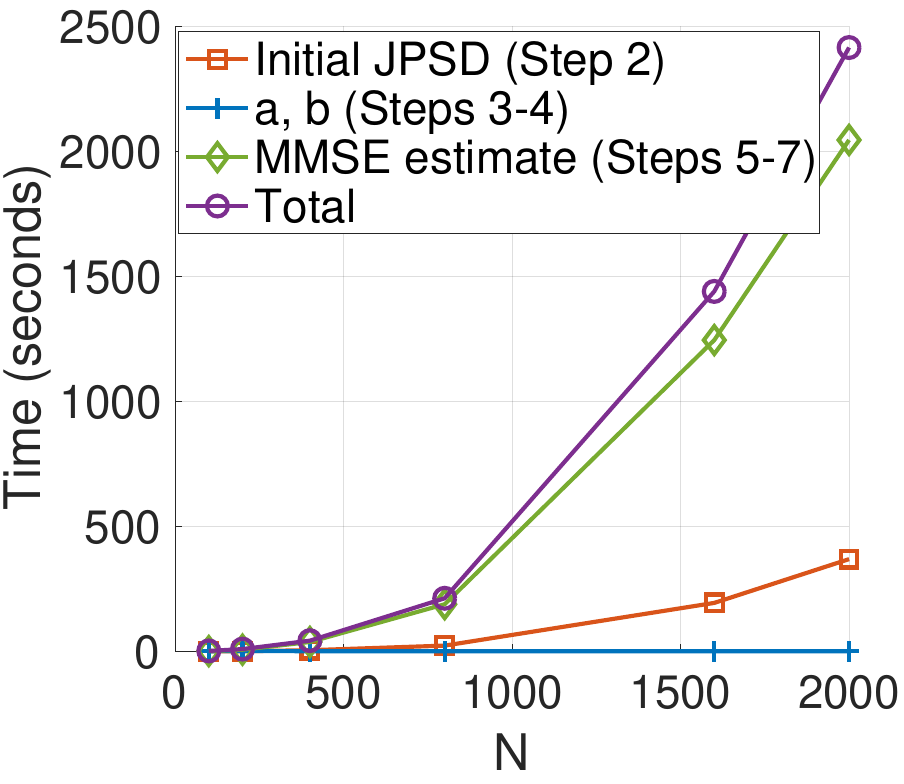}}
     \subfloat[]
     {\label{fig_time_comp_T}\includegraphics[height= 2.5cm]{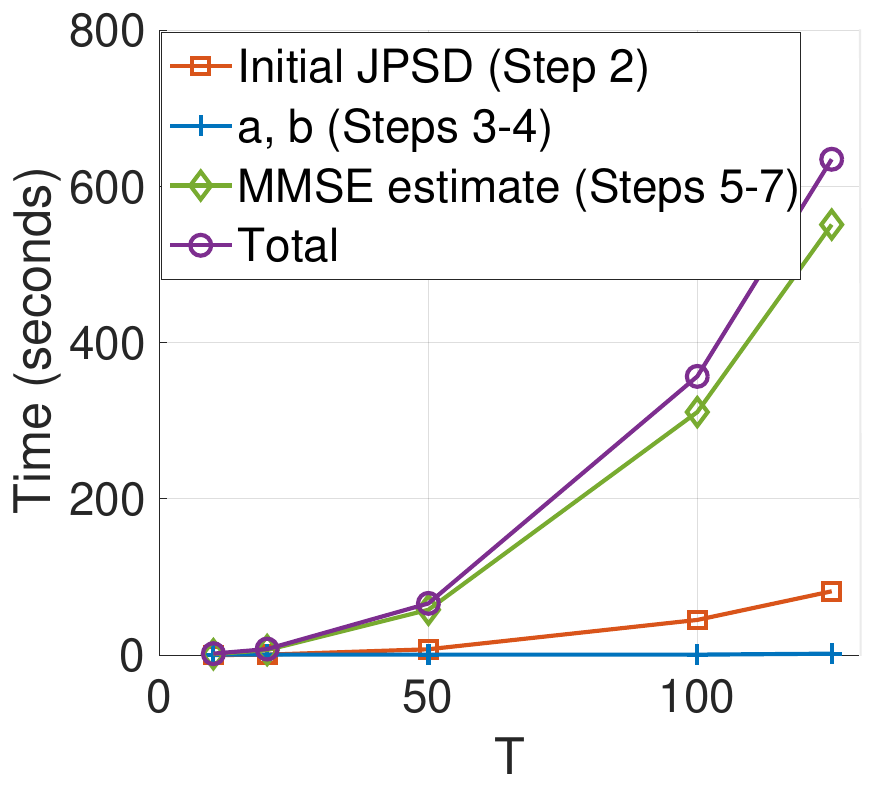}}
     \subfloat[]
     {\label{fig_time_comp_NT}\includegraphics[height=2.5cm]{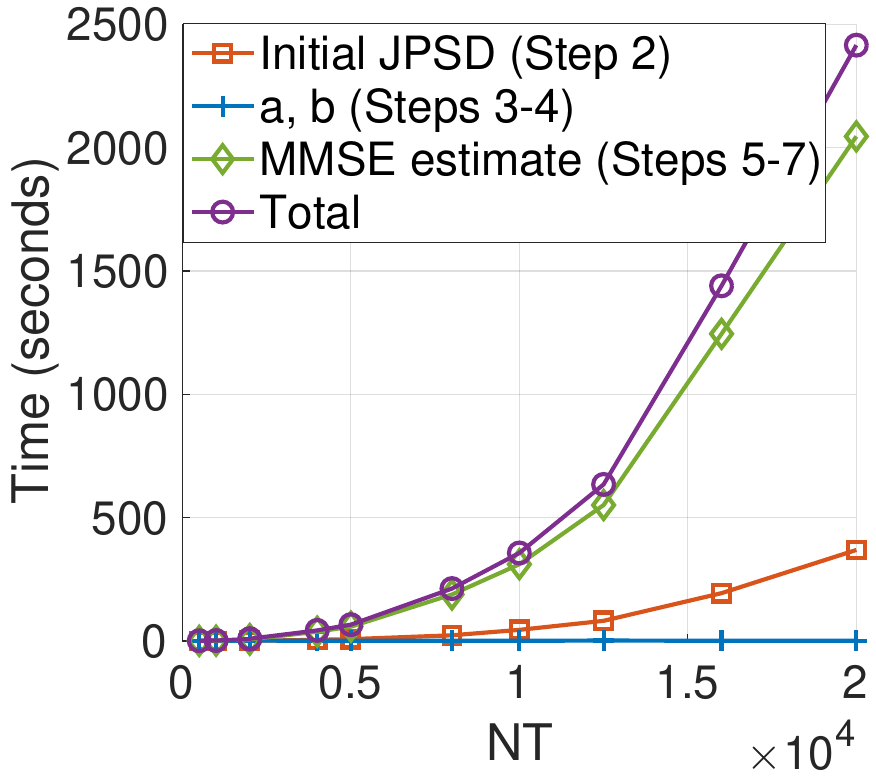}}
 \end{center}
	\caption{Variation of the runtime of JS-ARMA with graph size and time length}
 \label{fig_runtime_analysis}
\end{figure}

\subsubsection{Computational complexity evaluation}

We lastly analyze the computational complexity of the proposed method numerically. We construct a $7$-NN synthetic graph with Gaussian edge weights from $N$ nodes with random locations on a two-dimensional plane. We then generate $L=100$ realizations of a graph ARMA process of time length $T$, with model parameters $P=K=Q=1$, $M=0$ and coefficient vectors  $\avect = \big[ -0.6  \ \ 0.1 \big]^H$ and $ \bvect= \big[ 1 \ \ 1 \big]^H$. Randomly selected
$20\%$ of the process observations are considered as missing. A graph ARMA model is computed from the available observations, and the missing observations are estimated as described in Algorithm \ref{alg_tv_process} for variable $N$ and $T$ values. The experiment is done on a laptop computer with 32 GB RAM and 4.5 GHz processor using a MATLAB implementation. In order to better understand the computational complexity, in addition to the total runtime, we also report the individual runtimes of the following stages of our method in  Algorithm \ref{alg_tv_process}: The computation of the initial JPSD $\tilde h$ (Step 2); the computation of the model parameters $\avect$ and $\bvect$ (Steps 3-4); and the MMSE estimation of the missing observations $(\bar \zSignal^l)^\est $ from the learnt model (Steps 5-7). The runtimes are analyzed in Fig.~\ref{fig_runtime_analysis} for three different cases where graph size $N$ varies at fixed time length $T=10$;  time length $T$ varies at fixed graph size $N=100$; and the $NT$ product varies. (The JPSD estimation error and the NME of the MMSE estimations are verified to remain under 0.33 and 0.41 respectively in all experiments, confirming the validity of the computed model.)

In Fig.~\ref{fig_runtime_analysis}, the estimations of the initial JPSD (Step 2) and the missing observations (Steps 5-7) are seen to have significantly higher runtimes than the essential part of our method (Steps 3-4) where the model parameters $\avect, \bvect$ are computed by solving \eqref{eq_opt_AB_v5}.  In particular, the runtime of solving \eqref{eq_opt_AB_v5} in Steps 3-4 has always remained below 2 seconds in these experiments. We recall from the complexity analysis in Section \ref{ssec_complexity_anlys} that Steps 3-4 have linear complexity $O(NT)$ in graph size $N$ and time length $T$, while Step 2 and Steps 5-7 have cubic complexity $O(N^3 T^3)$. Although this results in an overall complexity of $O(N^3 T^3)$, in applications with large data sizes, one can reduce the complexity by preferring approximate implementations for Steps 2 and 5-7, where the primary computational bottleneck lies. In Step 2, the initial JPSD $\tilde h$ can be approximately estimated by computing only the largest eigenvalues of $\tilde\covMat_{\xv}$ via, e.g.,~ Krylov methods, which would have much lower complexity than a complete eigenvalue decomposition. Similarly, the MMSE estimate in Steps 5-7 can be substituted by an alternative less complex estimator, e.g., via sequential ARMA recursions based on the model parameters $\avect, \bvect$.

\subsection{Comparative Experiments on Real Data Sets}
\label{ssec_exp_realdata}

In this section, we evaluate the performance of our method with comparative experiments. The following time-vertex data sets have been used in the experiments:

\textit{1) Mol\`ene weather data set:} The experiment is conducted on hourly weather measurements collected in the Brittany region of France during January 2014 \cite{Girault-Stationarity}. We experiment on temperature measurements taken on $\graphDim=37$ different weather stations, each of which is represented as a graph node. We construct a $10$-NN graph with Gaussian edge weights as explained in Section \ref{sec_exp_sensit_analy}. We regard each $24$-hour measurement sequence as one realization of a time-vertex graph process $\timeVertexSignalMat $ with graph size $\graphDim=37$ and time length $T=24$, obtaining a total of $L=31$ realizations.

\textit{2) COVID-19 pandemic data set:} The experiments with COVID-19 data  \cite{Covid19data} are done on the number of daily new cases per country between February 15, 2020 and July 5, 2021. We include the $N=37$ European countries with highest populations in the experiment, where each country is considered as a graph node. A $4$-NN graph is constructed with Gaussian edge weights based on a hybrid distance measure that accounts for both geographical proximities and the number of flights (accessed through \cite{Eurostat}) between each pair of countries. The number of daily new cases are normalized by country populations and smoothed with a moving average filter over a time window of 7 days. The time length of the process is taken as $T=21$ days (three weeks). The experiments are conducted on $L=23$ realizations of the process. 

\textit{3) NOAA weather data set:} We experiment on hourly average temperature measurements from the NOAA weather data \cite{Arguez12} taken within a year from $N = 246$ weather stations across the United States. Each weather station is considered as a graph node and a $7$-NN graph is constructed with Gaussian edge weights. The 24-hour measurement sequences averaged over each week are regarded as a realization of the process. The experiments are therefore conducted on $L=52$ realizations of a process of time length $T=24$.

Since real data already has some natural deviation from the process model considered in our study, no extra noise is added to the data. In order to better interpret our estimation results, we first analyze the joint time-vertex stationarity, the vertex stationarity, and the time stationarity of each data set. Recalling from \eqref{eq_jpsd_est} that the covariance matrix of a time-vertex stationary process must be diagonalizable with the eigenvectors of the joint Laplacian, we follow the convention in \cite{Natheneal-Graph-Stationarity} and compute the time-vertex stationarity ratio of each data set as $\| \text{diag}(\tilde h(\graphEigenvalueMat_{\graph}, \timeEigenvalueMat))\| / \| \tilde h(\graphEigenvalueMat_{\graph}, \timeEigenvalueMat) \|_F$. The vertex stationarity ratio and the time stationarity ratio are computed similarly, by restricting the covariance matrix to the vertex domain or the time domain in \eqref{eq_jpsd_est}, and also replacing the eigenvector matrix with $\graphEigenvectorMat $ or $\timeEigenvectorMat $, respectively. The stationarity ratios of the data sets are reported in Table \ref{tab_stat_ratio}.

We study the signal estimation problem within the following scenarios:

\begin{enumerate}
\item \label{sco_salt_pepper} Missing observations occur at randomly and independently selected time-vertex pairs


\item  \label{sco_entire_graph} Missing observations occur on the entire graph at some randomly selected time instants 

\item \label{sco_forecasting} (Forecasting): Observations are available on the entire graph during the whole time interval $t=1, \dots, T-s$, and graph signals are predicted for the future time instant $t=T$, for forecasting time step $s$ for each realization. Realizations are divided equally into training and test.
\end{enumerate}

With the proposed JS-ARMA method, we learn a process model from the known observations by solving \eqref{eq_opt_AB_v5}.  The weight function $\mu(\lambda_n, \omega_\tau)$ in  \eqref{eq_opt_AB_v5} is set to be a Gaussian function that penalizes the error at low frequencies more severely, where the spectrum of most real graph processes is likely to be concentrated in practice. The MMSE estimate of the missing observations are then found as discussed in Section \ref{ssec_lmmse_estimate}.  We compare the estimation performance of JS-ARMA with the following approaches: Nonconvex version of our method by solving \eqref{eq_opt_prob1} (JSNC-ARMA), non-parametric JWSS process models\footnote{As the computation of the JFT is not possible in this setting with missing process observations, we use a variant of the original method \cite{Natheneal-Joint-Stationarity} by estimating the JPSD from the covariance matrix as in \eqref{eq_jpsd_est} and refining it by extracting its diagonal entries.} (JWSS) \cite{Natheneal-Joint-Stationarity}, graph vector autoregressive recursions (G-VAR) \cite{IsufiLPL19}, graph polynomial vector autoregressive recursions (GP-VAR) \cite{IsufiLPL19}, ARMA vertex process models (Vertex-ARMA) \cite{Marques-Stationarity}, vector autoregressive process models (VAR) \cite{Lutkepohl05}, AR time process models (AR) \cite{hayes96}, time-vertex signal reconstruction via Sobolev smothness (GraphTRSS) \cite{GiraldoMGTB22}, and deep algorithm unrolling (NestDAU) \cite{NagahamaYTCE22}. While the original deep unrolling method \cite{NagahamaYTCE22} (shown as NestDAU-Vertex in our experiments) addresses the reconstruction of signals in the vertex domain, we also adapt it to our time-vertex setting through the use of the Cartesian product graph, which is represented as NestDAU-TimeVertex.\footnote{Note that the vertex-domain methods NestDAU-Vertex and Vertex-ARMA are applicable only for Scenario \ref{sco_salt_pepper}. The NestDAU methods have been excluded from the experiments on the NOAA dataset as the simulations were repeatedly terminated by the computer due to lack of memory.}  For all methods, algorithm hyperparameters such as model orders and weight parameters are determined via validation.

\begin{table}[t]
	\centering
	\scriptsize
\begin{tabular}{|c|c|c|c|}
\hline
Dataset & Time Stationarity & Vertex Stationarity & Time-Vertex Stationarity\\ 
\hline
Mol\`ene & 0.8955 & 0.9365 & 0.9203\\
\hline
COVID-19 & 0.9963 & 0.7608 & 0.7525\\
\hline
NOAA & 0.9860 & 0.9263 & 0.9121\\
\hline
\end{tabular}
	\caption{Stationarity ratios of the data sets}
		\label{tab_stat_ratio}
\end{table}



\begin{figure}[t]
\begin{center}
     \subfloat[Mol\`ene weather data set]
     {\label{fig_ComparativeSaltPepperMoleneNMEResults}\includegraphics[height=3.6cm]{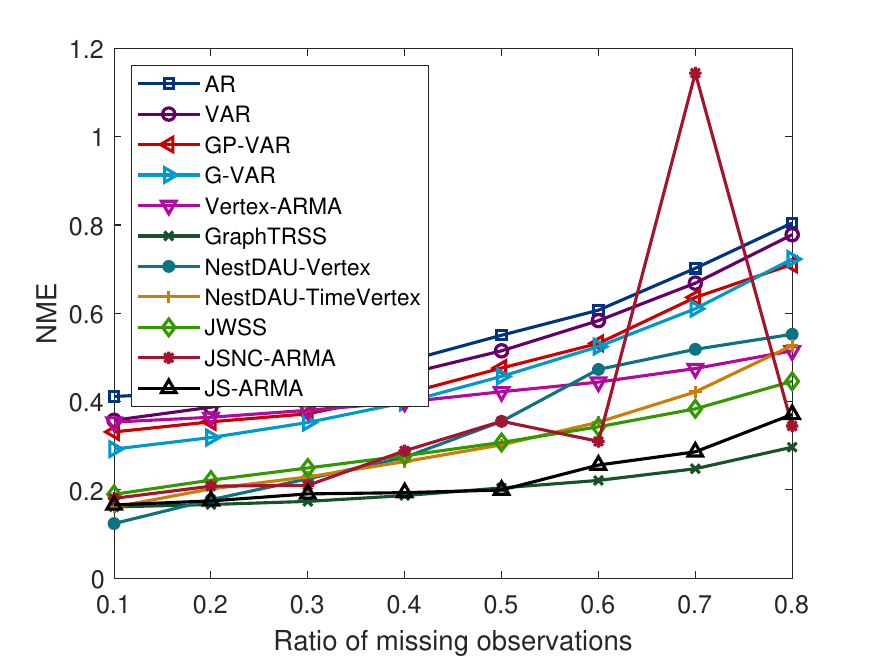}}
     \subfloat[COVID-19 data set]
     {\label{fig_ComparativeSaltPepperCovid19NMEResults}\includegraphics[height=3.6cm]{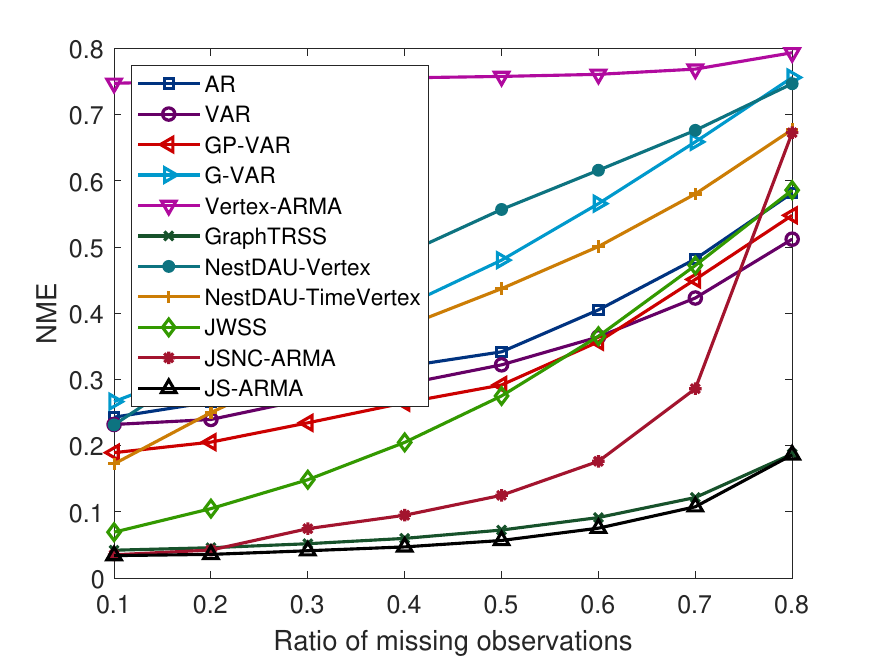}}\\
     \subfloat[NOAA weather data set]
     {\label{fig_ComparativeSaltPepperNOAANMEResults}\includegraphics[height=3.6cm]{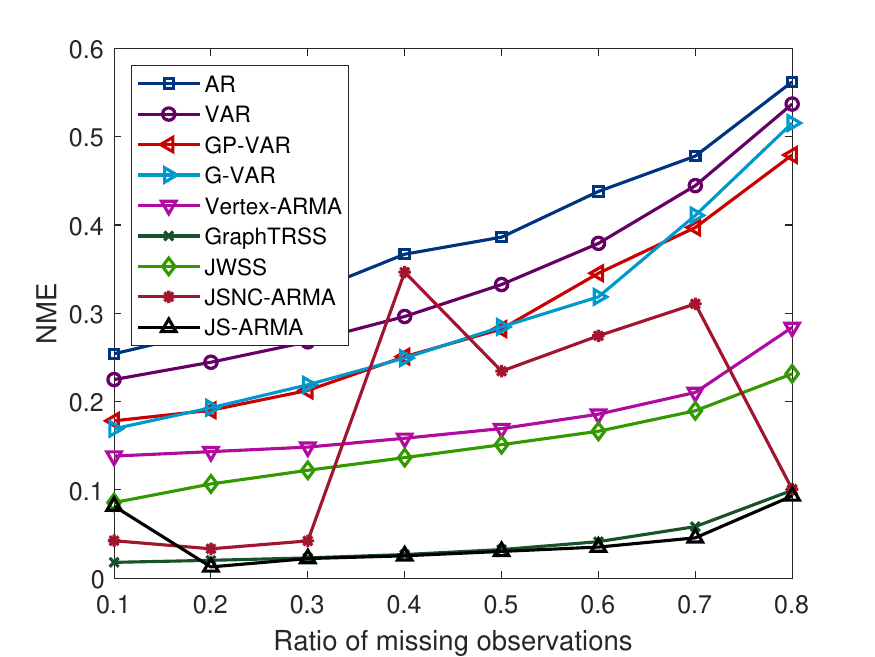}}
 \end{center}
	\caption{NME estimation errors for Scenario \ref{sco_salt_pepper}}
 \label{fig_nme_comparative_salt_pepper}
\end{figure}


\begin{figure}[th]
\begin{center}
     \subfloat[Mol\`ene weather data set]
     {\label{fig_ComparativeEntireGraphMoleneNMEResults}\includegraphics[height=3.4cm]{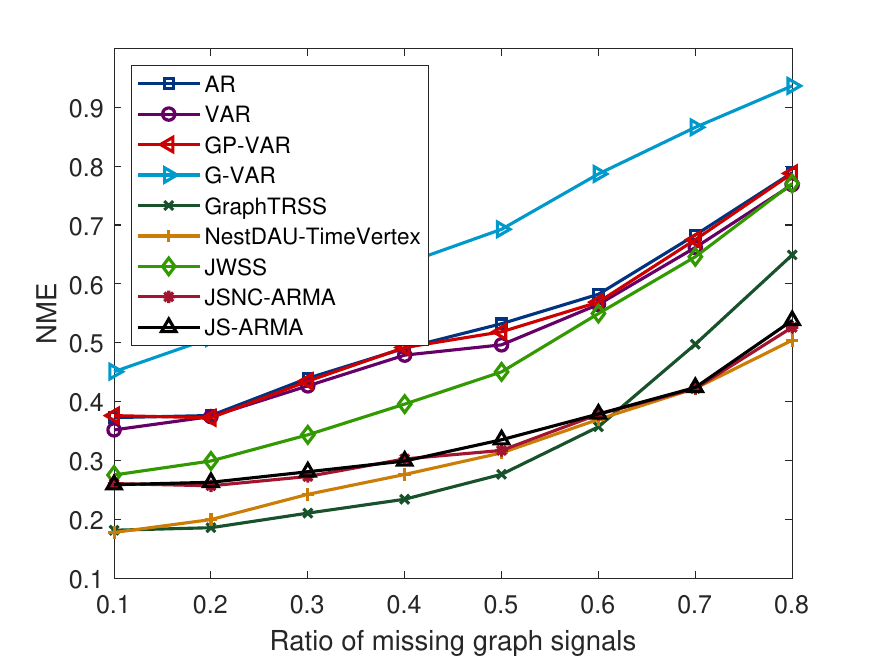}}
     \subfloat[COVID-19 data set]
     {\label{fig_ComparativeEntireGraphCovid19NMEResults}\includegraphics[height=3.4cm]{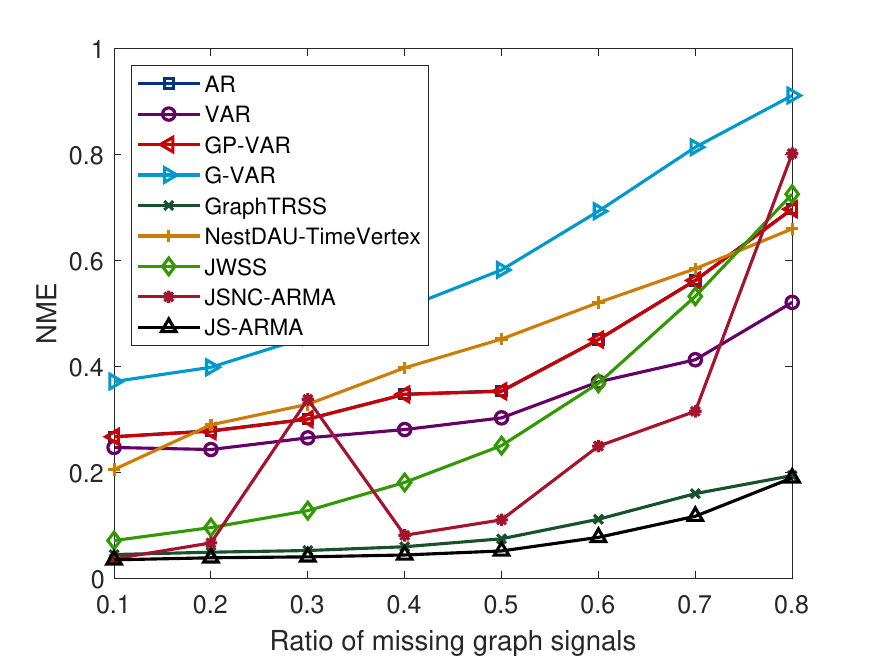}}\\
          \subfloat[NOAA weather data set]
     {\label{fig_ComparativeEntireGraphNOAANMEResults}\includegraphics[height=3.6cm]{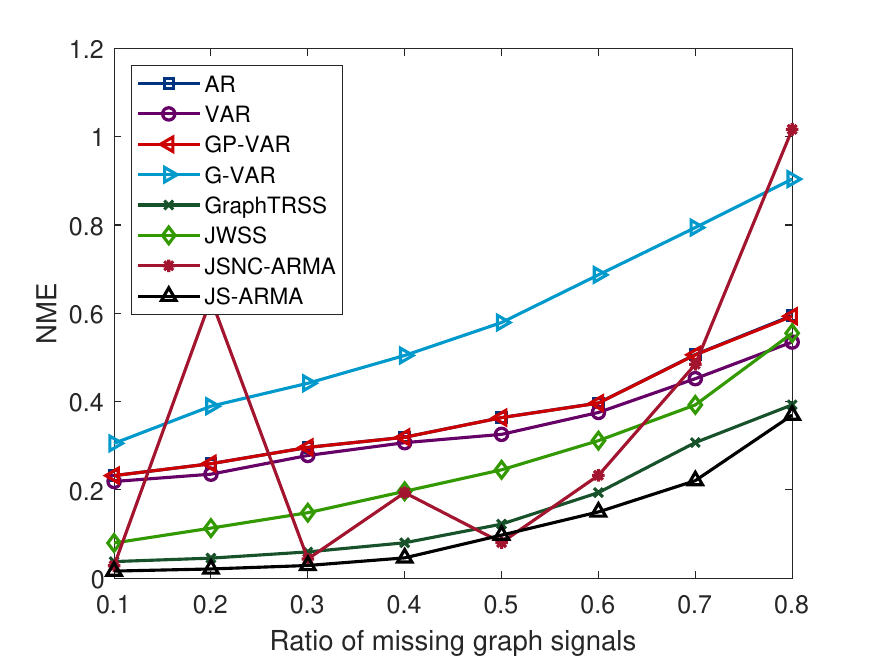}}
 \end{center}
	\caption{NME estimation errors for Scenario \ref{sco_entire_graph}}
 \label{fig_nme_comparative_entire_graph}
\end{figure}

\begin{figure}[th]
\begin{center}
     \subfloat[Mol\`ene weather data set]
     {\label{fig_ComparativeForecastingMoleneNMEResults}\includegraphics[height=3.4cm]{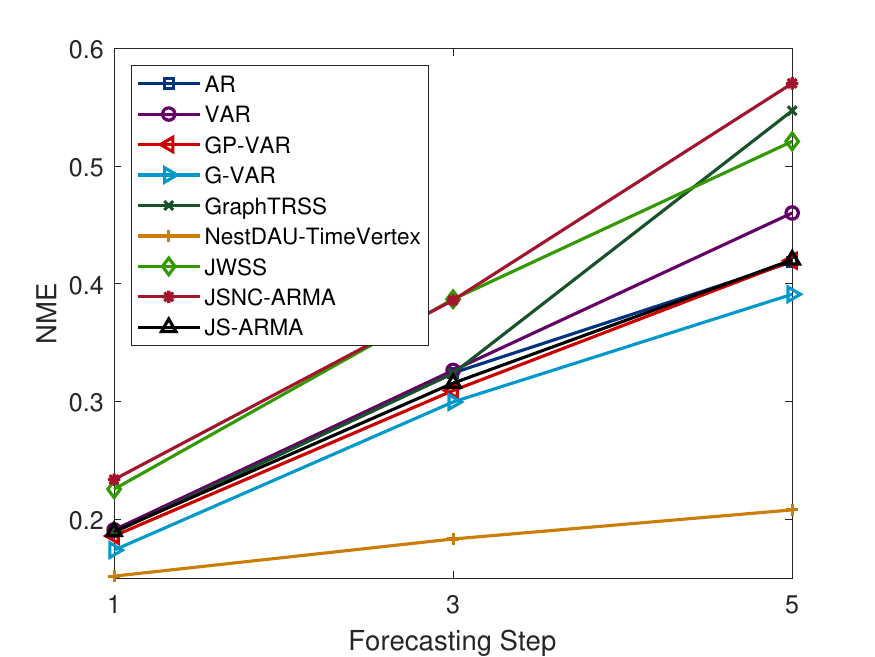}}
     \subfloat[COVID-19 data set]
     {\label{fig_ComparativeForecastingCovid19NMEResults}\includegraphics[height=3.4cm]{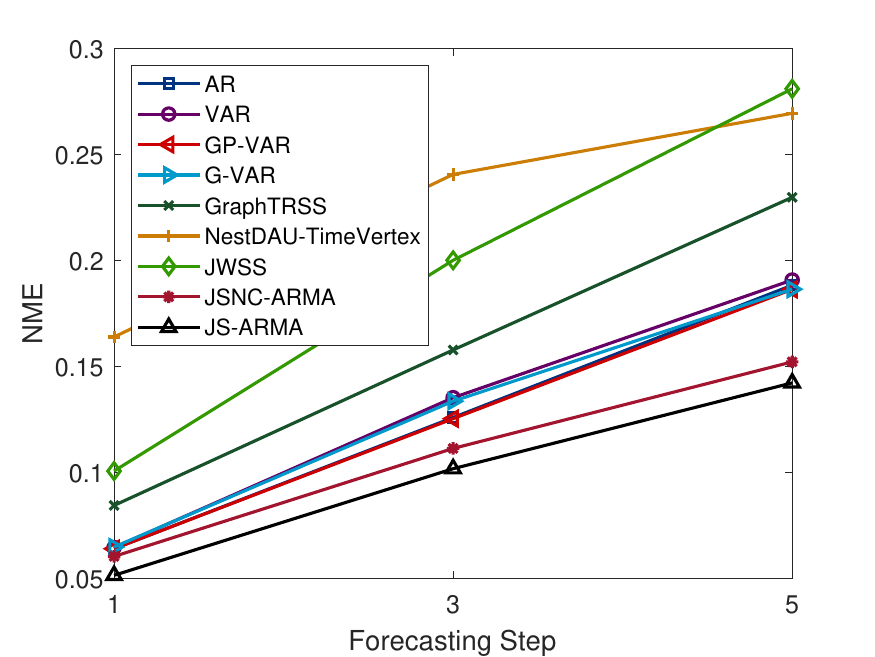}}\\
     \subfloat[NOAA weather data set]
     {\label{fig_ComparativeForecastingNOAANMEResults}\includegraphics[height=3.6cm]{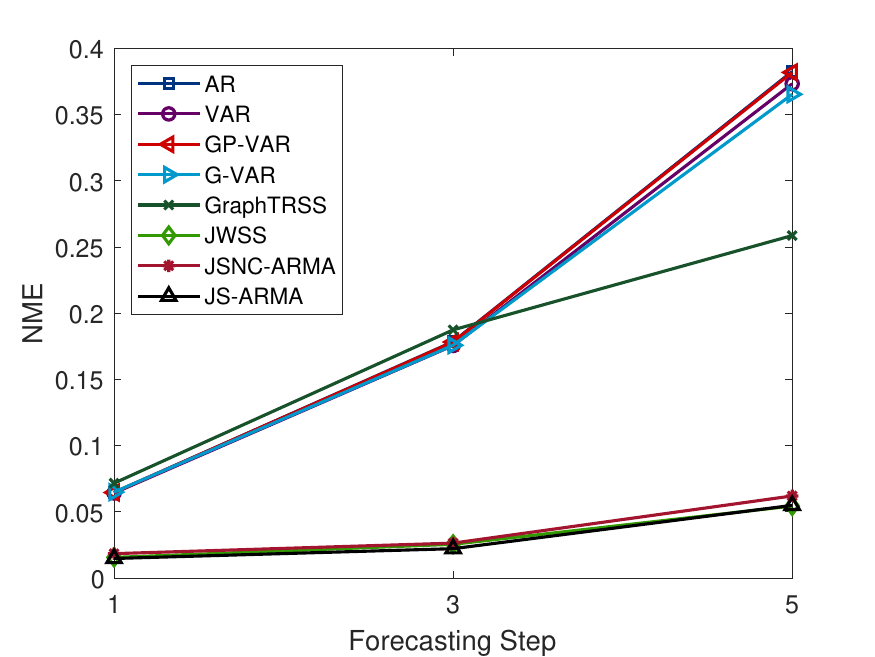}}
 \end{center}
	\caption{NME estimation errors for Scenario \ref{sco_forecasting} (Forecasting)}
 \label{fig_nme_comparative_forecasting}
\end{figure}

The performances of the algorithms are compared with respect to the normalized mean error (NME) of the estimates of the missing observations as defined in \eqref{eq_nme_defn}. The variation of the NME is plotted with respect to the ratio of missing observations in Fig.~\ref{fig_nme_comparative_salt_pepper}-\ref{fig_nme_comparative_entire_graph} for the signal interpolation problems in Scenarios \ref{sco_salt_pepper}-\ref{sco_entire_graph} and with respect to the forecasting time step $s$ in Fig.~\ref{fig_nme_comparative_forecasting} for Scenario \ref{sco_forecasting}. The compared methods have also been evaluated with respect to the  RMSE, MAPE, and the MAE metrics in Appendix E, which lead to similar conclusions.

For the signal interpolation problems in Scenarios \ref{sco_salt_pepper}-\ref{sco_entire_graph} reported in Figures \ref{fig_nme_comparative_salt_pepper}-\ref{fig_nme_comparative_entire_graph}, we observe that the proposed JS-ARMA method yields the best estimation performance among the methods relying on stochastic process models. Graph process models perform better than the VAR and AR models more often, which ignore the graph topology. An interesting exception to this occurs with the COVID-19 data set, where the AR and VAR methods outperform the graph-based G-VAR and Vertex-ARMA methods, which is coherent with the fact that the COVID-19 data set has very high time stationarity and weaker vertex or time-vertex stationarity. The proposed JS-ARMA method and the JWSS method, which employ the knowledge of the time-vertex joint spectrum of the process, perform better than G-VAR, GP-VAR, Vertex-ARMA, AR, and VAR, which do not exploit this information. Interestingly, this even holds for the COVID-19 data set with very high time stationarity and relatively low time-vertex stationarity, confirming that the joint time-vertex spectrum of a time-varying graph signal provides critical information about its characteristics that cannot be captured with vertex-only and time-only frequency analysis. The JSNC-ARMA method has fluctuating behavior, which stems from the difficulty of consistently obtaining an accurate solution due to the nonconvexity of its objective function \eqref{eq_opt_prob1}. The performance gap between JS-ARMA and JSNC-ARMA demonstrates the efficacy of the convex relaxations underlying the proposed algorithm. Regarding the non-stochastic methods, employing both the time-domain and the vertex-domain information,  NestDAU-TimeVertex is seen to perform better than NestDAU-Vertex in general. JS-ARMA often outperforms the NestDAU methods in the interpolation problems in Scenarios \ref{sco_salt_pepper}-\ref{sco_entire_graph}, NestDAU-TimeVertex performing slightly better only for the Mol\`ene data set in Scenario \ref{sco_entire_graph}. On the other hand, the GraphTRSS method is seen to be rather competitive with JS-ARMA. GraphTRSS is a spatio-temporal interpolation method whose objective function incorporates a Sobolev regularization term imposing that the time-derivative of the signal must vary smoothly over the graph \cite{GiraldoMGTB22}. The JS-ARMA and the GraphTRSS methods are essentially very different in nature: While the former learns a stochastic process model, the latter proposes a simple but elegant regularization idea without passing through a signal model. It remains as an interesting future direction to investigate whether and how the regularization technique in \cite{GiraldoMGTB22} can be coupled with the stochastic models learnt with the proposed JS-ARMA and similar approaches in order to push the signal estimation accuracy even further.

Lastly, for the forecasting problem in Scenario \ref{sco_forecasting},  for the Mol\`ene data set JS-ARMA performs similarly to the GP-VAR method which is specifically designed for forecasting problems, while NestDAU-TimeVertex provides the best forecasting performance in Fig.~\ref{fig_ComparativeForecastingMoleneNMEResults}. Meanwhile, the estimation error of NestDAU-TimeVertex is higher for the COVID-19 data set in Fig.~\ref{fig_ComparativeForecastingCovid19NMEResults}. The proposed JS-ARMA method is seen to well capture the strong time-stationarity behaviors of the COVID-19 and the NOAA data sets, yielding the best forecasting performance. An overall consideration of our comparative experiments suggests that the proposed JS-ARMA method is able to successfully combine the efficacy of parametric process models with the information of the time-vertex joint spectral characteristics of data, achieving state-of-the-art performance in the estimation of time-varying graph signals.



\section{Conclusion}
\label{sec_concl}
We have proposed a method for learning parametric stationary graph process models from time-vertex data sets. Our solution is based on fitting the parameters of a graph ARMA model to an initial rough estimate of the joint time-vertex spectrum of the process computed from possibly incomplete realizations of the process. The proposed method has been shown to achieve quite competitive performance for the inference of time-vertex signals, in comparison with reference approaches in the literature. The extension of the current study to time-varying graph structures or big network topologies are among the possible future directions of interest.


\appendices
\section*{Appendix A: Existence of the constants $\Tbnd$ and $\Kbnd$}

In this section, we discuss the existence of the geometric constants used in our sample complexity bounds. 
We first begin with the constant $\Tbnd$. We observe that the derivative of the JPSD  $h_{\cvect + t \unitvect}(\graphEigenvalue_n,\omega_\tau)$ at $t=0$ can be expressed in the form
\begin{equation}
\begin{split}
\frac{d}{dt} h_{\cvect + t \unitvect}(\graphEigenvalue_n,\omega_\tau) \bigg |_{t=0}  =( \rvect_{\cvect}(\graphEigenvalue_n,\omega_\tau))^\intercal  \unitvect 
\end{split}
\end{equation}
where the vector $\rvect_{\cvect} (\graphEigenvalue_n,\omega_\tau)\in \R^\dc $ is defined as
\begin{equation}
\begin{split}
  \frac{2}{| 1+ \avect^\intercal \,  \vvect_{n, \tau} |^4}
 \begin{bmatrix}
- |  \bvect^\intercal \uvect_{n, \tau} |^2 \, ( Re\{\vvect_{n, \tau}\} +  Re\{\vvect_{n, \tau} \vvect_{n, \tau}^H  \} \avect )  \\
| 1+ \avect^\intercal \vvect_{n, \tau}  |^2 Re\{\uvect_{n, \tau} \uvect_{n, \tau}^H \} \bvect
\end{bmatrix}
\end{split}
\end{equation}
for each frequency pair $(\graphEigenvalue_n,\omega_\tau)$. The tangent vector $\frac{d \, \hvect_{\cvect + t \unitvect}}{dt} \in \R^\NT $ at $t=0$ can then be expressed as
\begin{equation}
\begin{split}
\frac{d \, \hvect_{\cvect + t \unitvect}}{dt} \bigg |_{t=0} 
= \Rmat^\intercal  \, \unitvect
\end{split}
\end{equation}
where the matrix $\Rmat  \in \R^{\dc \times \NT}$ is defined as 
\begin{equation}
\begin{split}
\Rmat 
=[\rvect_{\cvect} (\graphEigenvalue_1,\omega_1) \  \rvect_{\cvect} (\graphEigenvalue_2,\omega_1) 
\ \dots \  \rvect_{\cvect} (\graphEigenvalue_N,\omega_T) ].
\end{split}
\end{equation}
Now, assuming that the length of the process $\xv$ is large enough to satisfy $NT > \dc = P(K+1)+(Q+1)(M+1)$, the matrix $\Rmat^\intercal \in \R^{\NT \times \dc}$ is a tall matrix. Hence,  the equation system $\Rmat^\intercal \,  \unitvect = 0$ is likely to be overdetermined in general and will not have an exact solution for unit-norm vectors with $\| \unitvect \| = 1$. This means that norms of the tangent vectors are positive,  thus we have $\big \|  \frac{d \, \hvect_{\cvect + t \unitvect}}{dt} |_{t=0} \big \|>0$ in general. Hence, provided that $NT > \dc $, the existence of a positive lower bound $\Tbnd $ on the tangent norms is a realistic and mild assumption.

Next, we discuss the existence of the curvature upper bound $\Kbnd$. Let us decompose the unit-norm parameter vector $\unitvect$ as $\unitvect=[\unitvect_a^\intercal  \ \unitvect_b^\intercal]^\intercal$ such that $\unitvect_a \in \R^{P(K+1)}$ and $\unitvect_b \in \R^{(Q+1)(M+1)}$. From \eqref{eq_defn_h_n_tau}, the JPSD at $\cvect + t \unitvect \in \csp$ can be written in the form
\begin{equation}
\begin{split}
h_{\cvect + t \unitvect}(\graphEigenvalue_n,\omega_\tau) = \frac{\beta(t)}{\alpha(t)}
\end{split}
\end{equation}
where $\alpha(t) \triangleq |  1 + (\avect + t \unitvect_a)^T \vvect_{n, \tau}  |^2$ and $\beta(t) \triangleq | (\bvect+ t \unitvect_b)^T \uvect_{n, \tau} |^2$. (The dependence of $\alpha(t) $ and $\beta(t)$ on $(\graphEigenvalue_n,\omega_\tau) $ is omitted from the notation for simplicity.) Since $\cvect + t \unitvect$ is taken to be in the bounded parameter space $\csp \in \R^\dc$, it is easy to show that the first- and second- order derivatives of $\alpha(t) $ and $\beta(t)$ with respect to $t$ are all bounded. Then with a simple inspection of the second derivative expression, we observe that an upper bound on $\left |   \frac{d^2 }{dt^2} \  h_{\cvect + t \unitvect}(\graphEigenvalue_n,\omega_\tau)    \right |$ exists, provided that  the denominator $| \alpha(t) |$ admits a positive lower bound on $\csp$. From our assumption that the spectrum is finite over $\csp$, it follows that $\alpha(t)$ must be nonzero at any $\cvect + t \unitvect \in \csp$. Since the parameter set $\csp$ is assumed to be compact, this implies
\begin{equation}
\begin{split}
\inf_{\cvect + t \unitvect \in \csp} | \alpha(t) | > 0.
\end{split}
\end{equation}
We thus conclude that for each frequency pair $(\graphEigenvalue_n,\omega_\tau) $, one can find a finite upper bound $\Kbnd_{n,\tau}$ on $\left |   \frac{d^2 }{dt^2} \  h_{\cvect + t \unitvect}(\graphEigenvalue_n,\omega_\tau)    \right |$, which indicates the existence of a finite global curvature upper bound $\Kbnd$ for the JPSD manifold $\Hc $.

\section*{Appendix B: Proof of Lemma \ref{lem_hard_bnd_hs_hg}}

\begin{proof}
First, we begin with developing a first-order approximation of the JPSD manifold $\Hc$. Fixing $n$ and $\tau$, let us regard $h_{\cvect + t \unitvect}(\graphEigenvalue_n,\omega_\tau) $ as a function of $t\in \R$. Taking $\cvect=\cs$, the Taylor expansion of $ h_{\cs + t \unitvect} (\graphEigenvalue_n,\omega_\tau) $ around $t=0$ can be written as
\begin{equation}
\label{eq_hcs_tu_expr1}
\begin{split}
 h_{\cs + t \unitvect} (\graphEigenvalue_n,\omega_\tau) 
 &=  h_{\cs} (\graphEigenvalue_n,\omega_\tau) 
 +  t  \, \left ( \frac{d}{d r}  h_{\cs + r \unitvect} (\graphEigenvalue_n,\omega_\tau)  \right )\bigg |_{r=0}  \\
 &+ \frac{t^2}{2}  \left ( \frac{d^2}{d r^2}  h_{\cs + r \unitvect} (\graphEigenvalue_n,\omega_\tau)  \right )\bigg |_{r=r_0}  
\end{split}
\end{equation}
for some $r_0 \in [0, t]$. Using the bound in \eqref{eq_defn_Kbnd_ntau}, we get
\begin{equation}
\begin{split}
&\bigg | h_{\cs + t \unitvect} (\graphEigenvalue_n,\omega_\tau) 
 - \bigg(  
h_{\cs} (\graphEigenvalue_n,\omega_\tau) \\
 &+  t  \, \left ( \frac{d}{d r}  h_{\cs + r \unitvect} (\graphEigenvalue_n,\omega_\tau)  \right )\bigg |_{r=0} 
 \bigg )
 \bigg |  
 \leq 
 \frac{t^2}{2} \  \Kbnd_{n,\tau}
\end{split}
\end{equation}
for any unit-norm $\unitvect $. Now if we take $t= \| \cg - \cs \|$ and $\unitvect =(\cg - \cs) / \| \cg - \cs \|$ in the above equation, we observe that the term $h_{\cs + t \unitvect}$ becomes equal to the true JPSD $h_\cg$, while the term 
\begin{equation}
\label{eq_hgapp_entries}
\begin{split}
h_{\cs} (\graphEigenvalue_n,\omega_\tau) +  t  \, \left ( \frac{d}{d r}  h_{\cs + r \unitvect} (\graphEigenvalue_n,\omega_\tau)  \right )\bigg |_{r=0} 
\end{split}
\end{equation}
can be regarded as a first-order approximation of $h_\cg = h_{\cs + t \unitvect}$ computed around $h_\cs$. Let us denote the first-order approximation of the vectorized JPSD $\hg \in \R^\NT $ as $\hgapp \in \R^\NT$, which is a vector with entries given in \eqref{eq_hgapp_entries}. Recalling the definition of the curvature upper bound in \eqref{eq_defn_Kbnd}, we get
\begin{equation}
\label{eq_bdn_curv_devi}
\begin{split}
\|  \hg - \hgapp \| \leq  \frac{\Kbnd}{2} \ \| \cg - \cs \|^2 .
\end{split}
\end{equation}

Next, we observe from \eqref{eq_defn_ab_star} that since the manifold point $\hs \in \Hc$ is the minimizer of the distance to the initial JPSD estimate $\hinit$ over the manifold $\Hc$, the error vector $\hinit - \hs$ must be orthogonal to any tangent to the manifold at $\hs \in \Hc$. Since the vector $\hgapp - \hs $ is tangent to the manifold at point $\hs$, it is orthogonal to $\hinit - \hs$, from which we get
\begin{equation}
\label{eq_dev_hinit_hgapp1}
\begin{split}
\| \hinit - \hgapp \|^2 = \|  \hinit - \hs \|^2 + \| \hs - \hgapp \|^2.
\end{split}
\end{equation}
Meanwhile, the first term in the above expression can be bounded as
\begin{equation}
\label{eq_dev_hinit_hgapp2}
\begin{split}
\| \hinit - \hgapp \| &= \| \hinit - \hg + \hg - \hgapp \| 
\leq \| \hinit - \hg \|   \\
&+ \|  \hg - \hgapp \| \leq \| \eh \| + \frac{\Kbnd}{2} \ \| \cg - \cs \|^2
\end{split}
\end{equation}
following the bound in \eqref{eq_bdn_curv_devi} and the definition of the initial estimation error $\eh$. From \eqref{eq_dev_hinit_hgapp1} and \eqref{eq_dev_hinit_hgapp2}, we get
\begin{equation}
\label{eq_dev_hs_hgapp}
\begin{split}
\| \hs - \hgapp \|^2 &=  \| \hinit - \hgapp \|^2 -   \|  \hinit - \hs \|^2 \\
& \leq  
\left(  \| \eh \| + \frac{\Kbnd}{2} \ \| \cg - \cs \|^2  \right)^2 -  \|  \hinit - \hs \|^2.
\end{split}
\end{equation}
We can then obtain
\begin{equation}
\label{eq_pre_bnd_hs_hg}
\begin{split}
 \| \hs - \hg \|  & \leq  \|  \hs - \hgapp \|  + \| \hgapp - \hg \| \\
& \leq 
 \bigg( 
\left(  \| \eh \| + \frac{\Kbnd}{2}  \| \cg - \cs \|^2  \right)^2 -  \|  \hinit - \hs \|^2
\bigg)^{1/2} \\
& + 
\frac{\Kbnd}{2}  \| \cg - \cs \|^2
\end{split}
\end{equation}
where the second equality follows from \eqref{eq_dev_hs_hgapp} and \eqref{eq_bdn_curv_devi}.

Next, we recall that by taking $t= \| \cg - \cs \|$  in \eqref{eq_hgapp_entries},
\begin{equation}
\begin{split}
\hgapp - \hs  = \| \cg - \cs  \|  \  \left( \frac{d}{d r}  \hvect_{\cs + r \unitvect}  \right) \bigg |_{r=0} 
\end{split}
\end{equation}
where $\frac{d \, \hvect_{\cs + r \unitvect}}{dr} |_{r=0} \in \R^\NT$ is the tangent to the manifold $\Hc$ at point $\hs$ along direction $\unitvect =(\cg - \cs) / \| \cg - \cs \|$. This gives
\begin{equation}
\label{eq_pre_bnd_cg_cs}
\begin{split}
\| \hgapp - \hs  \|^2 &= \| \cg - \cs  \|^2   \   \left \| \frac{d \, \hvect_{\cs + r \unitvect}}{d r}   \bigg |_{r=0} \right \|^2 \\
&\leq 
\left(  \| \eh \| + \frac{\Kbnd}{2} \ \| \cg - \cs \|^2  \right)^2 -  \|  \hinit - \hs \|^2
\end{split}
\end{equation}
where the inequality follows from \eqref{eq_dev_hs_hgapp}. Meanwhile, due to the lower bound $\Tbnd$ on the tangent norms in \eqref{eq_defn_Tbnd}, we have $ \| \frac{d \, \hvect_{\cs + r \unitvect}}{dr} |_{r=0}  \| \geq \Tbnd$, which gives from \eqref{eq_pre_bnd_cg_cs}
\begin{equation}
\begin{split}
\| \cg - \cs  \|^2  \leq \frac{1}{\Tbnd^2}
 \left(
\left(  \| \eh \| + \frac{\Kbnd}{2} \ \| \cg - \cs \|^2  \right)^2 -  \|  \hinit - \hs \|^2
\right).
\end{split}
\end{equation}
Using this bound in \eqref{eq_pre_bnd_hs_hg}, we get
\begin{equation}
\begin{split}
 & \| \hs - \hg \|  \leq  \bigg( 
\left(  \| \eh \| + \frac{\Kbnd}{2}  \| \cg - \cs \|^2  \right)^2 -  \|  \hinit - \hs \|^2
\bigg)^{1/2} \\
& + 
\frac{\Kbnd}{2 \Tbnd^2} 
\left(
 \left(  \| \eh \| + \frac{\Kbnd}{2} \ \| \cg - \cs \|^2  \right)^2 -  \|  \hinit - \hs \|^2
\right)
\end{split}
\end{equation}
which gives the inequality stated in the lemma.
\end{proof}

\section*{Appendix C: Proof of Theorem \ref{thm_jpsd_error_conv_rate}}

\begin{proof}
We begin with studying the rate of decrease of the initial JPSD estimation error $\eh = \hinit - \hg$ as the number of realizations $L$ increases. Denoting the true and unknown JPSD of the process as $h_\cg ( \graphEigenvalue_n, \omega_\tau)$, and the corresponding matrix form of the JPSD as $h_\cg (\graphEigenvalueMat_{\graph}, \timeEigenvalueMat)$, we have
\begin{equation}
\label{eq_form_e2_v1}
\begin{split}
&E[\, \| \eh \|^2 ] = E[ \, \| \hinit - \hg \|^2  ]
\leq  E[ \, \| \tilde h(\graphEigenvalueMat_{\graph}, \timeEigenvalueMat) - h_\cg (\graphEigenvalueMat_{\graph}, \timeEigenvalueMat) \|_F^2  ] \\
&=E[ \, \| \jointEigenvectorMat^H \tilde \covMat_{\xv} \jointEigenvectorMat  -  \jointEigenvectorMat ^H \covMat_{\xv} \jointEigenvectorMat   \|_F^2  ]
= E[ \, \|  \tilde \covMat_{\xv}  -  \covMat_{\xv}    \|_F^2  ] \\
&= E[ \, \|  L^{-1} \Wis -    L^{-1}  E[\Wis]    \|_F^2 ]
= \frac{1}{L^2}  E[ \, \|  \Wis -    E[\Wis]    \|_F^2 ]\\
&= \frac{1}{L^2}  \sum_{i=1}^{\NT}  \sum_{j=1}^{\NT} \text{Var}(\Wis_{ij} )
\end{split}
\end{equation}
where we define $\Wis =  \sum_{l=1}^L \xv^l (\xv^l)^\intercal$ and refer to the true covariance matrix of the process as $\covMat_{\xv}$. Since $\xv$ is assumed to be a Gaussian process in our study, the matrix $\Wis $ has a Wishart distribution with $L$ degrees of freedom, and its expectation is given by $E[\Wis]=L\covMat_{\xv}$. From \eqref{eq_form_e2_v1} and the variance of the Wishart distribution, we obtain
\begin{equation}
\begin{split}
E[\, \| \eh \|^2 ] & \leq \frac{1}{L^2}  \sum_{i=1}^{\NT}  \sum_{j=1}^{\NT} L \left( (\covMat_{\xv})_{ij}^2 + (\covMat_{\xv})_{ii} (\covMat_{\xv})_{jj} \right) \\
&= \frac{1}{L} \left( \| \covMat_{\xv} \|_F^2 + \tr(\covMat_{\xv} )^2  \right)
=\frac{1}{L} \Cg
\end{split}
\end{equation}
where $\Cg \triangleq \| \covMat_{\xv} \|_F^2 + \tr(\covMat_{\xv} )^2$ is a constant depending on the model parameters $\cg$. From Markov's equality for any $\epsilon>0$ we have
\begin{equation}
\label{eq_Pe2_v1}
\begin{split}
P(\|  \eh\|^2 \geq \epsilon^2) \leq \frac{E[ \, \|  \eh\|^2 ]}{\epsilon^2} 
\leq \frac{\Cg}{ L \, \epsilon^2}.
\end{split}
\end{equation}
Here the constant $\Cg$ grows at rate $O(N^2 T^2)$ with the dimension $NT$ of the process. As for its dependence on the dimension $d$ of the parameter space, we observe from \eqref{eq_defn_h_n_tau} that $h_\cg (\lambda_n, \omega_\tau)$ grows at a rate bounded by $O(\dc ^2)$, which implies $\| \Sigx \|_F^2 = \| \jointEigenvectorMat \, h_\cg(\graphEigenvalueMat_\graph , \timeEigenvalueMat) \, \jointEigenvectorMat ^H \|_F^2 = \| h_\cg(\graphEigenvalueMat_\graph , \timeEigenvalueMat) \|_F^2 = O(\dc ^4)$. Hence, we conclude that $\Cg =O(N^2 T^2 \dc ^4)$.  Defining $\delta=\frac{\Cg}{ L \, \epsilon^2}$ in \eqref{eq_Pe2_v1}, we get that with probability at least $1-\delta$,
\begin{equation}
\label{eq_rate_conv_eh}
\begin{split}
\| \eh\| \leq \epsilon = \sqrt{ \frac{\Cg}{L \delta} }= O \left(\sqrt{ \frac{ N^2 T^2 \dc ^4}{L\delta}  } \right).
\end{split}
\end{equation}
Hence, we have shown that the initial JPSD estimation error $\| \eh\| $ decreases at rate $O(\sqrt{N^2 T^2 \dc^4 / (L\delta)})$ as $L$ increases. In the sequel, we employ this result to study the rate of convergence of the JPSD estimation error $\|  \hs - \hg \| $ in \eqref{eq_jspd_error_lemma}. We first observe that as $\hs$ is the projection of $\hinit$ onto $\Hc$, we have $\| \hinit - \hs\| \leq   \| \hinit - \hg  \|  = \| \eh \|  $, which implies from \eqref{eq_rate_conv_eh}
\begin{equation}
\begin{split}
\label{eq_rate_conv_hinit_hs}
\| \hinit - \hs\|  = O \left(\sqrt{ \frac{ N^2 T^2 \dc^4}{L\delta}  } \right).
\end{split}
\end{equation}
Then, it remains to determine the rate of convergence of the estimation error $\| \cs - \cg \|$ of the process model parameters. We recall from the proof of Lemma \ref{lem_hard_bnd_hs_hg} that taking  $t= \| \cg - \cs \|$ and $\unitvect=(\cg - \cs) / \| \cg - \cs \|$ in \eqref{eq_hcs_tu_expr1}, we have 
\begin{equation}
\label{eq_hcg_expr2}
\begin{split}
 \hvect_{\cg} 
 &=  \hvect_{\cs} 
 +  \| \cg - \cs \|  \, \left ( \frac{d}{d r}  \hvect_{\cs + r \unitvect}   \right )\bigg |_{r=0}  \\
 &+ \frac{1}{2} \| \cg - \cs \|^2   \left ( \frac{d^2}{d r^2}  \hvect_{\cs + r \unitvect}   \right )\bigg |_{r=r_0}  
\end{split}
\end{equation}
where the tangent vector $\frac{d \, \hvect_{\cs + r \unitvect}}{dr} |_{r=0} \in \R^\NT$ and the curvature vector $\frac{d^2 \, \hvect_{\cs + r \unitvect}}{dr^2} |_{r=r_0} \in \R^\NT$ consist of the first- and the second-order derivatives to the manifold $\Hc$. From \eqref{eq_hcg_expr2} we get
\begin{equation}
\begin{split}
& \| \cg - \cs \|  \left \| \frac{d}{d r}  \hvect_{\cs + r \unitvect}   \bigg |_{r=0}  \right \| \\
& =  \left \|
  \hg - \hs - \frac{1}{2} \| \cg - \cs \|^2   \left ( \frac{d^2}{d r^2}  \hvect_{\cs + r \unitvect}   \right )\bigg |_{r=r_0}  
 \right \| \\
 & \leq  
\|  \hs -    \hg  \|  +  \frac{1}{2}  \| \cg - \cs \|^2 
\left \| \frac{d^2}{d r^2}  \hvect_{\cs + r \unitvect}   \bigg |_{r=r_0}  \right \|.
\end{split}
\end{equation}
Using the bounds \eqref{eq_defn_Tbnd} and \eqref{eq_defn_Kbnd} on the norms of the tangent and curvature vectors, we then obtain
\begin{equation}
\label{eq_cg_cs_ito_h}
\begin{split}
 \| \cg - \cs \|   \leq
 \frac{1}{\Tbnd}  \|  \hs -    \hg  \|  
 + \frac{\Kbnd}{2 \Tbnd}   \| \cg - \cs \|^2.  
\end{split}
\end{equation}
Now defining in \eqref{eq_jspd_error_lemma}
\begin{equation}
\label{eq_gh_defn}
\begin{split}
\gh 
&\triangleq 
 \left( \| \eh \| + \frac{\Kbnd}{2} \, \| \cg - \cs \|^2 \right)^2 
 - \| \hinit - \hs \|^2 \\
 &= \| \eh \|^2 + \Kbnd \, \| \eh \| \,  \| \cg - \cs \|^2 + \frac{\Kbnd^2}{4} \| \cg - \cs \|^4
 -  \| \hinit - \hs \|^2 
\end{split}
\end{equation}
and recalling the convergence rates \eqref{eq_rate_conv_eh} and \eqref{eq_rate_conv_hinit_hs}, we get
\begin{equation}
\label{eq_gh_O_Ldelta}
\begin{split}
\gh &\leq  O \left( \frac{ N^2 T^2 \dc^4}{L\delta}  \right) 
+ \Kbnd  \| \cg - \cs \|^2 \, O \left(\sqrt{ \frac{ N^2 T^2 \dc^4}{L\delta}  } \right) \\
&+\frac{\Kbnd^2}{4} \| \cg - \cs \|^4.
\end{split}
\end{equation}
From Lemma \ref{lem_hard_bnd_hs_hg}, we recall that 
\begin{equation}
\begin{split}
\|  \hs - \hg \|  \leq  
 \sqrt{\gh}
 +
 \frac{\Kbnd}{2 \Tbnd^2}  \gh
\end{split}
\end{equation}
which gives from \eqref{eq_cg_cs_ito_h}
\begin{equation}
\label{eq_cg_cs_ito_gh}
\begin{split}
 \| \cg - \cs \|   \leq 
 \frac{1}{\Tbnd}  \left( \sqrt{\gh} + \frac{\Kbnd}{2 \Tbnd^2}  \gh \right)  
 + \frac{\Kbnd}{2 \Tbnd}   \| \cg - \cs \|^2.  
\end{split}
\end{equation}
In order to determine the rate of convergence of  $\| \cg - \cs \| $, we make the following observations: First, as the number of realizations $L$ increases, we can ignore the effect of the term $\gh \, \Kbnd/ (2 \Tbnd^2)  $ in \eqref{eq_cg_cs_ito_gh}, since the term $\sqrt{\gh} $ will converge at a slower rate than $\gh$. Combining this observation with the bound in \eqref{eq_gh_O_Ldelta}, we can rewrite \eqref{eq_cg_cs_ito_gh} as
\begin{equation}
\label{eq_cg_cs_Ldelta_bnd1}
\begin{split}
 \| \cg - \cs \|  & \leq 
 c_1 \bigg( 
 O \left( \frac{ N^2 T^2 \dc^4 }{L\delta}  \right)  
 +  \| \cg - \cs \|^2  \\  O \left(\sqrt{ \frac{ N^2 T^2 \dc^4}{L\delta}  } \right) 
&+c_2 \, \| \cg - \cs \|^4
\bigg)^{1/2}
+ c_3 \, \| \cg - \cs \|^2
\end{split}
\end{equation}
for some constants $c_i$'s. We thus notice that $ \| \cg - \cs \|$ can not converge at a rate slower than $O(\sqrt{N^2 T^2 \dc^4 / ( L\delta) })$; otherwise the inequality would be violated for large $L$. We thus get 
\begin{equation}
\label{eq_cg_cs_ito_Ldelta}
\begin{split}
 \| \cg - \cs \| = O \left(\sqrt{ \frac{ N^2 T^2 \dc^4}{L\delta}  } \right).
\end{split}
\end{equation}
Using the convergence rates \eqref{eq_rate_conv_eh}, \eqref{eq_rate_conv_hinit_hs}, and \eqref{eq_cg_cs_ito_Ldelta} in \eqref{eq_gh_O_Ldelta}, one can find the rate of convergence of the term $\gh$. It can be verified that the curvature parameter $\Kbnd$ increases at rate $O(\sqrt{NT})$ as the dimensions $N$ and $T$ grow. Assuming that the number of realizations $L$ is sufficiently large to satisfy the rate $L=\Omega((NT)^{5/2} \dc^4)$ at large dimensions, one gets $\gh=O(N^2 T^2 \dc^4 /(L\delta))$, which finally gives from \eqref{eq_jspd_error_lemma}
\begin{equation*}
\begin{split}
\|  \hs - \hg \|  =  O \left(\sqrt{ \frac{ N^2 T^2 d^4}{L\delta}  } \right).
\end{split}
\end{equation*}
\end{proof}

\section*{Appendix D: Proof of Theorem \ref{thm_lmmse_error_conv_rate}}

\begin{proof}
From the relation \eqref{eq_sigma_from_jpsd} between the covariance matrix and the JPSD estimates, we first observe that
\begin{equation}
\label{eq_rate_sigxh_sigx}
\begin{split}
& \| \Sigxh - \Sigx \|_F^2 = \| \jointEigenvectorMat \, h_\cs (\graphEigenvalueMat_\graph , \timeEigenvalueMat) \, \jointEigenvectorMat ^H  
-
\jointEigenvectorMat \, h_\cg (\graphEigenvalueMat_\graph , \timeEigenvalueMat) \, \jointEigenvectorMat ^H
  \|_F^2 \\
  &
  =\|  h_\cs(\graphEigenvalueMat_\graph , \timeEigenvalueMat)   
-
 \,  h_\cg (\graphEigenvalueMat_\graph , \timeEigenvalueMat)   \|_F^2
 = \| \hs - \hg \|^2 \\
 &= O \left( \frac{N^2 T^2 \dc^4}{L \delta} \right)
\end{split}
\end{equation}
with probability at least $1-\delta$, where the last inequality follows from Theorem \ref{thm_jpsd_error_conv_rate}.

The deviation between the estimates $\ztests$ and $\ztestg$ can be bounded as
\begin{equation}
\label{eq_dev_zest_v1}
\begin{split}
& \| \ztests - \ztestg \| = 
\|  \Sigzytesth \, (\Sigytesth)^{-1}  \ytest
- 
\Sigzytest \, (\Sigytest)^{-1} \ytest
\| \\
&\leq 
\|  
\Sigzytesth \, (\Sigytesth)^{-1}  - \Sigzytest \, (\Sigytest)^{-1}
\| \,
\| \ytest \| \\
&= 
\|    
\Sigzytesth \, (\Sigytesth)^{-1}  - \delSigzytest \, (\Sigytesth)^{-1} + \delSigzytest \, (\Sigytesth)^{-1}  \\
& - \Sigzytest \, (\Sigytest)^{-1} \|
\, \| \ytest \| \\
&\leq
\|
\, \Sigzytest \, \big( (\Sigytesth)^{-1} - (\Sigytest)^{-1}  \big)+ \delSigzytest \, (\Sigytesth)^{-1}  
 \, \|
\  \| \ytest \| \\
& \leq
\|  \Sigzytest  \|  \  \left \| (\Sigytesth)^{-1} - (\Sigytest)^{-1}  \right \| \  \| \ytest \| 
 \\
&+ 
\|   \delSigzytest  \|  \   \|  (\Sigytesth)^{-1}  \|  \ \| \ytest \| 
\end{split}
\end{equation}
where we define $\delSigzytest  = \Sigzytesth  - \Sigzytest$ and $\delSigytest  = \Sigytesth  - \Sigytest$. 

The rest of the proof is based on studying the rates of convergence of the terms obtained in  \eqref{eq_dev_zest_v1}. We recall from the proof of Theorem \ref{thm_jpsd_error_conv_rate} that $\Cg = \| \covMat_{\xv} \|_F^2 + \tr(\covMat_{\xv} )^2= O(N^2 T^2 \dc^4)$, which implies $ \| \ytest \| = O(\sqrt{NT \dc^2})$.  We observe that $ \| \ytest \| $ does not depend on the number of realizations $L$ in \eqref{eq_dev_zest_v1}. Meanwhile, although the norms $\| \Sigzytest \|$ and $\| \delSigzytest \|$ of the covariance terms  increase at rate $ O(NT \dc^2)$ as the dimensions increase, this effect is typically expected to be neutralized by the inverse covariance terms in product with them in \eqref{eq_dev_zest_v1}. Therefore, in the sequel, when analyzing the covariance terms in \eqref{eq_dev_zest_v1}, we focus only on their dependence on the number of realizations $L$ and do not consider their dependence on $N$, $T$ and $\dc$. 

Regarding their dependence on only $L$, the terms $\|  \Sigzytest  \| $ and $ \|  (\Sigytesth)^{-1}  \| $ in \eqref{eq_dev_zest_v1} are of $O(1)$. Next, we have
\begin{equation}
\label{eq_conv_rate_deltazy}
\begin{split}
\|   \delSigzytest  \|  
& \leq    \|   \delSigzytest  \|_F 
= \|   \Sigzytesth  - \Sigzytest  \|_F
\leq   \| \Sigxh - \Sigx \|_F  \\
& = 
O \left( \frac{1}{\sqrt{L \delta}} \right)
\end{split}
\end{equation}
with probability at least $1-\delta$ due to \eqref{eq_rate_sigxh_sigx}. Lastly, we study the term
\begin{equation}
\label{eq_sigyhinv_sigyinv_v1}
\begin{split}
& \| (\Sigytesth)^{-1} - (\Sigytest)^{-1}  \|
 =  \|   (\Sigytest + \delSigytest)^{-1}  -   (\Sigytest)^{-1}    \| .\\
 \end{split}
\end{equation}
From  Woodbury matrix identity, we have
\begin{equation}
\begin{split}
&(\Sigytest + \delSigytest)^{-1} \\
&=  (\Sigytest)^{-1} -  (\Sigytest)^{-1}   
\left(  
(\delSigytest)^{-1}+
(\Sigytest)^{-1} 
\right)^{-1}
   (\Sigytest)^{-1} .
\end{split}
\end{equation}
Using this in \eqref{eq_sigyhinv_sigyinv_v1} we get
\begin{equation}
\label{eq_invsigyh_invsigy_v2}
\begin{split}
& \| (\Sigytesth)^{-1} - (\Sigytest)^{-1}  \|  \\
&= 
\|  
 (\Sigytest)^{-1}   
\left(  
(\delSigytest)^{-1}+
(\Sigytest)^{-1} 
\right)^{-1}
   (\Sigytest)^{-1}
\|  \\
& = \left 
\| (\Sigytest)^{-1} 
\left(  
\delSigytest - \delSigytest   (\Sigytesth)^{-1}  \delSigytest
\right)
(\Sigytest)^{-1}   
\right  \|  \\
& \leq
\| (\Sigytest)^{-1} \|^2 \
\| \delSigytest \|  \  \| \eye -    (\Sigytesth)^{-1}  \delSigytest\|
\end{split}
\end{equation} 
where the second equality is obtained by using  the Woodbury matrix identity, this time for the matrices $(\delSigytest)^{-1}$ and $ (\Sigytest)^{-1}$. In \eqref{eq_invsigyh_invsigy_v2}, we observe that the terms $\| (\Sigytest)^{-1} \|^2$ and  $\| \eye -    (\Sigytesth)^{-1}  \delSigytest\|$ are of $O(1)$ as $L$ increases. Meanwhile, similarly to \eqref{eq_conv_rate_deltazy}, it can be shown that with probability at least $1-\delta$,
\begin{equation}
\begin{split}
\| \delSigytest \| = O \left( \frac{1}{\sqrt{L \delta}} \right)
\end{split}
\end{equation}
which gives from \eqref{eq_invsigyh_invsigy_v2}
\begin{equation}
\label{eq_conv_rate_invsigyh_invsigy}
\begin{split}
\| (\Sigytesth)^{-1} - (\Sigytest)^{-1}  \|  
= 
O \left( \frac{1}{\sqrt{L \delta}} \right).
\end{split}
\end{equation}
Finally, combining the results \eqref{eq_conv_rate_deltazy} and \eqref{eq_conv_rate_invsigyh_invsigy} in \eqref{eq_dev_zest_v1} and recalling that $ \| \ytest \| = O(\sqrt{NT \dc^2})$,  we conclude that with probability at least $1-\delta$,
\begin{equation}
\begin{split}
 \| \ztests - \ztestg \| = O \left(\sqrt{\frac{NT \dc^2}{L \delta}} \right)
\end{split}
\end{equation}
which finishes the proof. 
\end{proof}

\section*{Appendix E: Evaluation of compared methods with respect to RMSE, MAPE, and MAE metrics}

In this section, we present additional results complementary to the plots obtained with the NME error measure in Section \ref{ssec_exp_realdata}. We evaluate the estimation errors of the methods for the same comparative experiments as in Section \ref{ssec_exp_realdata}, by measuring the errors with respect to the Root Mean Square Error (RMSE), Mean Absolute Error (MAE), and the Mean Absolute Percentage Error (MAPE) metrics defined as follows.

\begin{equation}
\label{eq_error_metrics_defn}
\begin{split}
RMSE &= \left( \frac{  \|   {\bar \zSignal} - {\bar \zSignal}^\est  \|^2}{L_{\bar \zSignal}}  \right)^{1/2} \\
MAE&=  \frac{  \|   {\bar \zSignal} - {\bar \zSignal}^\est  \|_1}{L_{\bar \zSignal}}  \\
MAPE &= \frac{1}{L_{\bar \zSignal}} \sum_{i=1}^{L_{\bar \zSignal}}
\left |  \frac{  \bar \zSignal(i) - {\bar \zSignal}^\est(i) }{ \bar \zSignal(i) }
 \right |
\end{split}
\end{equation}
Here ${\bar \zSignal}$ denotes a concatenated vector consisting of all missing observations in the experiment, ${\bar \zSignal}^\est$ denotes its estimate, ${\bar \zSignal}(i)$ denotes the $i$-th entry of ${\bar \zSignal}$, and  $L_{\bar \zSignal}$ denotes its length. The RMSE, MAE, and MAPE\footnote{While we remove the mean of the data in our experiments, when reporting the MAPE errors, in order to avoid unbounded error values we calculate the error with respect to the nonzero-mean version of the data.} errors of the methods obtained in Scenarios \ref{sco_salt_pepper}-\ref{sco_forecasting}  in Section \ref{ssec_exp_realdata} are presented in Fig.~\ref{fig_rmse_comparative_salt_pepper} - Fig.~\ref{fig_mape_comparative_forecasting} below.


\begin{figure}[t]
\begin{center}
     \subfloat[Mol\`ene weather data set]
     {\label{fig_ComparativeSaltPepperMoleneRMSEResults}\includegraphics[height=3.4cm]{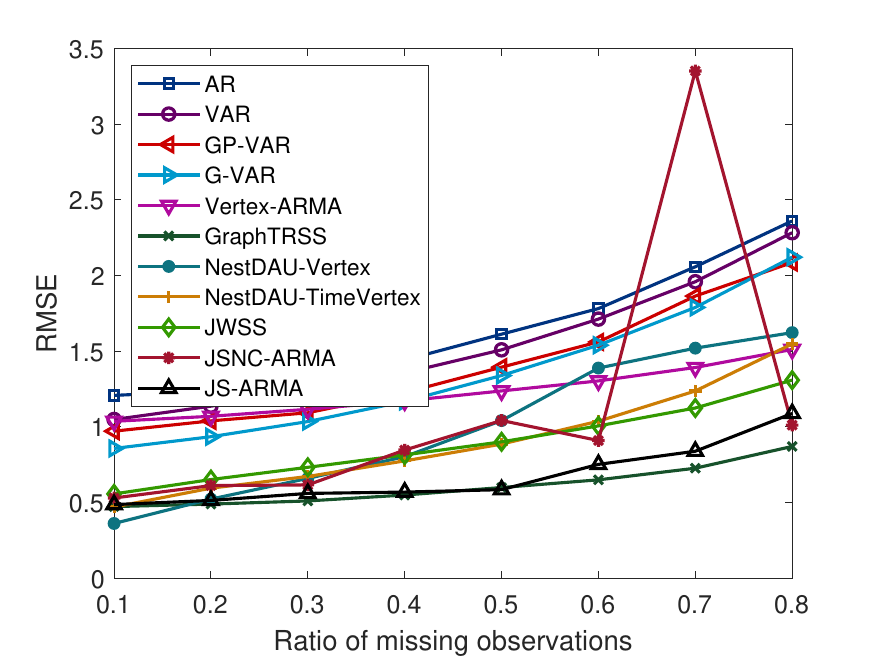}}
     \subfloat[COVID-19 data set]
     {\label{fig_ComparativeSaltPepperCovid19RMSEResults}\includegraphics[height=3.4cm]{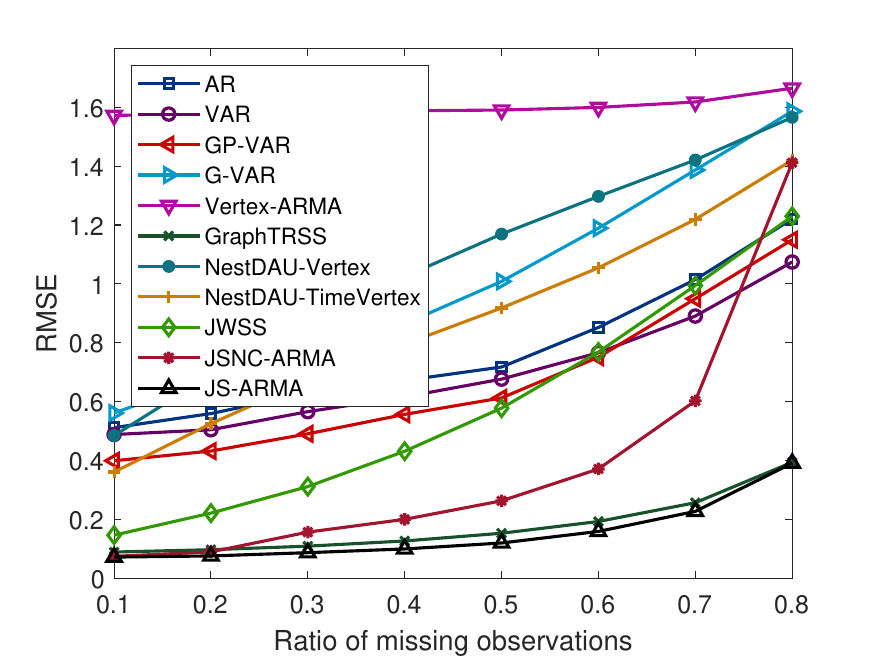}}\\
     \subfloat[NOAA weather data set]
     {\label{fig_ComparativeSaltPepperNOAARMSEResults}\includegraphics[height=3.4cm]{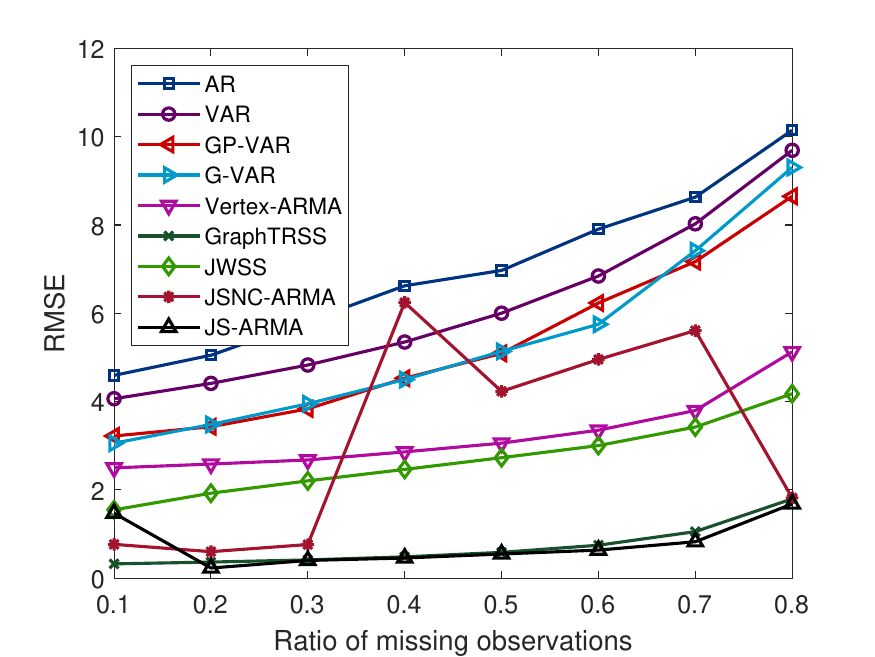}}
 \end{center}
	\caption{RMSE estimation errors for Scenario \ref{sco_salt_pepper} }
 \label{fig_rmse_comparative_salt_pepper}
\end{figure}

\begin{figure}[h]
\begin{center}
     \subfloat[Mol\`ene weather data set]
     {\label{fig_ComparativeSaltPepperMoleneMAEResults}\includegraphics[height=3.4cm]{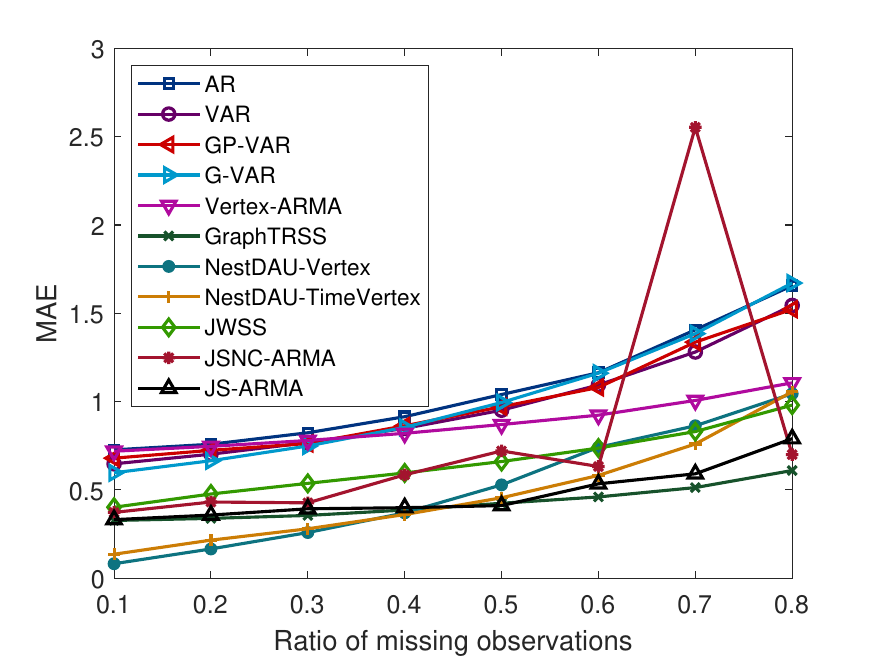}}
     \subfloat[COVID-19 data set]
     {\label{fig_ComparativeSaltPepperCovid19MAEResults}\includegraphics[height=3.4cm]{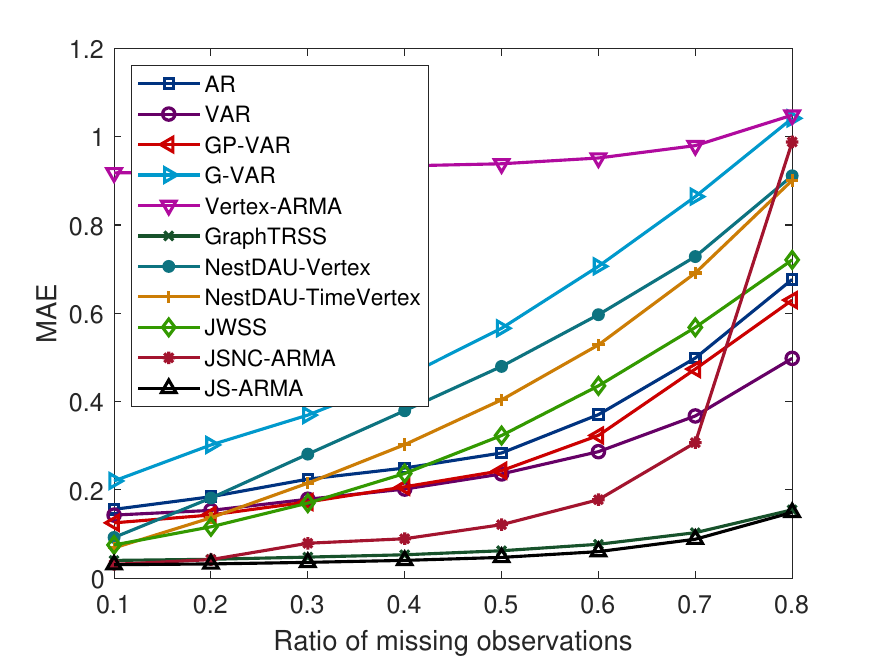}}\\
     \subfloat[NOAA weather data set]
     {\label{fig_ComparativeSaltPepperNOAAMAEResults}\includegraphics[height=3.4cm]{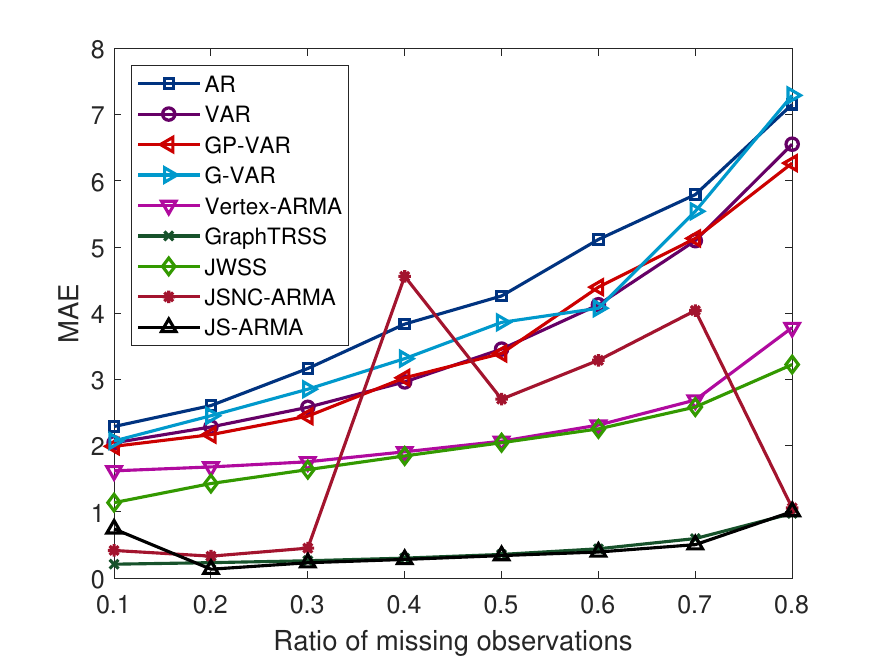}}
 \end{center}
	\caption{MAE estimation errors for Scenario \ref{sco_salt_pepper} }
 \label{fig_mae_comparative_salt_pepper}
\end{figure}

\begin{figure}[h]
\begin{center}
     \subfloat[Mol\`ene weather data set]
     {\label{fig_ComparativeSaltPepperMoleneMAPEResults}\includegraphics[height=3.4cm]{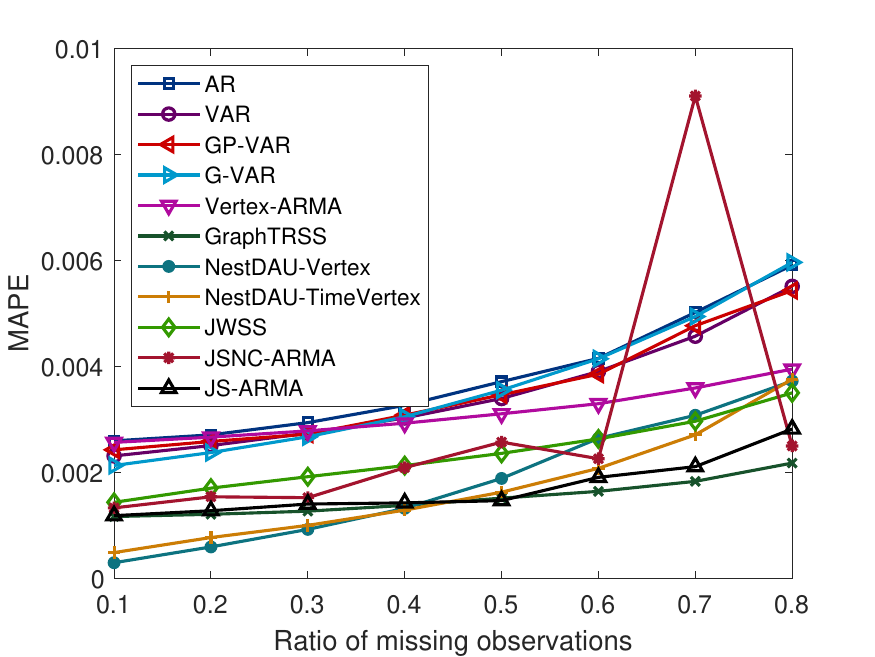}}
     \subfloat[COVID-19 data set]
     {\label{fig_ComparativeSaltPepperCovid19MAPEResults}\includegraphics[height=3.4cm]{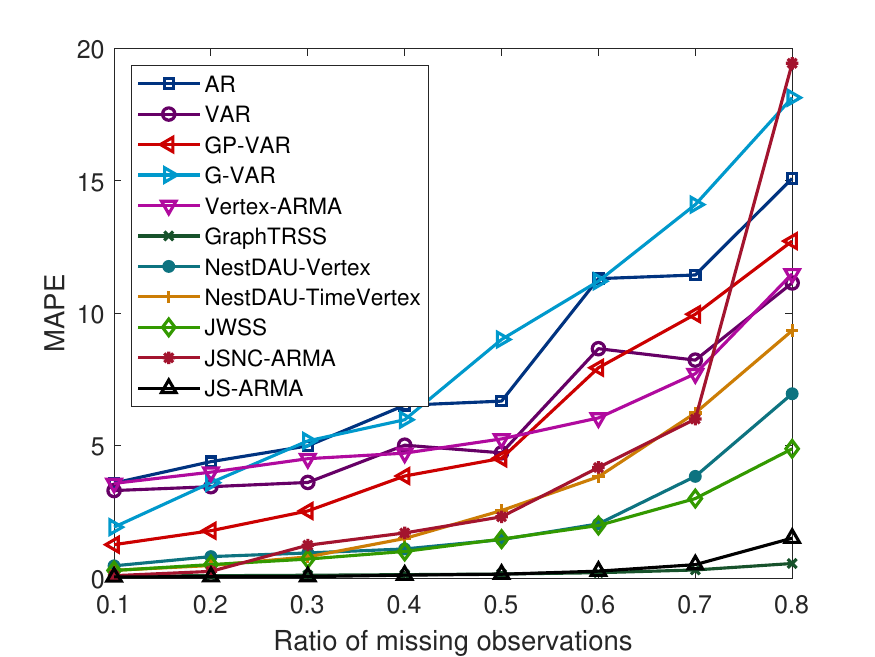}}\\
     \subfloat[NOAA weather data set]
     {\label{fig_ComparativeSaltPepperNOAAMAPEResults}\includegraphics[height=3.4cm]{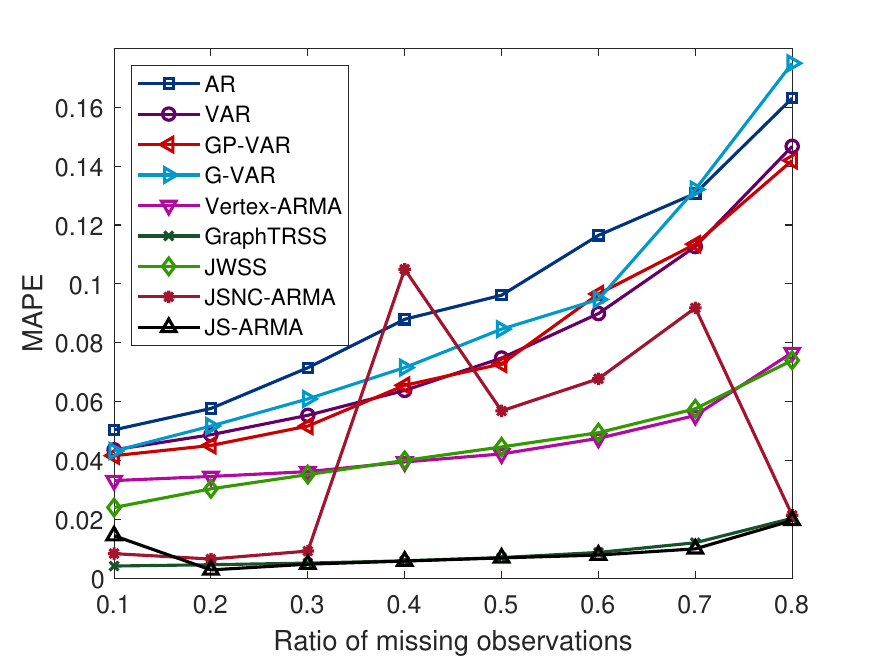}}
 \end{center}
	\caption{MAPE estimation errors for Scenario \ref{sco_salt_pepper} }
 \label{fig_mape_comparative_salt_pepper}
\end{figure}

\begin{figure}[t!]
\begin{center}
     \subfloat[Mol\`ene weather data set]
     {\label{fig_ComparativeEntireGraphMoleneRMSEResults}\includegraphics[height=3.4cm]{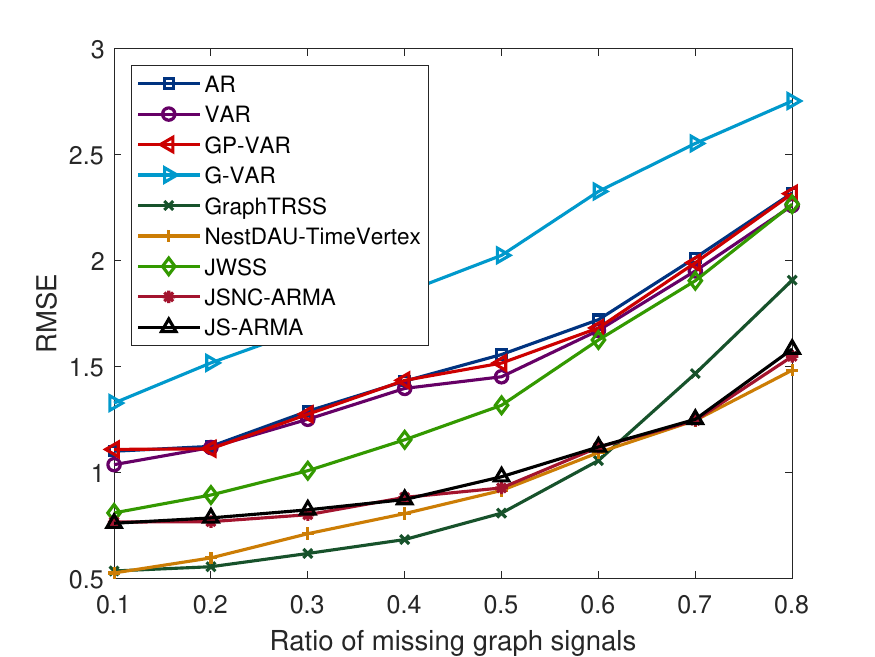}}
     \subfloat[COVID-19 data set]
     {\label{fig_ComparativeEntireGraphCovid19RMSEResults}\includegraphics[height=3.4cm]{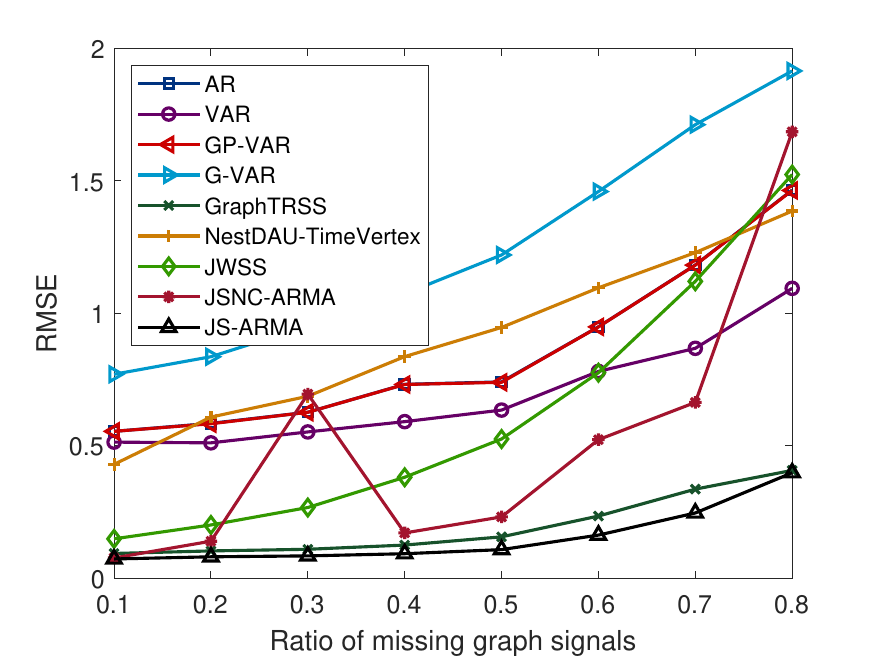}}\\
      \subfloat[NOAA weather data set]
     {\label{fig_ComparativeEntireGraphNOAARMSEResults}\includegraphics[height=3.4cm]{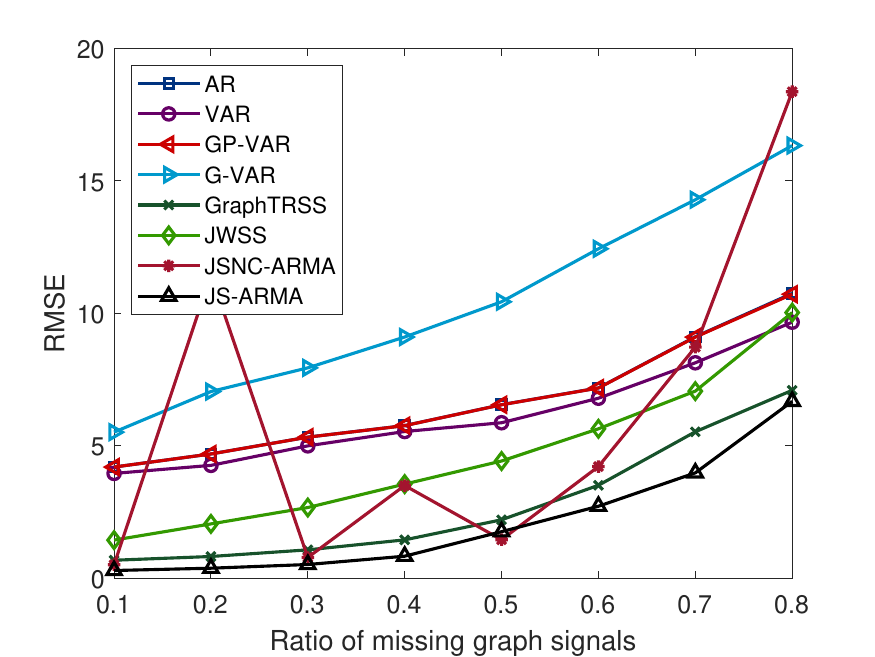}}
 \end{center}
	\caption{RMSE estimation errors for Scenario \ref{sco_entire_graph}}
 \label{fig_rmse_comparative_entire_graph}
\end{figure}

\begin{figure}[h]
\begin{center}
     \subfloat[Mol\`ene weather data set]
     {\label{fig_ComparativeEntireGraphMoleneMAEResults}\includegraphics[height=3.4cm]{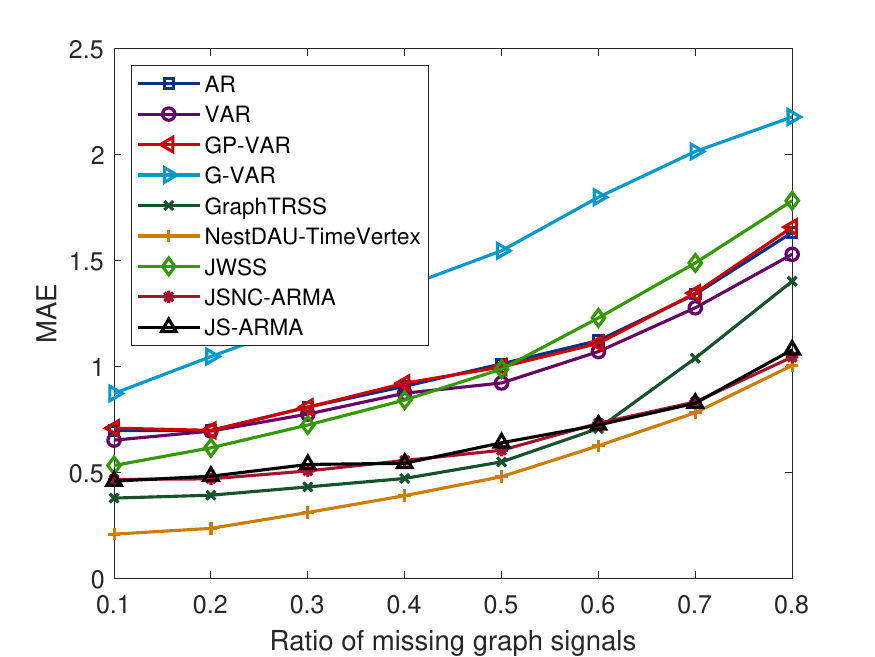}}
     \subfloat[COVID-19 data set]
     {\label{fig_ComparativeEntireGraphCovid19MAEResults}\includegraphics[height=3.4cm]{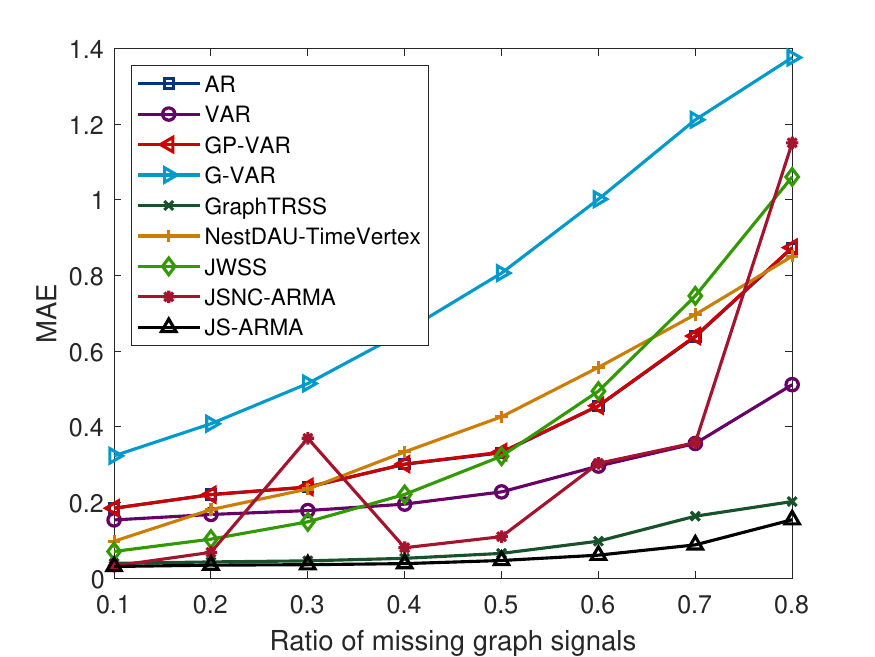}}\\
           \subfloat[NOAA weather data set]
     {\label{fig_ComparativeEntireGraphNOAAMAEResults}\includegraphics[height=3.4cm]{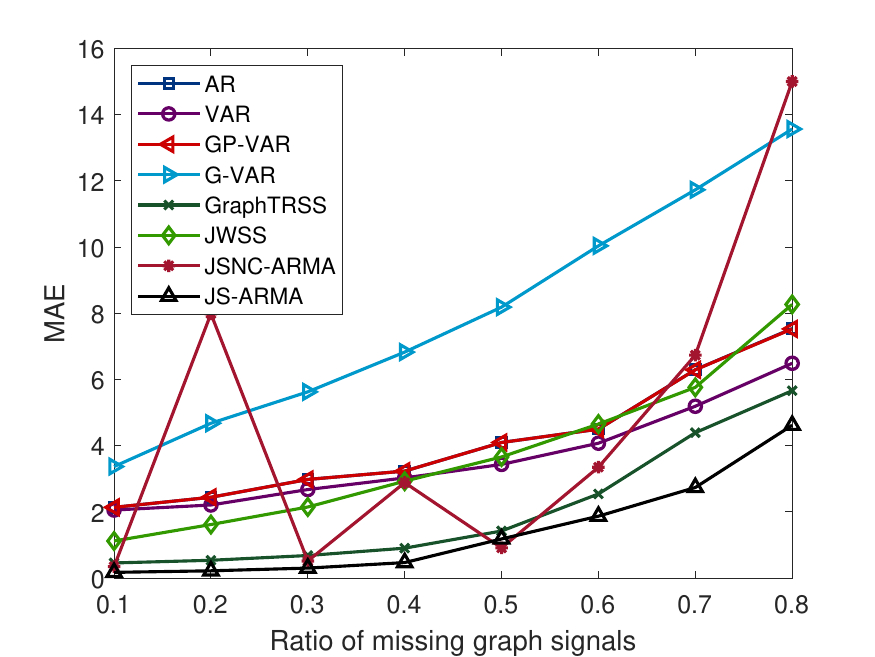}}
 \end{center}
	\caption{MAE estimation errors for Scenario \ref{sco_entire_graph}}
 \label{fig_mae_comparative_entire_graph}
\end{figure}

\begin{figure}[h]
\begin{center}
     \subfloat[Mol\`ene weather data set]
     {\label{fig_ComparativeEntireGraphMoleneMAPEResults}\includegraphics[height=3.4cm]{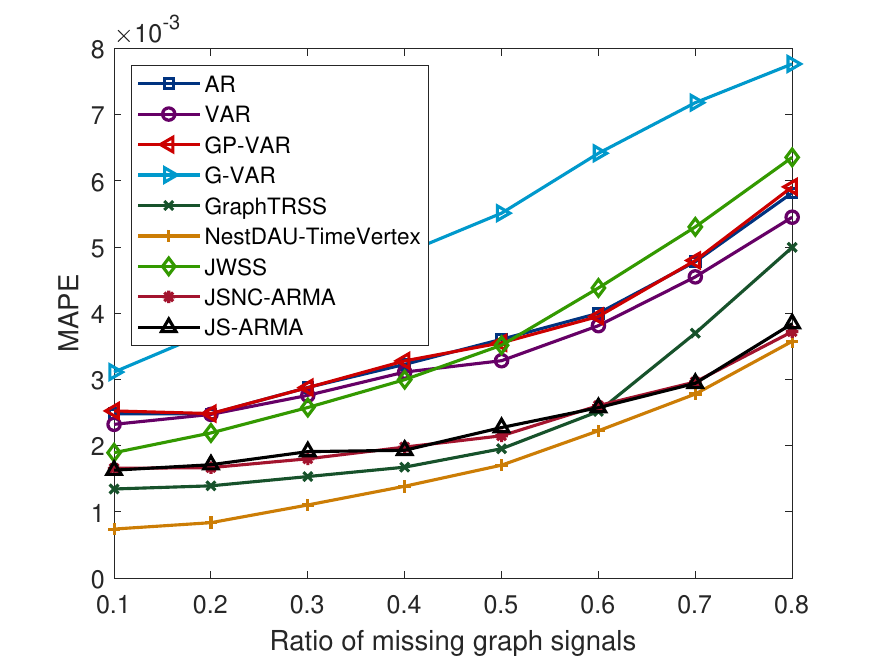}}
     \subfloat[COVID-19 data set]
     {\label{fig_ComparativeEntireGraphCovid19MAPEResults}\includegraphics[height=3.4cm]{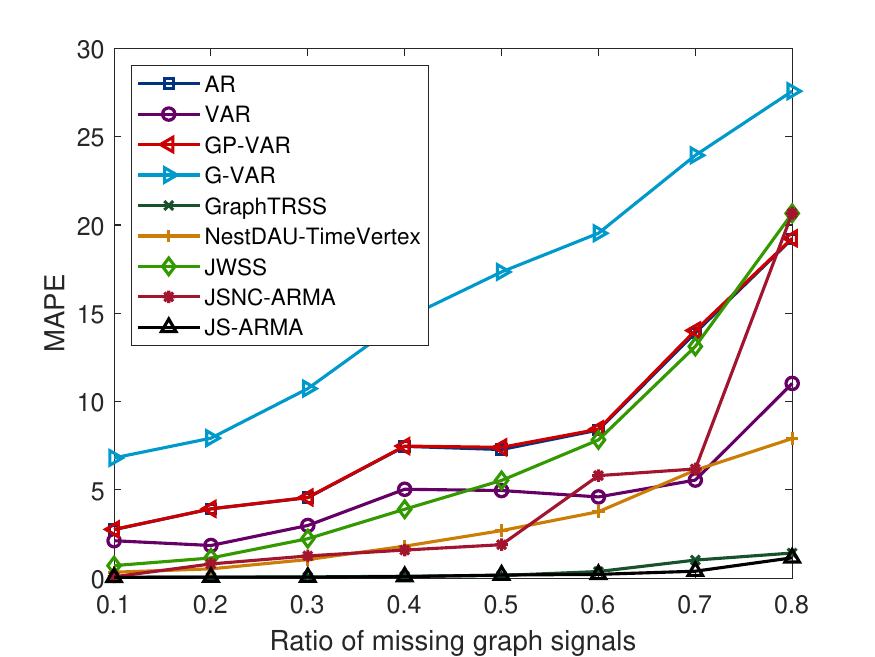}}\\
           \subfloat[NOAA weather data set]
     {\label{fig_ComparativeEntireGraphNOAAMAPEResults}\includegraphics[height=3.4cm]{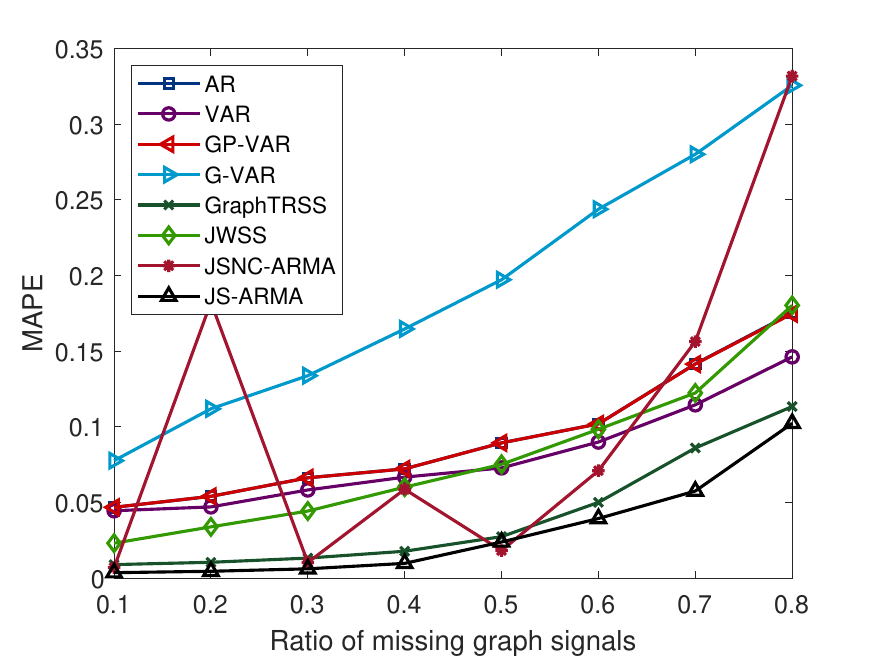}}
 \end{center}
	\caption{MAPE estimation errors for Scenario \ref{sco_entire_graph}}
 \label{fig_mape_comparative_entire_graph}
\end{figure}


\begin{figure}[t!]
\begin{center}
     \subfloat[Mol\`ene weather data set]
     {\label{fig_ComparativeForecastingMoleneRMSEResults}\includegraphics[height=3.4cm]{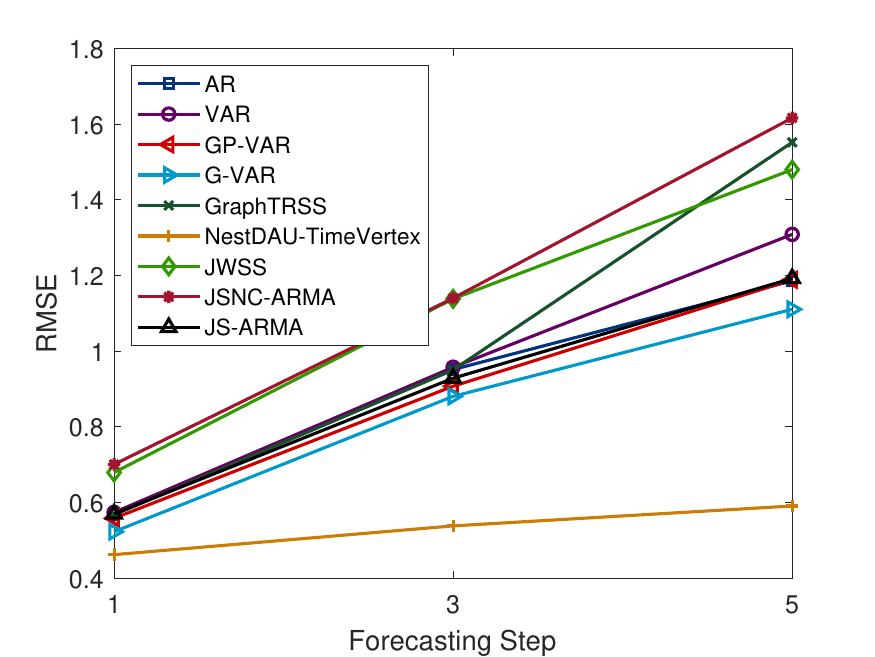}}
     \subfloat[COVID-19 data set]
     {\label{fig_ComparativeForecastingCovid19RMSEResults}\includegraphics[height=3.4cm]{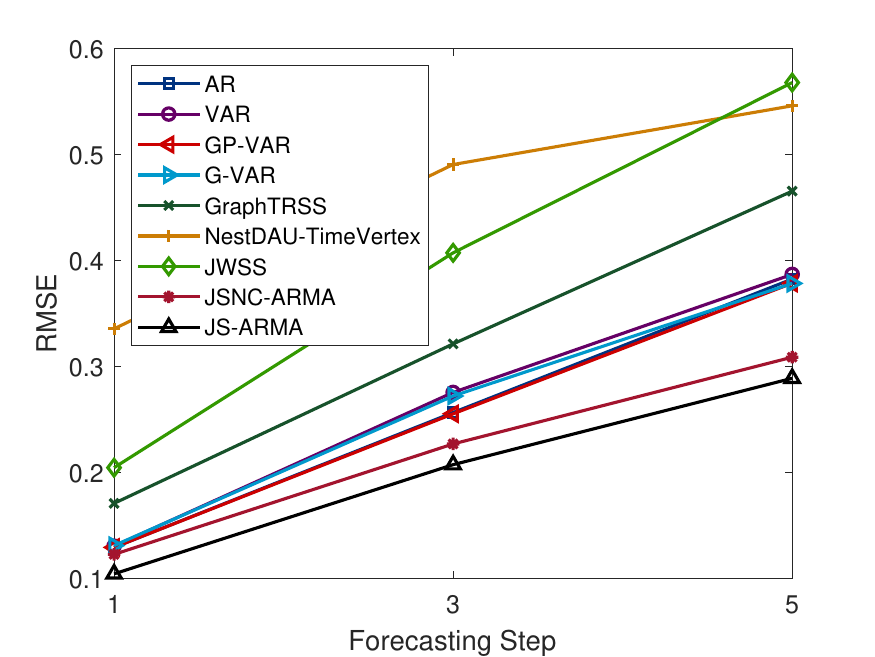}}\\
          \subfloat[NOAA weather data set]
     {\label{fig_ComparativeForecastingNOAARMSEResults}\includegraphics[height=3.4cm]{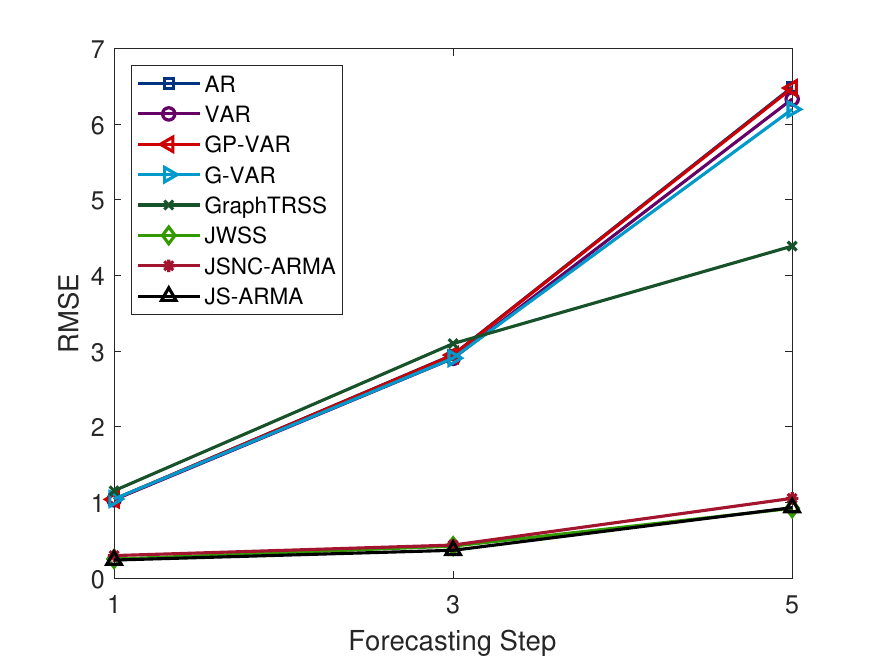}}\\
 \end{center}
	\caption{RMSE estimation errors for Scenario \ref{sco_forecasting} (Forecasting)}
 \label{fig_rmse_comparative_forecasting}
\end{figure}

\begin{figure}[h]
\begin{center}
     \subfloat[Mol\`ene weather data set]
     {\label{fig_ComparativeForecastingMoleneMAEResults}\includegraphics[height=3.4cm]{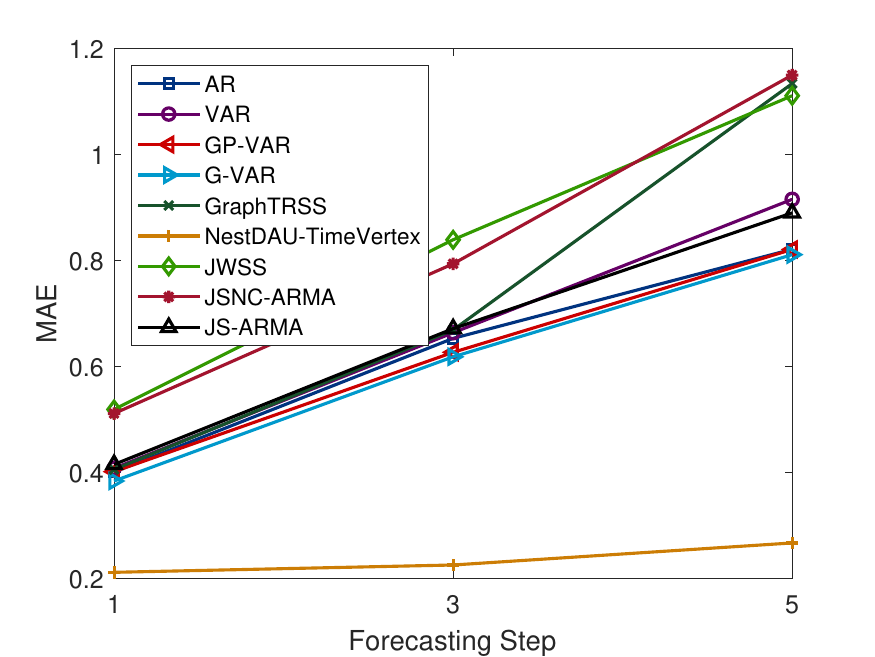}}
     \subfloat[COVID-19 data set]
     {\label{fig_ComparativeForecastingCovid19MAEResults}\includegraphics[height=3.4cm]{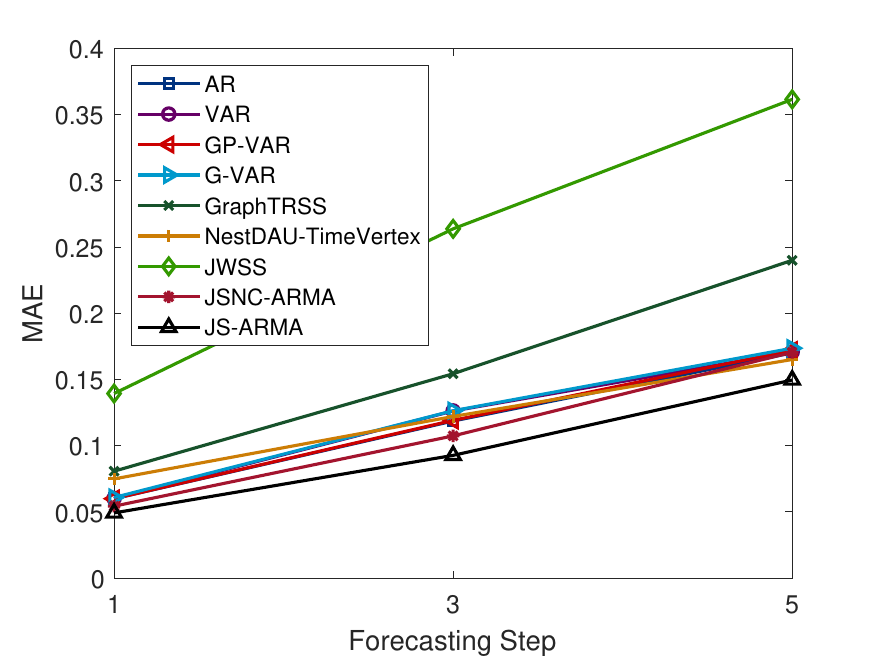}}\\
       \subfloat[NOAA weather data set]
     {\label{fig_ComparativeForecastingNOAAMAEResults}\includegraphics[height=3.4cm]{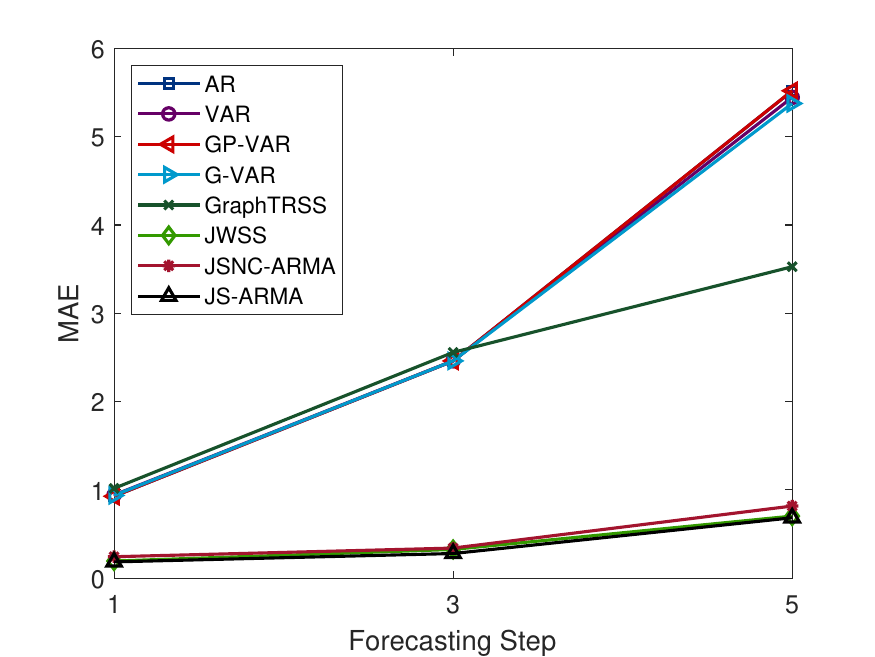}}\\
 \end{center}
	\caption{MAE estimation errors for Scenario \ref{sco_forecasting} (Forecasting)}
 \label{fig_mae_comparative_forecasting}
\end{figure}

\begin{figure}[h]
\begin{center}
     \subfloat[Mol\`ene weather data set]
     {\label{fig_ComparativeForecastingMoleneMAPEResults}\includegraphics[height=3.4cm]{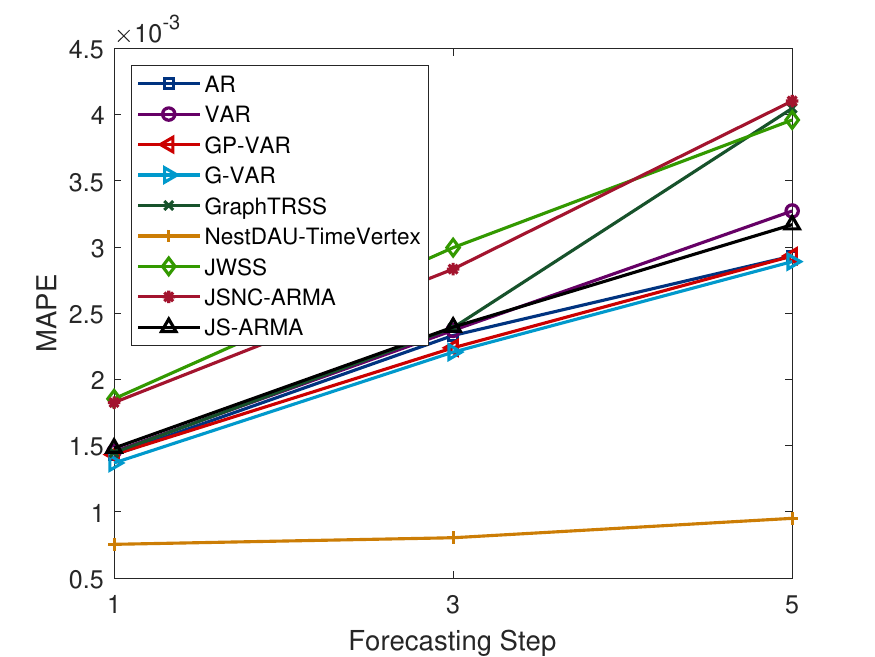}}
     \subfloat[COVID-19 data set]
     {\label{fig_ComparativeForecastingCovid19MAPEResults}\includegraphics[height=3.4cm]{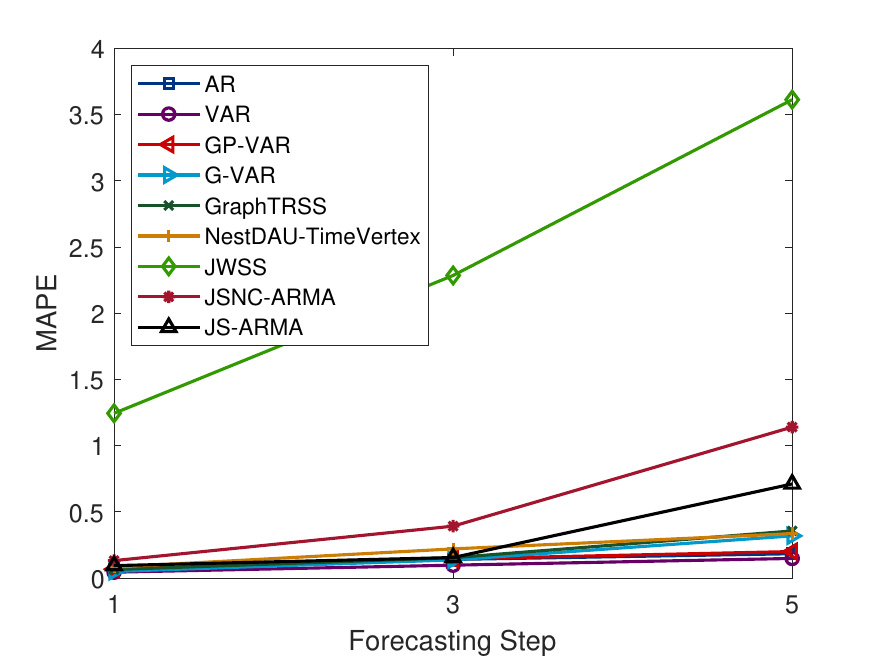}}\\
     \subfloat[NOAA weather data set]
     {\label{fig_ComparativeForecastingNOAAMAPEResults}\includegraphics[height=3.4cm]{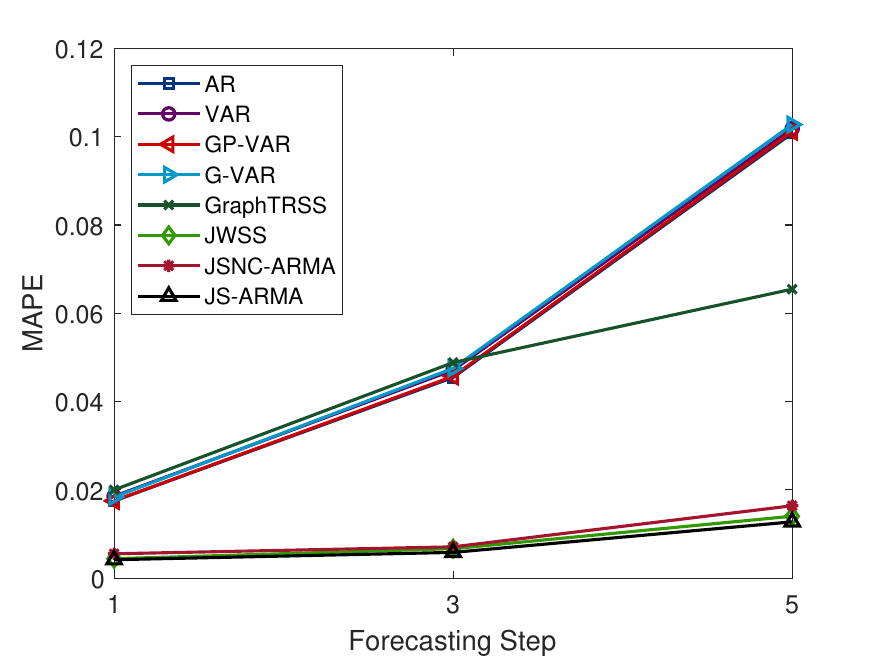}}\\
 \end{center}
	\caption{MAPE estimation errors for Scenario \ref{sco_forecasting} (Forecasting)}
 \label{fig_mape_comparative_forecasting}
\end{figure}

\bibliographystyle{IEEEtran}
\bibliography{refs.bib}

\begin{thebibliography}{10}
\providecommand{\url}[1]{#1}
\csname url@samestyle\endcsname
\providecommand{\newblock}{\relax}
\providecommand{\bibinfo}[2]{#2}
\providecommand{\BIBentrySTDinterwordspacing}{\spaceskip=0pt\relax}
\providecommand{\BIBentryALTinterwordstretchfactor}{4}
\providecommand{\BIBentryALTinterwordspacing}{\spaceskip=\fontdimen2\font plus
\BIBentryALTinterwordstretchfactor\fontdimen3\font minus
  \fontdimen4\font\relax}
\providecommand{\BIBforeignlanguage}[2]{{%
\expandafter\ifx\csname l@#1\endcsname\relax
\typeout{** WARNING: IEEEtran.bst: No hyphenation pattern has been}%
\typeout{** loaded for the language `#1'. Using the pattern for}%
\typeout{** the default language instead.}%
\else
\language=\csname l@#1\endcsname
\fi
#2}}
\providecommand{\BIBdecl}{\relax}
\BIBdecl

\bibitem{Girault-Isometric-shift}
B.~{Girault}, P.~{Gon\c{c}alves}, and E.~{Fleury}, ``Translation on graphs: An
  isometric shift operator,'' \emph{IEEE Signal Processing Letters}, vol.~22,
  no.~12, pp. 2416--2420, Dec 2015.

\bibitem{Natheneal-Graph-Stationarity}
N.~{Perraudin} and P.~{Vandergheynst}, ``Stationary signal processing on
  graphs,'' \emph{IEEE Transactions on Signal Processing}, vol.~65, no.~13, pp.
  3462--3477, July 2017.

\bibitem{Marques-Stationarity}
A.~G. {Marques}, S.~{Segarra}, G.~{Leus}, and A.~{Ribeiro}, ``Stationary graph
  processes and spectral estimation,'' \emph{IEEE Trans. Sig. Proc.}, vol.~65,
  no.~22, Nov 2017.

\bibitem{Natheneal-Joint-Stationarity}
A.~{Loukas} and N.~{Perraudin}, ``Stationary time-vertex signal processing,''
  \emph{EURASIP Journal on Advances in Signal Processing}, vol. 2019, no.~1,
  p.~36, 2019.

\bibitem{MeiM17}
J.~Mei and J.~M.~F. Moura, ``Signal processing on graphs: Causal modeling of
  unstructured data,'' \emph{{IEEE} Trans. Signal Process.}, vol.~65, no.~8,
  pp. 2077--2092, 2017.

\bibitem{IsufiLPL19}
E.~Isufi, A.~Loukas, N.~Perraudin, and G.~Leus, ``Forecasting time series with
  {VARMA} recursions on graphs,'' \emph{{IEEE} Trans. Signal Process.},
  vol.~67, no.~18, pp. 4870--4885, 2019.

\bibitem{ZhouBLWS03}
D.~Zhou, O.~Bousquet, T.~N. Lal, J.~Weston, and B.~Sch{\"{o}}lkopf, ``Learning
  with local and global consistency,'' in \emph{Advances in Neural Inf. Proc.
  Sys.}, 2003, pp. 321--328.

\bibitem{JungHMJHE19}
A.~Jung, A.~O. Hero~III, A.~C. Mara, S.~Jahromi, A.~Heimowitz, and Y.~C. Eldar,
  ``Semi-supervised learning in network-structured data via total variation
  minimization,'' \emph{{IEEE} Trans. Signal Process.}, vol.~67, no.~24, pp.
  6256--6269, 2019.

\bibitem{BergerHM17}
P.~Berger, G.~Hannak, and G.~Matz, ``Graph signal recovery via primal-dual
  algorithms for total variation minimization,'' \emph{{IEEE} J. Sel. Top.
  Signal Process.}, vol.~11, no.~6, pp. 842--855, 2017.

\bibitem{GiraldoMGTB22}
J.~H. Giraldo, A.~Mahmood, B.~Garc{\'{\i}}a-Garc{\'{\i}}a, D.~Thanou, and
  T.~Bouwmans, ``Reconstruction of time-varying graph signals via {S}obolev
  smoothness,'' \emph{{IEEE} Trans. Signal Inf. Process. over Networks},
  vol.~8, pp. 201--214, 2022.

\bibitem{JiangTSO20}
J.~Jiang, D.~Tay, Q.~Sun, and S.~Ouyang, ``Recovery of time-varying graph
  signals via distributed algorithms on regularized problems,'' \emph{{IEEE}
  Trans. Signal Inf. Process. over Networks}, vol.~6, pp. 540--555, 2020.

\bibitem{QiuMSWLG17}
K.~Qiu, X.~Mao, X.~Shen, X.~Wang, T.~Li, and Y.~Gu, ``Time-varying graph signal
  reconstruction,'' \emph{{IEEE} J. Sel. Top. Signal Process.}, vol.~11, no.~6,
  pp. 870--883, 2017.

\bibitem{ChenE21a}
S.~Chen and Y.~C. Eldar, ``Time-varying graph signal inpainting via unrolling
  networks,'' in \emph{{IEEE} International Conference on Acoustics, Speech and
  Signal Processing}, 2021, pp. 8092--8097.

\bibitem{KojimaNYT23}
H.~Kojima, H.~Noguchi, K.~Yamada, and Y.~Tanaka, ``Restoration of time-varying
  graph signals using deep algorithm unrolling,'' in \emph{{IEEE} International
  Conference on Acoustics, Speech and Signal Processing}, 2023.

\bibitem{NagahamaYTCE22}
M.~Nagahama, K.~Yamada, Y.~Tanaka, S.~H. Chan, and Y.~C. Eldar, ``Graph signal
  restoration using nested deep algorithm unrolling,'' \emph{{IEEE} Trans.
  Signal Process.}, vol.~70, pp. 3296--3311, 2022.

\bibitem{HadouKR23}
S.~Hadou, C.~I. Kanatsoulis, and A.~Ribeiro, ``Space-time graph neural networks
  with stochastic graph perturbations,'' in \emph{IEEE Int. Conf. Acoustics,
  Speech and Signal Proc.}, 2023, pp. 1--5.

\bibitem{CastroCorreaGMBBM23}
J.~A. Castro-Correa~et al., ``Time-varying signals recovery via graph neural
  networks,'' in \emph{IEEE Int. Conf. Acoustics, Speech and Signal Proc.},
  2023, pp. 1--5.

\bibitem{LorenzoBBS16}
P.~Di~Lorenzo, S.~Barbarossa, P.~Banelli, and S.~Sardellitti, ``Adaptive least
  mean squares estimation of graph signals,'' \emph{{IEEE} Trans. Signal Inf.
  Process. over Networks}, vol.~2, no.~4, pp. 555--568, 2016.

\bibitem{LorenzoBIBL18}
P.~Di~Lorenzo, P.~Banelli, E.~Isufi, S.~Barbarossa, and G.~Leus, ``Adaptive
  graph signal processing: Algorithms and optimal sampling strategies,''
  \emph{{IEEE} Trans. Signal Process.}, vol.~66, no.~13, pp. 3584--3598, 2018.

\bibitem{YangYYH21}
G.~Yang, L.~Yang, Z.~Yang, and C.~Huang, ``Efficient node selection strategy
  for sampling bandlimited signals on graphs,'' \emph{{IEEE} Trans. Signal
  Process.}, vol.~69, pp. 5815--5829, 2021.

\bibitem{ChamonR18}
L.~F.~O. Chamon and A.~Ribeiro, ``Greedy sampling of graph signals,''
  \emph{{IEEE} Trans. Signal Process.}, vol.~66, no.~1, pp. 34--47, 2018.

\bibitem{IsufiBLL20}
E.~Isufi, P.~Banelli, P.~Di~Lorenzo, and G.~Leus, ``Observing and tracking
  bandlimited graph processes from sampled measurements,'' \emph{Signal
  Process.}, vol. 177, p. 107749, 2020.

\bibitem{JiangFTX21}
J.~Jiang, H.~Feng, D.~B. Tay, and S.~Xu, ``Theory and design of joint
  time-vertex nonsubsampled filter banks,'' \emph{{IEEE} Trans. Signal
  Process.}, vol.~69, pp. 1968--1982, 2021.

\bibitem{OrtizJimenezCC18}
G.~Ortiz{-}Jim{\'{e}}nez, M.~Coutino, S.~P. Chepuri, and G.~Leus, ``Sampling
  and reconstruction of signals on product graphs,'' in \emph{2018 {IEEE}
  Global Conference on Signal and Information Processing}.\hskip 1em plus 0.5em
  minus 0.4em\relax {IEEE}, 2018, pp. 713--717.

\bibitem{RomeroIG17}
D.~Romero, V.~N. Ioannidis, and G.~B. Giannakis, ``Kernel-based reconstruction
  of space-time functions on dynamic graphs,'' \emph{{IEEE} J. Sel. Top. Signal
  Process.}, vol.~11, no.~6, pp. 856--869, 2017.

\bibitem{Girault-Stationarity}
B.~{Girault}, ``Stationary graph signals using an isometric graph
  translation,'' in \emph{2015 23rd European Signal Processing Conference
  (EUSIPCO)}, Aug 2015, pp. 1516--1520.

\bibitem{Lutkepohl05}
H.~L\"utkepohl, \emph{New introduction to multiple time series analysis}.\hskip
  1em plus 0.5em minus 0.4em\relax Springer, 2005.

\bibitem{Jung15}
A.~Jung, ``Learning the conditional independence structure of stationary time
  series: {A} multitask learning approach,'' \emph{{IEEE} Trans. Signal
  Process.}, vol.~63, no.~21, pp. 5677--5690, 2015.

\bibitem{Natheneal-Towards-Joint-Stationarity}
N.~{Perraudin}, A.~{Loukas}, F.~{Grassi}, and P.~{Vandergheynst}, ``Towards
  stationary time-vertex signal processing,'' in \emph{Proc. IEEE ICASSP},
  2017, pp. 3914--3918.

\bibitem{Time-vertex-signal-processing}
F.~{Grassi}, A.~{Loukas}, N.~{Perraudin}, and B.~{Ricaud}, ``A time-vertex
  signal processing framework: Scalable processing and meaningful
  representations for time-series on graphs,'' \emph{IEEE Transactions on
  Signal Processing}, vol.~66, no.~3, pp. 817--829, Feb 2018.

\bibitem{IsufiLSL16}
E.~Isufi, A.~Loukas, A.~Simonetto, and G.~Leus, ``Separable autoregressive
  moving average graph-temporal filters,'' in \emph{Proc. 24th {EUSIPCO}},
  2016, pp. 200--204.

\bibitem{hayes96}
M.~Hayes, \emph{Statistical Digital Signal Processing and Modeling}.\hskip 1em
  plus 0.5em minus 0.4em\relax Wiley, 1996.

\bibitem{GuneyiCV21}
E.~T. G{\"{u}}neyi, A.~Canbolat, and E.~Vural, ``Learning parametric
  time-vertex graph processes from incomplete realizations,'' in \emph{{IEEE}
  Int. Workshop Machine Learning for Signal Processing}, 2021.

\bibitem{DavidEmerging}
D.~I. {Shuman}, S.~K. {Narang}, P.~{Frossard}, A.~{Ortega}, and
  P.~{Vandergheynst}, ``The emerging field of signal processing on graphs:
  Extending high-dimensional data analysis to networks and other irregular
  domains,'' \emph{IEEE Signal Processing Magazine}, vol.~30, no.~3, pp.
  83--98, May 2013.

\bibitem{cvx}
M.~Grant and S.~Boyd, ``{CVX}: Matlab software for disciplined convex
  programming, version 2.1,'' Mar. 2014.

\bibitem{gb08}
------, ``Graph implementations for nonsmooth convex programs,'' in
  \emph{Recent Advances in Learning and Control}.\hskip 1em plus 0.5em minus
  0.4em\relax Springer-Verlag Limited, 2008, pp. 95--110.

\bibitem{optimization-SDPT3}
R.~H. T\"ut\"unc\"u, K.~C. Toh, and M.~J. Todd, ``{SDPT3} — a {MATLAB}
  software package for semidefinite programming, version 1.3,'' vol. 11:1-4,
  pp. 545--581, 1999.

\bibitem{optimization-quadratic-linear}
------, ``Solving semidefinite-quadratic-linear programs using {SDPT3},'' vol.
  95:2, pp. 189--217, 2003.

\bibitem{Covid19data}
\BIBentryALTinterwordspacing
``{COVID}-19 coronavirus pandemic data.'' [Online]. Available:
  \url{https://www.worldometers.info/coronavirus/}
\BIBentrySTDinterwordspacing

\bibitem{Eurostat}
\BIBentryALTinterwordspacing
``Eurostat: An official website of the {E}uropean {U}nion.'' [Online].
  Available: \url{https://ec.europa.eu/eurostat}
\BIBentrySTDinterwordspacing

\bibitem{Arguez12}
{A. Arguez et al.}, ``{NOAA}’s 1981-2020 {U.S.} climate normals: {A}n
  overview,'' in \emph{Proc. Bull. Amer. Meteorol. Soc.}, 2012, pp. 1687--1697.

\end{thebibliography}

%
%
%
%

\vfill

\end{document}